%% file: viplan.tex
\documentclass[11pt]{article}

\usepackage[final]{acl}
\usepackage{times}
\usepackage{latexsym}
\usepackage[T1]{fontenc}
\usepackage[utf8]{inputenc}
\usepackage{microtype}
\usepackage{inconsolata}
\usepackage{graphicx}
\usepackage{amsmath}
\usepackage{amssymb}

\usepackage{booktabs}

\usepackage{xcolor}      
\definecolor{myblue}{RGB}{68,166,216}
\definecolor{myred}{RGB}{244,99,103}
\usepackage{comment}
\usepackage{caption}
\usepackage{multirow}
\usepackage{tcolorbox}
\usepackage{subcaption}

\tcbuselibrary{listings,skins,breakable}

\newtcblisting{pddlcode}[1][]{
  listing engine=listings,
  enhanced,
  breakable,
  colback=gray!10,
  colframe=gray!70,
  boxrule=0.5pt,
  arc=1mm,
  outer arc=1mm,
  left=2mm,
  right=2mm,
  top=1mm,
  bottom=1mm,
  #1,
  listing only,
  listing options={
    basicstyle=\ttfamily\small,
    keywordstyle=\bfseries\color{blue!70!black},
    commentstyle=\itshape\color{gray!60},
    stringstyle=\color{red!50!black},
    morekeywords={define,domain,action,parameters,precondition,effect,objects,init,goal,requirements,types},
    alsoletter={:},
    literate={:}{\textcolor{blue!70!black}{:}}1,
    breaklines=true,
    keepspaces=true,
    columns=fullflexible,
  }
}

\newtcblisting{prompt}[1][]{
  listing engine=listings,
  enhanced, breakable,
  colback=gray!10, colframe=gray!70,
  boxrule=0.5pt, arc=1mm, outer arc=1mm,
  left=2mm, right=2mm, top=1mm, bottom=1mm,
  #1, listing only,
  listing options={
    basicstyle=\small,
    keywordstyle=\bfseries\color{blue!70!black},
    morekeywords={
      \<system\>,\</system\>,
      \<user\>,\</user\>,
      \<explanation\>,\</explanation\>,
      \<answer\>,\</answer\>,
      \{image\}
    },
    alsoletter={\<,\>,/},
    literate=
      {<system>}{{\color{blue!70!black}<system>}}1
      {</system>}{{\color{blue!70!black}</system>}}1
      {<user>}{{\color{blue!70!black}<user>}}1
      {</user>}{{\color{blue!70!black}</user>}}1
      {<explanation>}{{\color{blue!70!black}<explanation>}}1
      {</explanation>}{{\color{blue!70!black}</explanation>}}1
      {<answer>}{{\color{blue!70!black}<answer>}}1
      {</answer>}{{\color{blue!70!black}</answer>}}1
      {\{image\}}{{\color{red!70!black}\{image\}}}1,
    breaklines=true,
    breakindent=0pt,
    breakautoindent=false,
    columns=fullflexible,
  }
}

\title{ViPlan: A Benchmark for Visual Planning with Symbolic Predicates and Vision-Language Models}

\author{
  \textbf{Matteo Merler\textsuperscript{1,2}}\thanks{Equal contribution.},
  \textbf{Nicola Dainese\textsuperscript{1}}\footnotemark[1],
  \textbf{Minttu Alakuijala\textsuperscript{1}},
  \textbf{Giovanni Bonetta\textsuperscript{2}},
  \textbf{Pietro Ferrazzi\textsuperscript{2,3}},
  \\
  \textbf{Yu Tian\textsuperscript{1}},
  \textbf{Bernardo Magnini\textsuperscript{2}},
  \textbf{Pekka Marttinen\textsuperscript{1}}
\\
  \textsuperscript{1}Department of Computer Science, Aalto University, Espoo, Finland
\\
  \textsuperscript{2}Fondazione Bruno Kessler, Trento, Italy
\\
  \textsuperscript{3}Department of Mathematics, Università degli Studi di Padova, Padova, Italy
\\
    \textbf{Correspondence:} \texttt{pekka.marttinen@aalto.fi}
}

\begin{document}
\maketitle
\begin{abstract}
  Integrating Large Language Models with symbolic planners is a promising direction for obtaining verifiable and grounded plans, with recent works extending this idea to visual domains using Vision-Language Models (VLMs). 
  However, an open-source benchmark for comparing these approaches under matched conditions is missing, due to a lack of visual benchmarks that support symbolic planning.
  We present \textbf{ViPlan}, the first open-source benchmark for comparing VLM-grounded symbolic approaches (VLM-as-grounder) with direct VLM planning methods (VLM-as-planner).
  ViPlan introduces a series of increasingly challenging tasks in two visual domains: a visual variant of the classic Blocksworld planning problem and a simulated household robotics environment. 
  Averaged across methods, we find VLM-as-grounders to outperform direct VLM planning in Blocksworld (solving 46\% of the tasks against 9\%), where image grounding is both crucial and accurate. However, in the household robotics tasks, where linguistic knowledge helps, VLM-as-planner methods are greatly superior to VLM-as-grounder approaches (solving 34\% of the tasks against 5\%), which are hindered by partial observability. 
  Thus, ViPlan domains capture fundamental shortcomings of both planning approaches, which we further diagnose with a qualitative failure analysis.
  Finally, across methods, we observe no consistent benefit from Chain-of-Thought prompting, suggesting persistent limitations in current VLMs’ visual reasoning abilities.
\end{abstract}

\section{Introduction}
\label{sec:intro}

Automated planning is a fundamental capability for general-purpose agents, consisting of the ability to generate action plans based on their knowledge of the environment they act in. This enables them to make decisions, adapt to dynamic environments and achieve complex goals~\citep{ghallab2004automated}. The rise in popularity of foundation models~\citep{bommasani2021opportunities}, including Large Language Models (LLMs), has prompted many to test their ability for planning~\citep{ahn2022can, huang2022language, huang2023inner, hao2023reasoning}. Others have been more critical, arguing that LLMs remain unreliable and lack the capacity for formal planning~\citep{valmeekam2023planning, kambhampati2024position} and recommending to integrate them with symbolic planners. Such planners rely on formal logic, for example using the Planning Domain Definition Language~\citep[PDDL;][]{mcdermott1998pddl}, to specify the relevant properties of the world state and how actions affect them in order to solve tasks.

\begin{figure*}[t]
    \centering
    \includegraphics[width=\linewidth]{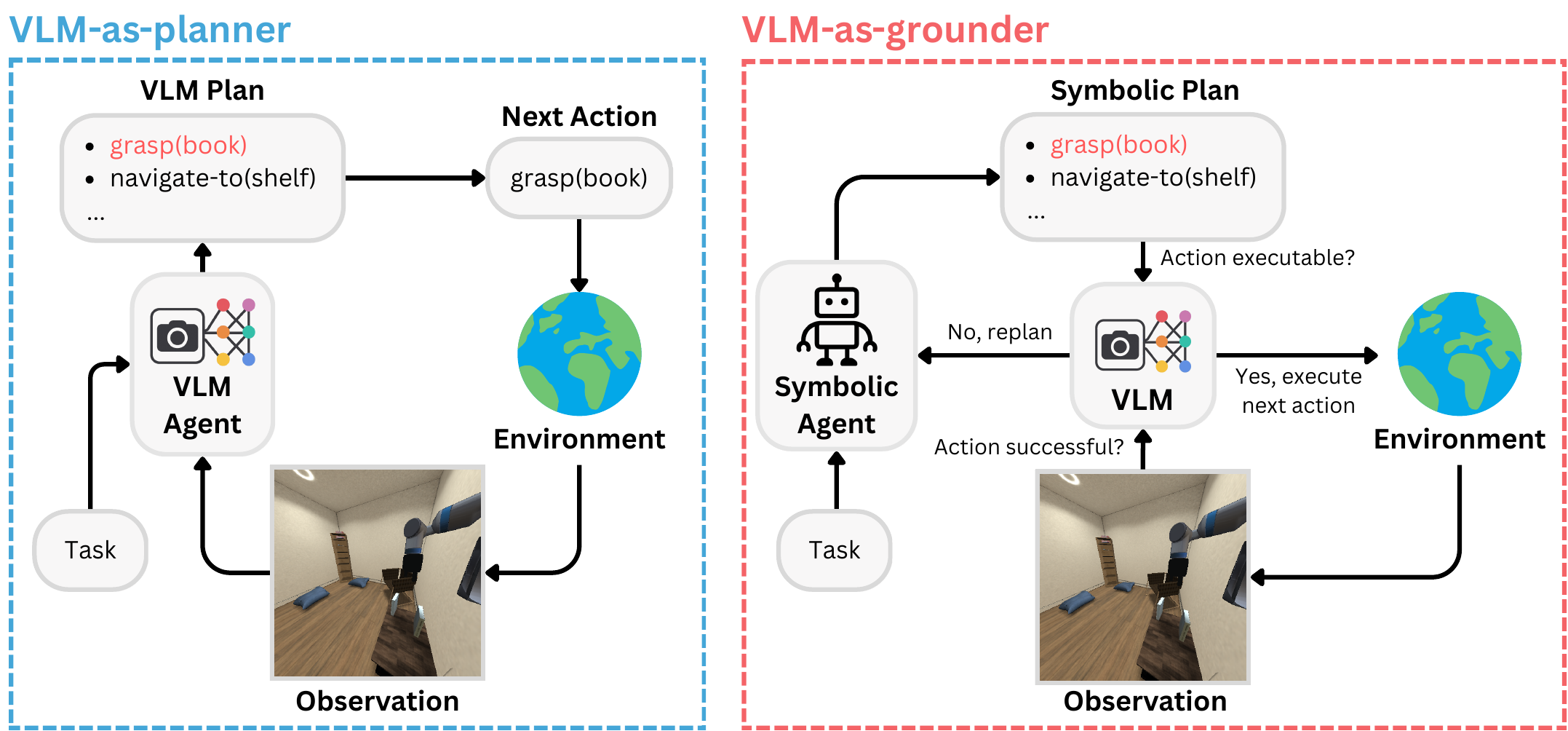}
    \caption{\textbf{Planning with VLMs.} Two classes of VLM-based methods for planning a set of actions to reach a goal. \textcolor{myblue}{\textbf{VLM-as-planner}} uses the VLM directly to produce a new plan after every action. \textcolor{myred}{\textbf{VLM-as-grounder}} uses the VLM to ground a symbolic agent's plans to the observations from the environment. Grounding takes the form of yes-no question-answering about whether the conditions that make an action executable are met and whether the expected outcomes of the action are realized. }
    \label{fig:figure1}
\end{figure*}

Vision-Language Models~\citep[VLMs;][]{liu2023visual, li2023blip} have gathered interest for their ability to process visual observations jointly with language. Inspired by LLM approaches to planning, this has led to two complementary uses of VLMs, distinguished by the role of the model (see Figure~\ref{fig:figure1}). The first, which we name \textcolor{myblue}{\textbf{VLM-as-planner}}, uses the model to directly select actions from visual inputs. The second, \textcolor{myred}{\textbf{VLM-as-grounder}}, instead uses the VLM to ground symbolic representations of the world, such as the boolean variables defined in PDDL, named \emph{predicates}, in perception. The latter is particularly important in open-world domains, such as embodied AI tasks, where the state of the world can change in ways that symbolic planners alone cannot handle due to their lack of built-in perceptual grounding.

However, recent evidence has shown that current state-of-the-art VLMs can still struggle with detecting objects~\citep{augustin2025dash} and identifying relationships between entities~\citep{tong2024eyes}, crucial abilities for grounding planners to observations. Furthermore, they inherit the same limitations of LLMs in terms of formal planning abilities.
As a result, both approaches exhibit distinct failure modes, yet their relative strengths and weaknesses remain poorly understood. To date, no open-source benchmark enables a comparison of the two under matched environments and protocols, with previous studies either focusing on direct VLM planning~\citep{liu2025visualagentbench}, or working with private benchmarks and closed-source models~\citep{zhang2025dkprompt}, making it difficult to assess when each approach is preferable.

To address this gap, we present \textbf{ViPlan}, the first open-source visual planning benchmark designed to compare VLM-as-planner and VLM-as-grounder approaches.
Our main contributions are:
\begin{enumerate}
    \item We introduce \textbf{ViPlan}, an open-source\footnote{We open-source the benchmark at \url{https://github.com/merlerm/ViPlan}.} and extensible benchmark for comparing VLM-as-planner and VLM-as-grounder approaches under a matched protocol, comprising two interactive domains: ViPlan-Blocksworld (ViPlan-BW) and ViPlan-Household (ViPlan-HH), with three difficulty splits of 25 tasks each. We screen 21 open-source VLMs and evaluate eight method variants on the five models selected for the full benchmark.

    \item We find a domain-dependent trade-off: VLM-as-grounder outperforms VLM-as-planner in ViPlan-BW (46\% vs.\ 9\%), where visual grounding is reliable and precise state tracking is crucial, while the opposite is true in ViPlan-HH (5\% vs.\ 34\%), where grounding errors compound and direct VLM planners can better exploit linguistic priors. Supporting blank-image and predicate-level analyses corroborate this explanation.

    \item We analyze the behavior and failure modes of both paradigms, finding no consistent benefit from Chain-of-Thought prompting~\citep[CoT;][]{wei2022chain}, even after controlling for generation-budget effects, and identifying recurring grounding, non-executable-action, and parsing failures.
\end{enumerate}

\section{Background}
\label{sec:background}

\begin{figure*}[t]
    \centering
    \includegraphics[width=\linewidth]{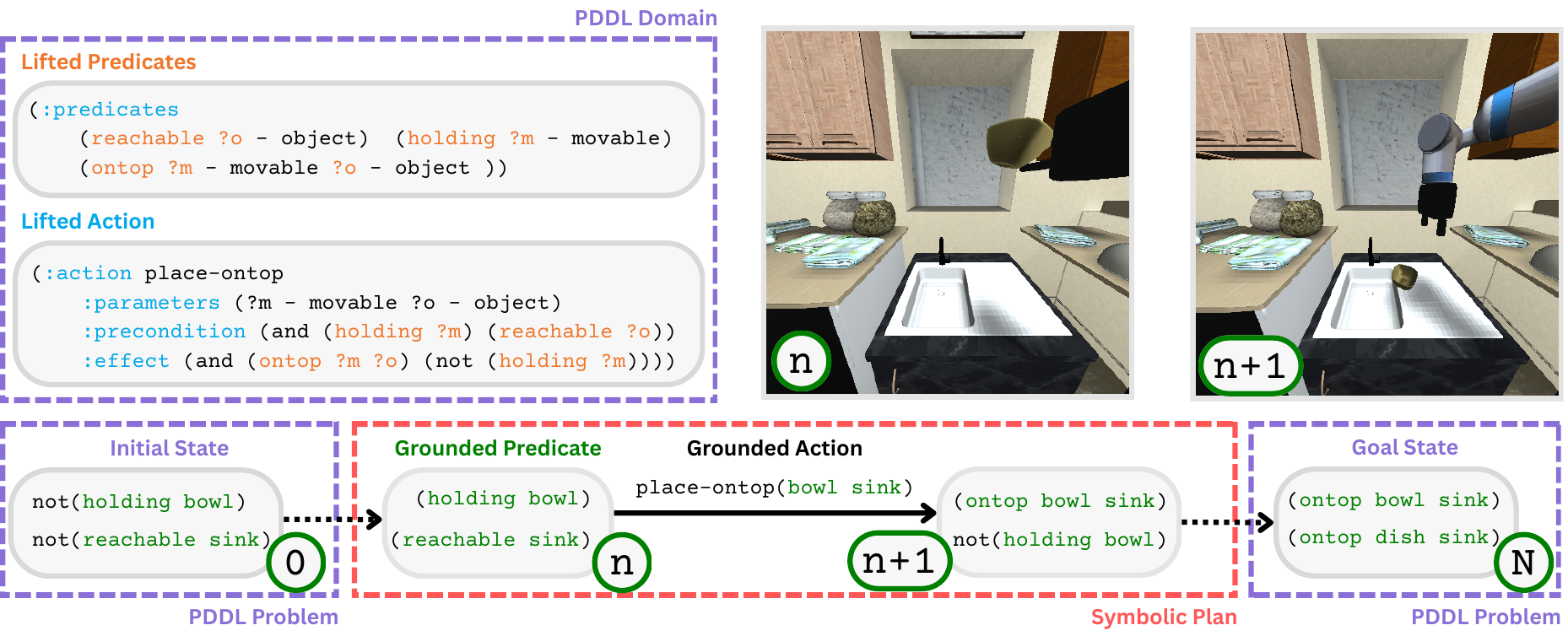}
    \caption{\textbf{Example of the basic components for formal planning with PDDL.} A PDDL domain includes the list of possible lifted actions, which are then grounded by a PDDL problem, that provides the initial and goal state. A symbolic planner takes as input the PDDL domain and problem to generate a symbolic plan to reach the desired goal state through a sequence of \textit{N} action.}
    \label{fig:pddl_diagram}
\end{figure*}

We investigate \textbf{classical planning problems}~\citep{ghallab2004automated}. These involve finding a sequence of actions that get an agent to achieve a goal, given a model of the environment.

A classical planning problem is defined as a tuple $\Pi = \langle \mathcal{D}, \mathcal{I}, \mathcal{G} \rangle$, where $\mathcal{D}$ is the problem \textbf{domain}, $\mathcal{I}$ is the \textbf{initial state} and $\mathcal{G}$ is the desired \textbf{goal state}. 
The domain $\mathcal{D}$ is further defined as $\mathcal{D} = \langle \mathcal{P}, \mathcal{A} \rangle$, where $\mathcal{P}$ is the set of all boolean \textbf{predicates} used to describe the environment's state, and $\mathcal{A}$ is the set of all \textbf{actions} the agent can perform. See Figure~\ref{fig:pddl_diagram} for an example. 

Predicates represent properties of the world and can be either \textbf{lifted} (parameterized) or \textbf{grounded} (instantiated). For example, the lifted predicate \texttt{(holding ?m - movable}\footnote{\texttt{movable} is an example of object type, used to restrict certain actions or predicates to a given object category.}) becomes grounded as \texttt{(holding bowl)} when the argument is instantiated. Only grounded predicates can be evaluated as true or false. $\mathcal{I}$, $\mathcal{G}$, and generally any state $s_t$ are specified as the set of true grounded predicates at timestep $t$, with the rest assumed to be false under the closed-world assumption.

Each action in $\mathcal{A}$ is defined by a set of \textbf{preconditions} over grounded predicates that must hold in the current state for the action to be applicable, and a set of \textbf{effects} that describe how grounded predicates truth values change after the action is executed. Actions, like predicates, can be lifted and then grounded by instantiating their parameters. Actions are typically executed by low-level controllers assumed to perfectly achieve their effects, following the \textbf{downward refinement property}~\citep{bacchus1991downward} and this allows planners to ignore intermediate states. However, real-world controllers may fail or face changing conditions, and as such grounding is required to identify these failures. 
    
Classical planning problems are typically described using the PDDL formal language~\citep{mcdermott1998pddl}. A PDDL domain file defines the space of available predicates and actions (along with object types), while a PDDL problem file defines the initial and goal states (in terms of grounded predicate truth values) along with available objects (i.e., specific instances of object types). Example PDDL domain and problem files are provided in Appendix~\ref{app:domains}.

We use the term \textbf{problem} to refer to a specific PDDL problem file (i.e., $\langle \mathcal{I}, \mathcal{G} \rangle$ plus object list), and \textbf{task} to refer to a concrete instance of that problem (e.g., a specific environment layout using the same domain). 
A planner takes a domain and problem file and returns a \textbf{plan}, which is a list of grounded actions that leads from $\mathcal{I}$ to $\mathcal{G}$.

Note that in classical planning, the planner operates with the symbolic state $s_t$, which is assumed to be given. We instead study a more realistic scenario, which we refer to as \textbf{visual planning}, where only the image $x_t$ of the environment at time $t$ is available, while $s_t$ must be extracted (e.g., with a VLM). This also implies that $\mathcal{I}$ is not known, but must be established from the first image $x_0$.

\section{Related Work}
\label{sec:related_work}

\paragraph{Plan Synthesis with VLMs.} 
A prominent line of research employs LLMs and VLMs as planners, prompted to directly select actions from a predefined list~\citep{ahn2022can, huang2023inner}. Several works integrate visual feedback through object detectors~\citep{song2023llm, chen2023llm, wu2023tidybot} or vision oracles~\citep{ren2023robots}, while others use VLMs to monitor plan execution and trigger replanning on failure~\citep{guo2024doremi, mei2024replanvlm}. Most relevant to one of our VLM-as-planner implementations, ViLa~\citep{hu2024look} uses a VLM in a closed-loop fashion by generating a new plan at each step. Crucially, all of these methods build systems upon foundation models without closely investigating the choice of the VLM itself or formalizing a benchmark.

\paragraph{Visual Symbolic Planning.}
The integration of LLMs and symbolic planning frameworks, such as PDDL, was first proposed by LLM+P~\citep{liu2023llm+}, with this line of work typically relying on the language model to generate a symbolic domain to be used with classical planners~\citep{huang2024understanding}. Several recent methods extend this to the visual setting, using VLMs to generate PDDL predicates and domains from images~\citep{liang2024visualpredicator, dang2025planning}, to learn predicates and actions through interaction~\citep{athalye2024predicate}, or to refine plans based on detected failures~\citep{duan2024aha}. VLM-TAMP~\citep{yang2024guiding} instead uses the VLM to provide a high-level plan that is then refined through classical motion planning.

Recently, VLMs have emerged as powerful tools for grounding classical planners with visual perception, which was previously only possible through domain-specific models~\citep{migimatsu2022grounding} trained end-to-end. S3E~\citep{azran2025s3e} enumerates all possible predicate-object combinations and converts them into natural language questions for a VLM at every step of the plan. The method however is not scalable and the work does not investigate the success rate of planning with the estimated states, which is hard to predict from prediction accuracy alone.

Most relevant to our work, TP-VQA~\citep{zhang2023grounding} and the later DKPROMPT~\citep{zhang2025dkprompt} use VLMs to monitor the truth values of predicates by asking specific questions based on an action's preconditions and effects.
DKPROMPT also directly compares its symbolic approach with a VLM-planner baseline on five OmniGibson tasks, reporting a large advantage for the symbolic approach. Importantly, however, the execution protocols differ: DKPROMPT updates its symbolic state and replans online when preconditions or effects are not satisfied, while its direct VLM-planner baseline is open-loop. Its evaluation is also limited to closed-source VLMs and a private set of five household tasks, with no publicly available implementation. ViPlan instead provides an open-source, multi-domain benchmark in which both paradigms operate under the same closed-loop protocol.
Our implementations of the VLM-as-grounder method class, while independently developed, resemble the DKPROMPT method, as asking yes-no questions about visual predicates is one of the most straightforward ways of interfacing a VLM and a PDDL planner.

\paragraph{Visual Planning Benchmarks.} Finally, while benchmarks like VisualAgentBench~\citep{liu2025visualagentbench} and EmbodiedBench~\citep{yang2025embodiedbench} evaluate VLM-based agents acting as planners across multiple interactive domains, they are not designed to support symbolic planners and thus cannot evaluate the VLM-as-grounder approach.
To the best of our knowledge, no open-source benchmark enables a comparison of VLM-as-planner and VLM-as-grounder methods in visual planning tasks under matched environments and protocols.

For further discussion on works related to benchmarking VLMs for hallucination detection and Visual Question Answering (VQA), we refer the reader to Appendix~\ref{app:additional_related_work}. 

\section{The ViPlan Benchmark}
\label{sec:benchmark}

This Section describes the two ViPlan domains, as well as the two classes of methods tested, together with the specific method instances. We also introduce the methodology for selecting the VLMs used for our experiments.

\subsection{Domains}
\label{subsec:domains}
\begin{figure}[t]
    \centering
    \begin{subfigure}[b]{0.59\linewidth}
        \centering
        \includegraphics[width=\linewidth]{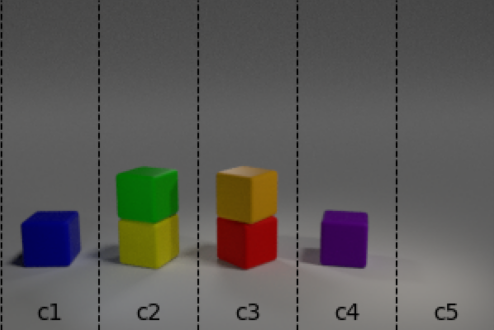}
        \caption{ViPlan-BW}
        \label{fig:blocks_example}
    \end{subfigure}
    \hfill
    \begin{subfigure}[b]{0.39\linewidth}
        \centering
        \includegraphics[width=\linewidth]{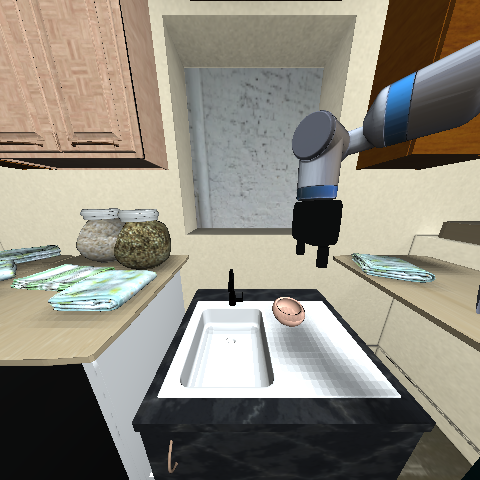}
        \caption{ViPlan-HH}
        \label{fig:house_example}
    \end{subfigure}

    \caption{\textbf{Example Domain Visualizations.} We show an example rendering for a state in ViPlan-BW (left) and ViPlan-HH (right).}
    \label{fig:side_by_side}
\end{figure}

ViPlan is comprised of two domains: \textbf{ViPlan-Blocksworld} (ViPlan-BW) and \textbf{ViPlan-Household} (ViPlan-HH), specifically chosen to represent both abstract fully observable, and realistic partially observable environments, which present opposing challenges. Example states for both domains can be seen in Figure~\ref{fig:side_by_side}. 

\paragraph{ViPlan-Blocksworld.}
Blocksworld is a popular classical planning domain~\citep{mcdermott1998ipc} in which the agent must stack blocks on a table in a specific configuration.
Our ViPlan-BW domain is written for the photorealistic Blocksworld renderer introduced by~\citet{asai2018photo}. To match the visual state, we use a simplified formulation with a single \texttt{moveBlock} action rather than the standard action set. Blocks can only be moved to the top of a column and can only be moved if they do not have other blocks on top. We consider blocks with identical sizes and shapes, but varying in color. ViPlan-BW is composed of 25 distinct procedurally generated problems for three difficulty splits: simple, medium and hard (see Appendix~\ref{app:domains_Blocksworld}).

\paragraph{ViPlan-Household.} 

While ViPlan-BW provides a controlled environment, we are also interested in evaluating planning in a domain closer to real-world embodied AI scenarios. For this, we implement ViPlan-HH on top of iGibson 2.0~\citep{li2022igibson}, specifically using the Fetch robot. iGibson is an open-source household simulator, where the agent can interact with various objects to complete tasks (e.g., setting the table or cleaning the dishes). 

As PDDL deals with high-level actions, we implement each one by bypassing low-level control and instead directly transitioning the environment to one of the action's valid terminal states, saving time and simulation cost. In order to introduce realistic variability, object positions are sampled within allowable regions, a process which also introduces occasional action failures that need to be detected. Due to this randomness, we verify that each task is solvable using an oracle symbolic planner which receives the full symbolic state from the simulator. As in ViPlan-BW, we divide ViPlan-HH in three splits with 25 tasks each. All our original contributions, implementation details, domain and task information can be found at Appendix~\ref{app:domains_iGibson}.

In ViPlan-HH, partial observability means that not all predicates are visible from the agent's current viewpoint, and so we assume the initial state $\mathcal{I}$ is partially known, with non-visible predicates initialized using privileged information from the simulator. This could alternatively be provided through other means in practical settings, such as natural language communication. We provide both method classes with equivalent privileged information about non-visible predicates throughout the episode, as detailed in Section~\ref{sec:methods}.

Furthermore, some ViPlan-HH tasks include multiple instances of the same object, which look identical and would be impossible to disambiguate. In such cases, in order to isolate legitimate errors from ambiguous object detection mistakes, we add labeled bounding boxes with object names (e.g., "book\_1", "book\_2") only to the duplicate objects. This disambiguates object references, but adds no relational information, thus the VLM must still reason about actions and object relationships.

\subsection{Visual Planning Methods} 
\label{sec:methods}

We consider two classes of methods, \textcolor{myblue}{\textbf{VLM-as-planner}} and \textcolor{myred}{\textbf{VLM-as-grounder}}, each implemented with four specific instances, to investigate design choices within each class.

\paragraph{\textcolor{myblue}{VLM-as-planner.}} At each timestep, the VLM is provided with an image of the current state, a textual description of the goal and the set of all available actions; additionally, in the case of ViPlan-HH, we provide natural language information about non-visible objects. Depending on the specific method instance, the VLM is then tasked with generating either a full plan in a parsable format (JSON) that satisfies the goal (\textcolor{myblue}{Plan} variant), or the next action (\textcolor{myblue}{Action} variant). We further consider two variants where the VLM is asked to perform a CoT reasoning trace before generating the plan or the action, named respectively \textcolor{myblue}{Plan + CoT} and \textcolor{myblue}{Action + CoT} variants.
In the case of the two Plan variants, only the first action of the plan is executed and the VLM is prompted to generate a full plan again at the next step. This kind of closed-loop planning is well-suited for our tasks, as it can naturally recover from action failures.

\begin{figure*}
    \includegraphics[width=\textwidth]{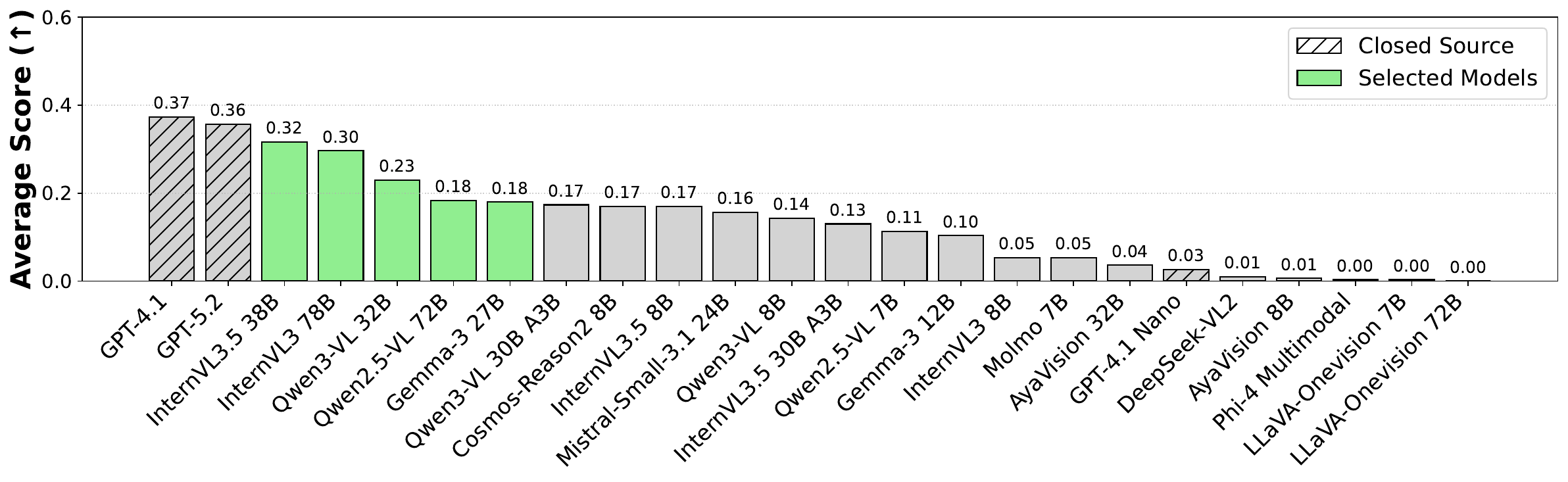}
    \caption{\textbf{Models selection.} We show a ranking of all tested VLMs, with an average score across both the \textcolor{myblue}{Plan} and \textcolor{myred}{Ground} methods, both the ViPlan-BW and ViPlan-HH domains, and all difficulty splits. The suite of models selected for our benchmark are highlighted in green. }
    \label{fig:models_preselection}
\end{figure*}

\paragraph{\textcolor{myred}{VLM-as-grounder.}} The VLM-as-grounder approach uses the VLM to ground a symbolic planner (specifically, the Fast Downward planner~\citep{helmert2006fast}, following Unified Planning's~\citep{unified_planning_softwarex2025} implementation). 

First, the VLM is tasked with filling in the starting truth values for all possible grounded predicates in the initial state $\mathcal{I}$ (which we call \textbf{predicate enumeration}), so that the symbolic planner can generate an initial plan. After, for every action in the plan to be executed, the VLM is first prompted to verify that the preconditions hold, and then to verify that all effects are observed as expected after execution, as shown in Figure~\ref{fig:figure1}.
We verify each predicate by first converting it to a natural language yes-no question through a pre-defined template (e.g., \texttt{(holding bowl)} becomes "Is the agent holding a bowl?") and then prompting the VLM to answer the question based on the current state image $x_t$. 
In ViPlan-HH, at each timestep we only query the VLM about predicates visible from the current viewpoint. Non-visible predicates are instead filled using privileged simulator information, or left unchanged from the previously grounded state when unaffected by the current action, providing the same state information that VLM-as-planner receives in natural language.
We consider two base variants: \textcolor{myred}{Ground}, which uses this standard yes-no question-answering, and \textcolor{myred}{Ground + CoT}, which augments each question with a request for CoT reasoning before the answer.

If, at any point, there is a mismatch between the VLM answers and the values expected by the PDDL action, the state of the world is deemed inconsistent. In this case, the action might have had unintended consequences, and changed other predicates besides the ones specified by its effects. For example, a bowl could be knocked over while moving another one due to a failed low-level execution. Thus, we must re-establish the state through predicate enumeration, generate a new plan, and continue the task.

\paragraph{Memory-augmented variants.} When the VLM determines that preconditions are not met, or when an attempted action is found non-executable, this signals a potential grounding error. We test variants that augment the VLM's context during the subsequent replanning cycle with \textbf{memory} of the issue: the action previously attempted, the precondition questions with their VLM answers, and whether the issue was caught during precondition checking or only discovered upon execution attempt. These memory-augmented variants are \textcolor{myred}{Ground + Mem} and \textcolor{myred}{Ground + Mem + CoT}, which combines memory with CoT reasoning.

\subsection{Vision-Language Model Selection}

\paragraph{VLM Pool.}
To ensure comprehensive coverage of the current VLM landscape, we evaluate a diverse pool of 21 open-source vision-language models spanning multiple model families and scales. Our model pool includes:
AyaVision (8B, 32B), Cosmos-Reason 2 (8B), DeepSeek-VL2 (27B A4.5B), Gemma-3 (12B, 27B), InternVL3 (8B, 78B), InternVL3.5 (8B, 38B, 30B A3B), Mistral-Small-3.1 (24B), Molmo (7B), Llava-OneVision (7B, 72B), Phi-4 Multimodal (5.6B), Qwen2.5-VL (7B, 72B) and Qwen3-VL (8B, 32B, 30B A3B)\footnote{For mixture-of-experts (MoE) models, we report parameter counts using the notation \{\emph{total}\}B A\{\emph{activated}\}B, where \emph{total} denotes the total number of parameters and \emph{activated} the number of parameters used per forward pass.}. Additionally, we include three closed-source models for reference: GPT-5.2, GPT-4.1 and GPT-4.1 Nano. We exclude these models from the selection process in order to keep the benchmark based on open-source models, but still test them to provide valuable context for assessing the gap between open and closed-source VLMs. We report the full model names as found in Hugging Face or in the APIs in Appendix~\ref{app:model_information}, while referring to their shorthands in all plots and tables for readability. For all models, we use a temperature value of 0 to ensure a deterministic outcome.

\paragraph{Model Selection Process.}
\label{sec:preselection}

Given the computational cost of evaluating 21 VLMs across all eight method variants and 150 tasks (75 per domain $\times$ 2 domains), we identify the most capable models through a selection procedure: we evaluate all 21 open-source models on two representative methods (Plan and Ground, the core variants without CoT or memory augmentations), compute each model's average success rate across both methods, both domains, and all three difficulty splits, rank by this aggregate score, and select the \textbf{top five open-source models} to form our evaluation suite for the main benchmark results. We weight Plan and Ground equally so that selection reflects competence across both planning paradigms. Empirically, this retains the strongest tested open-source grounders, the strongest open-source planner, and models competitive on both methods (Appendix~\ref{app:complete_benchmark_results}). This ensures our benchmark is easily reproducible, and new methods only need to test their performance across the selected models.

\paragraph{Selection Results.} 

Figure~\ref{fig:models_preselection} presents the ranking. We observe clear performance tiers, with larger models generally (though not universally) achieving higher scores, and a small gap between large closed-source and open-source models; notably, we find little difference between GPT-4.1 and GPT-5.2. The five best performing open-source models are: \textbf{InternVL3.5 38B} (0.32), \textbf{InternVL3 78B} (0.30), \textbf{Qwen3-VL 32B} (0.23), \textbf{Qwen2.5-VL 72B} (0.18), and \textbf{Gemma3-27B} (0.18). These VLMs, spanning 27B to 78B parameters across different architectures, form the basis for all subsequent benchmark evaluations reported in Section~\ref{sec:results}. 

\begin{table}[t]
\caption{\textbf{Results of methods (averaged over models).} We report the average success rates (mean across the selected VLMs) for each method (as defined in Section~\ref{sec:methods}), as well as averages across method classes; standard error of the mean across models is shown in parenthesis. The highest value in each column is bolded.}\label{tab:methods_results}
\resizebox{\linewidth}{!}{
\begin{tabular}{l c c c}
\toprule
\textbf{Method} & \textbf{ViPlan-BW} & \textbf{ViPlan-HH} & \textbf{Combined} \\
\midrule
\textcolor{myblue}{\textbf{Plan}}  & 0.07 \scriptsize{(0.03)} & 0.39 \scriptsize{(0.05)} & 0.23 \scriptsize{(0.02)} \\
\textcolor{myblue}{\textbf{Plan + CoT}} & 0.10 \scriptsize{(0.04)} & 0.23 \scriptsize{(0.04)} & 0.16 \scriptsize{(0.03)} \\
\textcolor{myblue}{\textbf{Action}} & 0.07 \scriptsize{(0.03)} & \textbf{0.41 \scriptsize{(0.03)}} & 0.24 \scriptsize{(0.01)} \\
\textcolor{myblue}{\textbf{Action + CoT}} & 0.14 \scriptsize{(0.04)} & 0.35 \scriptsize{(0.04)} & 0.25 \scriptsize{(0.03)} \\
\cmidrule(lr){0-0}
\textcolor{myblue}{\textbf{VLM-as-planner Avg.}} & 0.09 \scriptsize{(0.02)} & 0.34 \scriptsize{(0.02)} & 0.22 \scriptsize{(0.01)} \\
\midrule
\textcolor{myred}{\textbf{Ground}} & \textbf{0.47 \scriptsize{(0.14)}} & 0.04 \scriptsize{(0.01)} & 0.25 \scriptsize{(0.07)} \\
\textcolor{myred}{\textbf{Ground + CoT}} & 0.44 \scriptsize{(0.14)} & 0.06 \scriptsize{(0.02)} & 0.25 \scriptsize{(0.07)} \\
\textcolor{myred}{\textbf{Ground + Mem}} & \textbf{0.47 \scriptsize{(0.16)}} & 0.05 \scriptsize{(0.02)} & \textbf{0.26 \scriptsize{(0.08)}} \\
\textcolor{myred}{\textbf{Ground + Mem + CoT}} & 0.45 \scriptsize{(0.14)} & 0.05 \scriptsize{(0.01)} & 0.25 \scriptsize{(0.07)} \\
\cmidrule(lr){0-0}
\textcolor{myred}{\textbf{VLM-as-grounder Avg.}} & 0.46 \scriptsize{(0.07)} & 0.05 \scriptsize{(0.01)} & 0.25 \scriptsize{(0.04)} \\

\bottomrule
\end{tabular}
}
\end{table}

\begin{figure}[t]
    \centering
    \includegraphics[width=\linewidth]{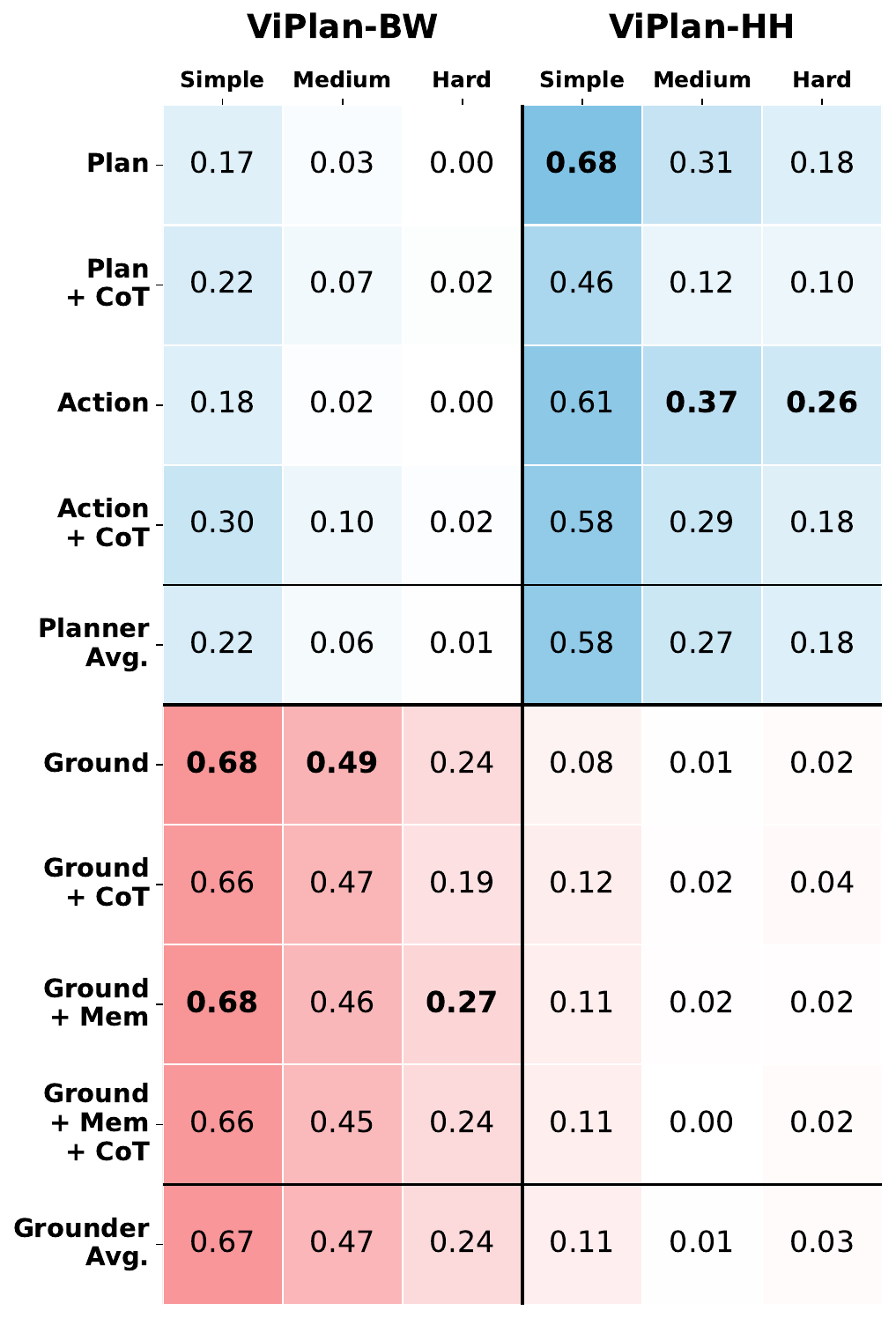}
    \caption{\textbf{Method performances across difficulty splits.} We show the average scores across models for each tested method, as well as the VLM-as-planner and VLM-as-grounder overall averages (rows), for each individual difficulty split of the domains (columns). Bolded values indicate the highest score for each column.}
    \label{fig:heatmap}
\end{figure}

\section{Results}
\label{sec:results}

We present our results, starting with the performances of VLM-as-planner and VLM-as-grounder variants over the two domains, then assessing the impact of CoT on the various methods and finally presenting a qualitative analysis of method failure cases.
A compact overview of the results across methods and domains is available in Table~\ref{tab:methods_results}, while Figure~\ref{fig:heatmap} shows a more detailed overview, separating scores across the three difficulty splits. For all methods, we report the \textbf{success rate}, i.e., the fraction of completed tasks, as a directly comparable measure of performance. 

\paragraph{\textcolor{myblue}{VLM-as-planner.}} We find VLM-as-planner performing unevenly across domains: it performs relatively well in ViPlan-HH (with the planner variants solving on average 34\% of tasks), while it struggles in ViPlan-BW, with the best method (Action + CoT) surprisingly solving only 14\% of tasks. A blank-image ablation with InternVL3.5 38B supports the hypothesis that non-visual task information contributes to this difference (Appendix~\ref{app:blank_image}). Removing visual state information reduces Plan success from 16\% to 4\% in ViPlan-BW, while preserving the 24\% average success rate in ViPlan-HH, even improving the simple split. This is consistent with direct VLM planning being able to leverage linguistic priors in the household domain, whereas ViPlan-BW depends more strongly on accurate perception and state tracking. For both domains, performance decreases uniformly when increasing the difficulty level (see Figure~\ref{fig:heatmap}).

\paragraph{\textcolor{myred}{VLM-as-grounder.}} Conversely, VLM-as-grounder has complementary strengths and weaknesses to VLM-as-planner approaches: it excels in ViPlan-BW, solving on average 46\% of the tasks, yet the performance drops drastically in ViPlan-HH, reaching at most a success rate of 6\%. Per-predicate results explain this difference (Appendix~\ref{app:predicate_accuracy_results}): in ViPlan-BW, stronger models achieve near-perfect accuracy on predicates such as \texttt{leftOf}, \texttt{rightOf}, and \texttt{on}, although \texttt{clear} remains harder. In ViPlan-HH, predicates such as \texttt{nextto}, \texttt{open}, and \texttt{reachable} remain below 90\% accuracy even for the strongest models. These local errors compound through the symbolic planner: an episode can require up to 120 predicate predictions, and a single incorrect value can corrupt the symbolic state and prevent task completion (Figure~\ref{fig:predictions_solved}). Similarly to VLM-as-planner methods, we find the performance to decrease with higher split difficulties.

\paragraph{Overall Comparison.}
Aggregated across both domains, no method or method class clearly outperforms the others (22\% for the VLM-as-planner average against 25\% for VLM-as-grounder, see Table~\ref{tab:methods_results}), and performance is far from saturated. Our findings rather reveal two complementary frameworks, with implementation details within the classes being less important. The two paradigms instead exhibit complementary weaknesses: direct VLM planning struggles when precise visual information is required and linguistic priors do not help, while symbolic grounding struggles with local perceptual errors in ambiguous visual states that compound over the course of a task. For reference, GPT-4.1 and GPT-5.2 show the same domain-dependent effect as the selected open-source models, favoring Ground in ViPlan-BW and Plan in ViPlan-HH; full closed-source results are reported in Appendix~\ref{app:complete_benchmark_results}.

\paragraph{Impact of Chain-of-Thought.}

Examining the impact of CoT more closely (Table~\ref{tab:methods_results}), we observe that adding CoT prompting yields no consistent improvement for most methods (Action, Ground, and Ground+Mem). Surprisingly, it actively hurts performance in the Plan method for the ViPlan-HH domain, where performance drops from 39\% (Plan) to 23\% (Plan+CoT), while it slightly benefits the same method for the ViPlan-BW domain.

This is robust to the generation token budget (Appendix~\ref{app:cot_budget}). Only Qwen3-VL 32B exceeds the budget among the selected models, and the comparison holds even when excluding it. Increasing its budget from 1024 to 4096 tokens reduces truncation, but raises the five-model ViPlan-HH average only from 23\% to 26\%, still below the 39\% obtained without CoT. Thus, truncation can affect reasoning-heavy models, but does not explain the overall lack of a consistent benefit from CoT.

We conclude that when the task requires more complex reasoning (reflecting on a multi-step plan, versus selecting a single action or answering a yes-no question) or the prompt is longer (as in ViPlan-HH versus ViPlan-BW), current VLMs struggle to maintain coherent reasoning chains during CoT, which we find to be in line with recent findings~\citep{chen2024measuring, shiri2024empirical}.

\paragraph{Qualitative Failure Analysis.}
We manually inspect several unsuccessful planning attempts to identify the most common failure modes; we report a detailed analysis with illustrative examples in Appendix~\ref{app:qualit_analysis}.

For VLM-as-planner methods, agents typically fail either by repeatedly predicting non-executable actions or by failing to follow the required output format (parsing errors). The first mode is more prevalent in ViPlan-BW than ViPlan-HH, in line with the relative performance of direct VLM planning in these domains. The second depends mainly on the model and affects both domains equally; we find the most recent models in our evaluation suite, Qwen3-VL 32B and InternVL3.5 38B, to be particularly susceptible, echoing concerns that fine-tuning on reasoning traces can lead to counterproductive overthinking~\citep{cuadron2025danger}.

For VLM-as-grounder methods, failures are primarily driven by incorrect grounding of action preconditions, with asymmetric consequences: incorrectly invalidated preconditions lead to dead-end symbolic states, while mistakenly validated ones cause repeated illegal action attempts. The use of memory in the VLM's context (Ground + Mem variants) does not significantly mitigate this. In ViPlan-BW, such errors stem from clear model mistakes, whereas in ViPlan-HH they are exacerbated by partial observability, viewpoint-dependent ambiguity, and the need to verify a larger number of interacting predicates.

\section{Conclusions}
\label{sec:conclusions}

In this work, we introduced ViPlan, the first open-source benchmark for comparing VLM-grounded symbolic approaches (VLM-as-grounder) with direct VLM planning methods (VLM-as-planner). ViPlan tests eight method variants from these two classes, across two interactive domains with complementary perceptual and reasoning demands.
We find that VLM-as-planner methods are more effective when they can leverage linguistic priors, whereas VLM-as-grounder methods are better suited to tasks that require precise state grounding and explicit symbolic structure. These differences are further supported by our qualitative analysis, which sheds light on concrete failure cases that each method should tackle to improve its performance, and further supported evidence is provided by a blank-image ablation and per-predicate results.
Furthermore, we show that Chain-of-Thought prompting provides little to no improvement across the majority of models and approaches, highlighting the ongoing difficulties with current VLMs and visual reasoning. 

Overall, our benchmark presents itself as an unsolved challenge, calling for novel robust planning methods capable of both classic planning rigor and handling ambiguous situations. 

\section*{Limitations}

In terms of method classes, we assume that in ViPlan-HH some privileged information is given to the VLM both when serving as a planner and as a grounder. This is because a scenario where the model has no information about where certain objects can be found would require a big amount of exploration. It is outside the scope of this work to determine how to avoid using privileged information, but future directions should investigate how to lift this assumption, especially in the case of VLM-as-grounder, where PDDL is not easily extendable to unobserved predicates otherwise. We assume a hierarchical control interface where low-level controllers handle action execution and termination. Consequently, if an agent attempts an action whose preconditions are not met in the ground truth state, the execution is rejected, the state remains unchanged, and the agent receives feedback to trigger replanning; this aligns with standard TAMP architectures where infeasible refinements cause backtracking. Furthermore, we assume the low-level controller signals when an action has terminated, tasking the VLM solely with verifying the high-level effects or selecting the next step, although recent work suggests VLMs could also assist in detecting termination conditions~\citep{du2023vision}.

More in general, ViPlan-HH presents non-negligible stochasticity which could lead to fluctuations across simulator seeds. Our reported errors capture variability across the 25 problems in each split, rather than repeated-seed uncertainty: our computational budget and the large number of models tested (see Appendix~\ref{app:resources}), prevented running multiple seeds per experiment. Thus, small differences between methods within the same paradigm should not be interpreted as robust rankings, and future experiments with the benchmark would benefit from multi-seed evaluation.
We also assume, as is typical in the classical planning literature, that the domain $\mathcal{D}$ is given. However recent works, such as \citet{guan2023leveraging}, explore how the domain could be generated using LLMs. We also relied on templates to translate grounded predicates into questions. Recent work \citep{azran2025s3e} has shown how this can be automated with LLMs, but as the scope of our benchmark is evaluating VLMs, we avoided this for simplicity.

Finally, since VLM selection uses the benchmark tasks themselves, performance of the selected models may be optimistic as an estimate of model-level generalization; our selection is intended to identify a capable evaluation suite rather than to rank individual VLMs, and our results depend on comparisons and not absolute metrics.

 \paragraph{Potential Risks.} 
 ViPlan evaluates VLMs in simulated planning settings and does not involve real-world deployment. A possible indirect risk is that strong benchmark results could be over-interpreted as evidence that current VLM-based planners are ready for safety-critical embodied deployment; our findings (success rates well below saturation, compounding grounding errors, and CoT failing to consistently help) argue against such an interpretation. We release ViPlan to encourage rigorous evaluation rather than to endorse any particular system for real-world use.

\section*{Acknowledgments}
This work was supported by the Research Council of Finland (Flagship programme: Finnish Center for Artificial Intelligence FCAI, and grants 352986, 358246), EU (H2020 grant 101016775 and NextGenerationEU) and by the PNRR project FAIR - Future AI Research (PE00000013), under the NRRP MUR
program funded by NextGenerationEU. We thank Aalto-IT (IT Services of Aalto University, Finland) for the provided support with computational resources. 
Giovanni Bonetta and Bernardo Magnini were partially supported by the PNRR MUR project \href{https://fondazione-fair.it/}{PE0000013-FAIR} (Spoke 2).

\bibliography{custom}

\input{appendix}

\end{document}

%% file: appendix.tex
\setcounter{footnote}{0} 

\appendix
\clearpage

\section{Failure Cases Analysis}
\label{app:qualit_analysis}
In this Section, we present a series of representative failure cases of both classes of methods across the two ViPlan domains that complement our analysis in the main paper. The main failure cases for VLM-as-planner and VLM-as-grounder can happen regardless of whether CoT is employed. We report cases accompanied by CoTs (edited for compactness and clarity), as we find them insightful. We additionally present one failure case specific of VLM-as-planner methods that employ CoT.

\subsection{VLM-as-planner.}
VLM-as-planner variants failure cases can be divided in actions that are parsable but non-executable and actions that fail to be parsed, which we present in this order.

\begin{figure*}
    \centering
    \includegraphics[width=\textwidth]{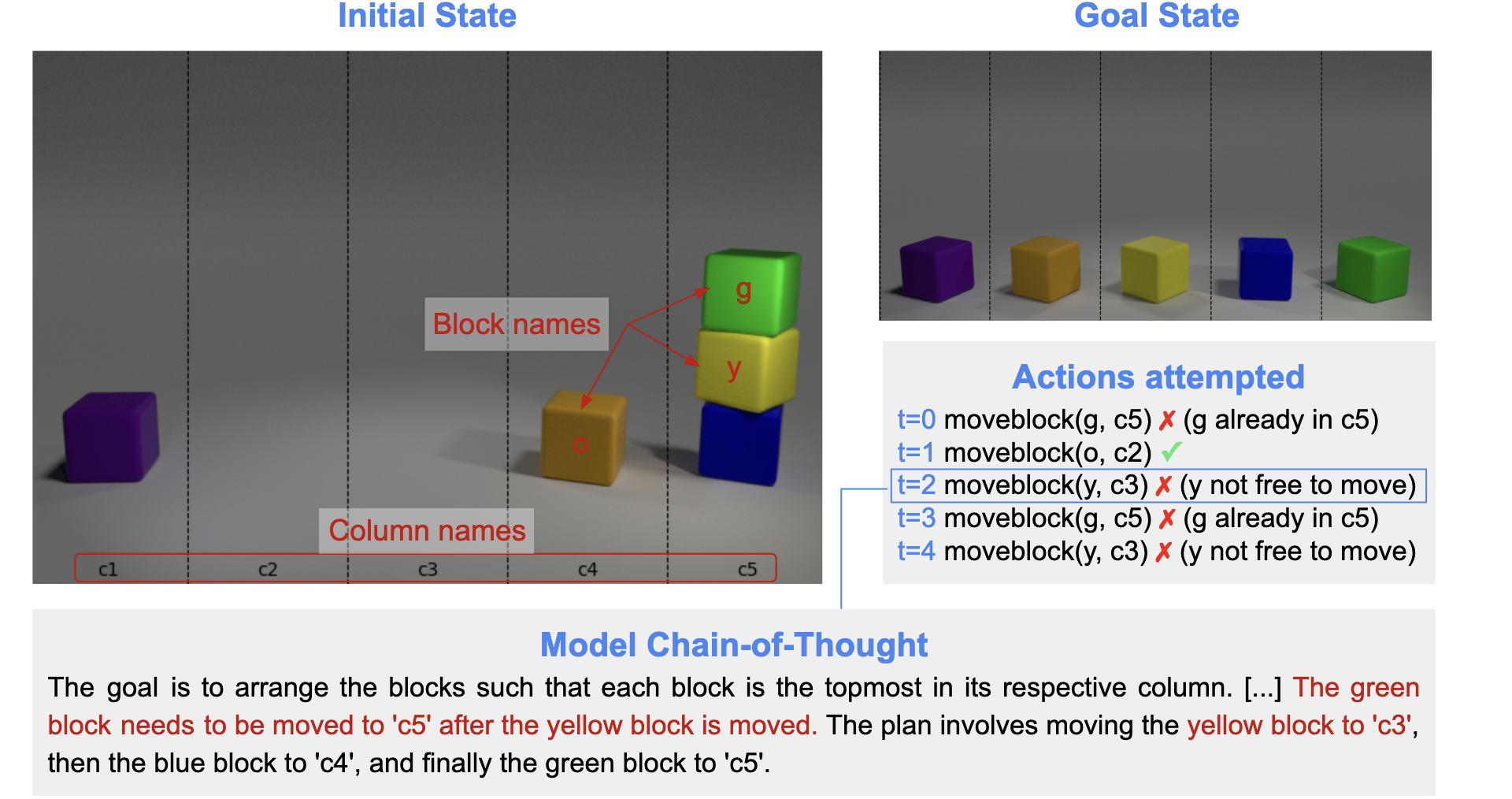}
    \caption{\textbf{Failure case of a VLM-as-planner method in ViPlan-BW.}
The agent oscillates between multiple non-executable actions due to inconsistent implicit state tracking. After one successful move, the model repeatedly attempts to move blocks that are either not free to move, or already in the target column, violating action preconditions and failing to make progress toward the goal.}
    \label{fig:vlm_planner_failure_bw}
\end{figure*}

\begin{figure*}
    \centering
    \includegraphics[width=0.8\textwidth]{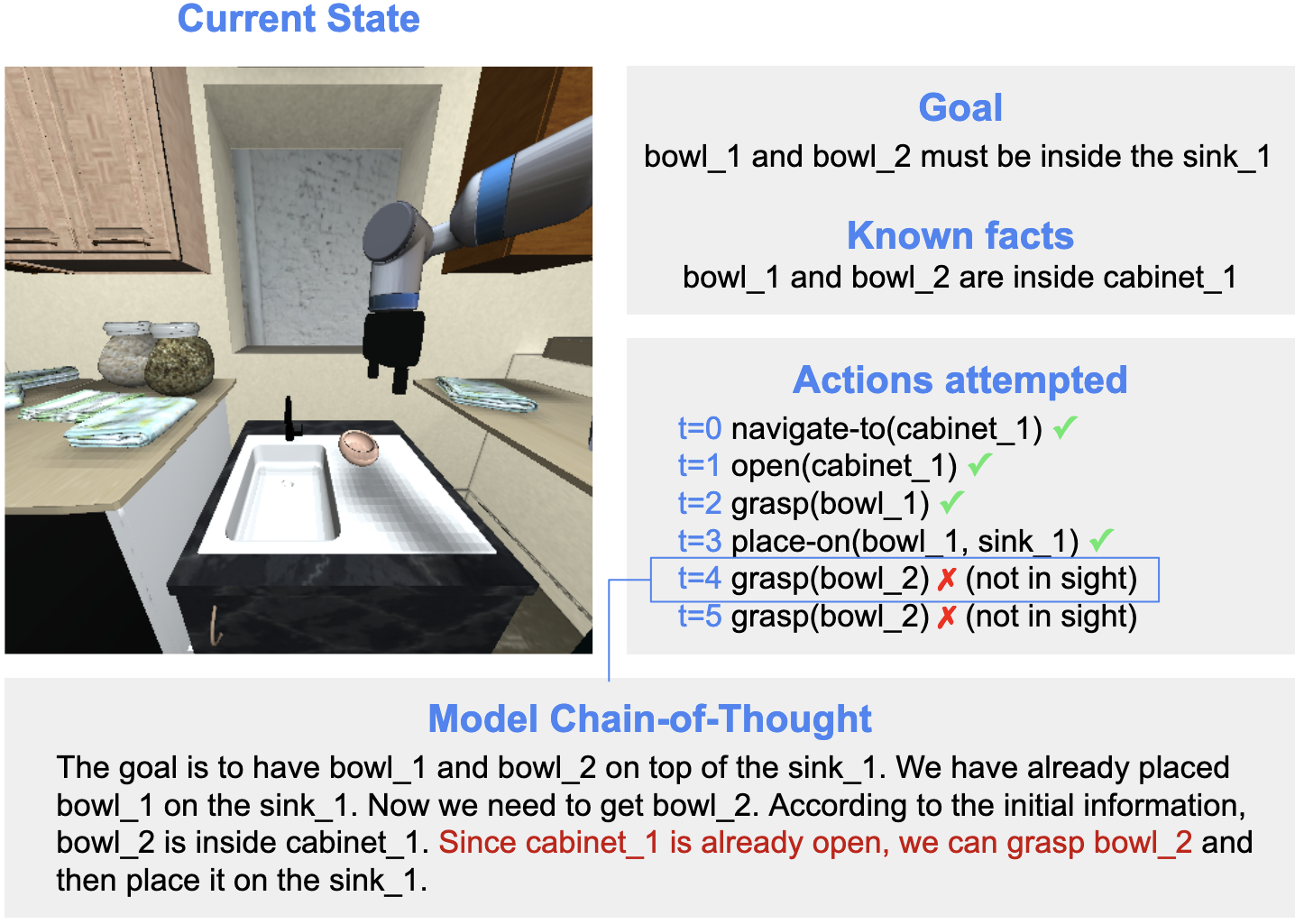}
    \caption{\textbf{Failure case of a VLM-as-planner method in ViPlan-HH.} The agent attempts a non-executable action. After the first attempt, it fails to diagnose the cause for the failure and enters a loop of re-attempts.
}
    \label{fig:vlm_planner_failure_hh}
\end{figure*}

\paragraph{Non-executable action failures.} We observe two main dynamics that arise from a single failed action and propagate into a sequence of failures until the termination of an episode: 
\begin{enumerate}
    \item The agent oscillates between two non-executable actions, as shown in Figure~\ref{fig:vlm_planner_failure_bw}.
    \item The same action is repeated again and again, as possible to see in Figure~\ref{fig:vlm_planner_failure_hh}.
\end{enumerate}
In both cases, a mistaken diagnosis about why the action is non-executable is responsible for these failing dynamics. It is possible that moving to more advanced VLM agents with better memory and replanning procedures could solve some of these failures.

\begin{figure*}
    \centering
    \includegraphics[width=0.8\textwidth]{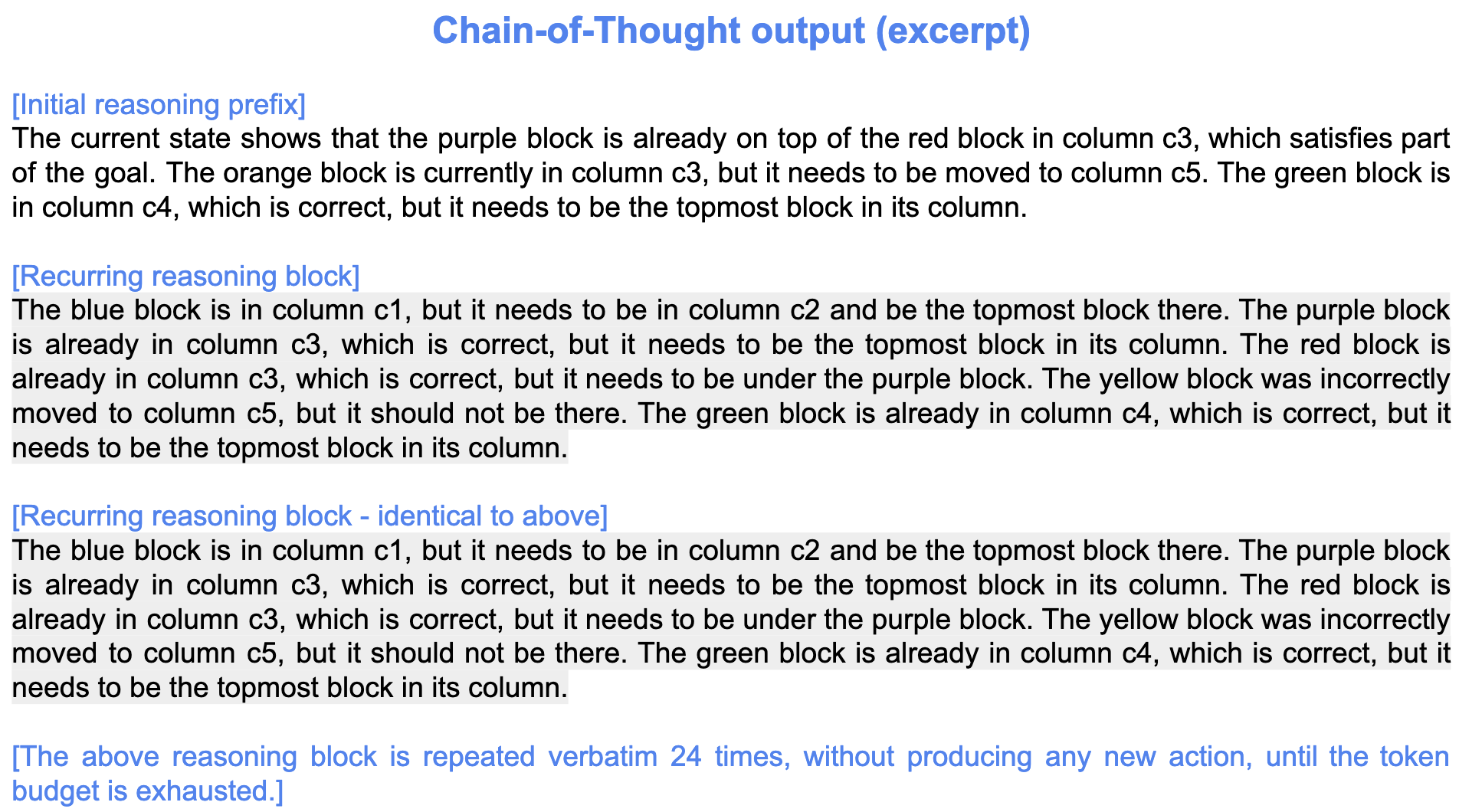}
    \caption{\textbf{Failure case illustrating the negative impact of Chain-of-Thought prompting in ViPlan-BW.} The VLM initially produces a coherent reasoning prefix, but its reasoning then converges to a recurring reasoning block that is repeated verbatim across subsequent steps, preventing the generation of new actions and exhausting the token budget.}
    \label{fig:cot_loop}
\end{figure*}

\paragraph{Non-parsable action failures.} As we require all model outputs to be in JSON format, all parsing errors are due to incorrect formatting and we are not aware of any shortcomings of our parsing procedure that would discard otherwise valid outputs. Ultimately, outputs with the wrong format can be re-conducted to the inability of a model to follow instructions. We inspect one particularly interesting case of parsing error (see Figure~\ref{fig:cot_loop}), caused by the model CoT repeating itself in loops until it exhausts the token budget without producing any valid JSON output. This case is observed especially in the Plan+CoT variant of VLM-as-planner, as it tends to arise from the interaction between the request for a plan in the prompt and the usage of CoT.

\subsection{VLM-as-grounder.} 
\begin{figure*}
    \centering
    \includegraphics[width=\textwidth]{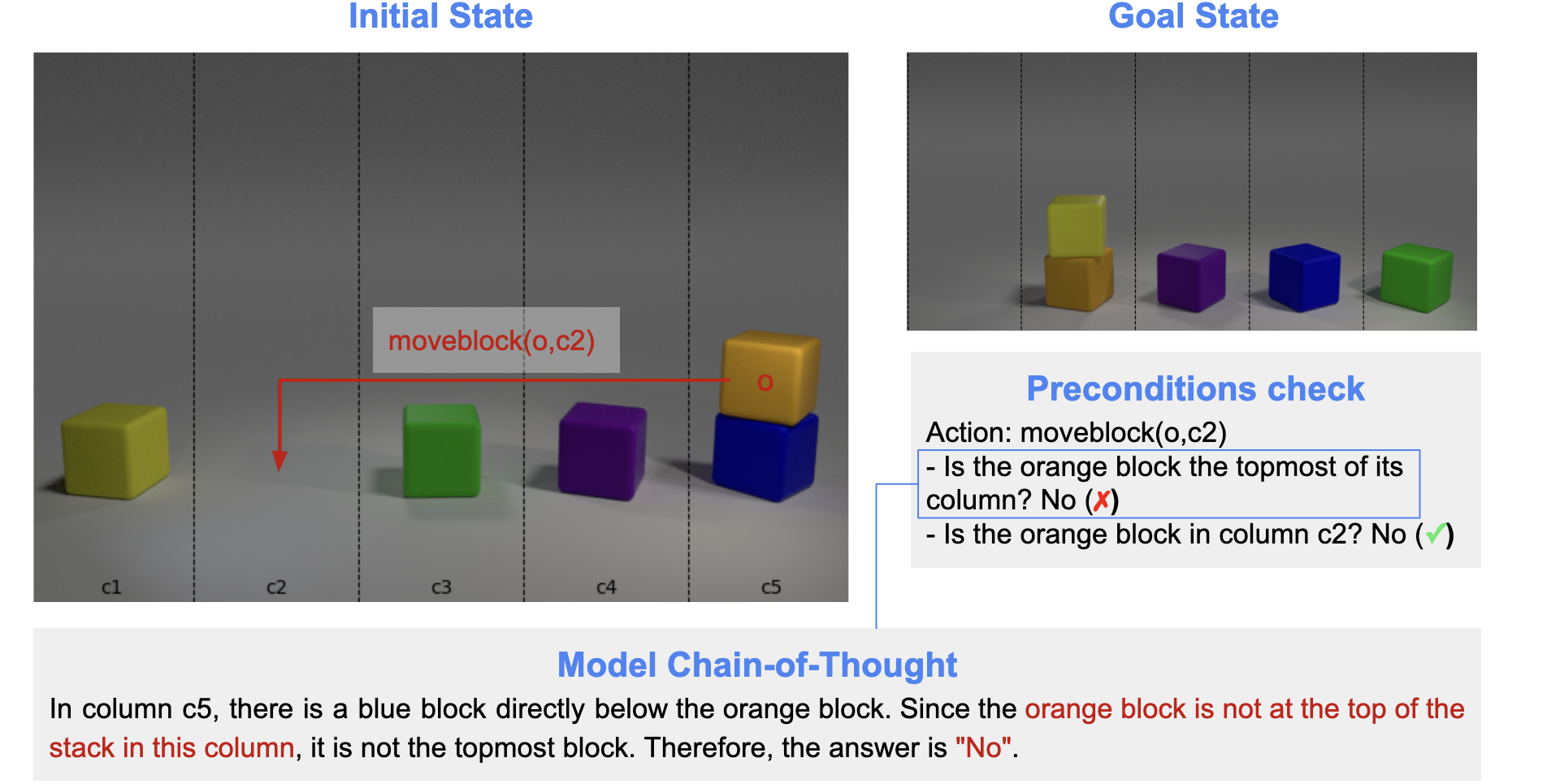}
    \caption{\textbf{Failure case of a VLM-as-grounder method in ViPlan-BW.} During precondition grounding, the VLM incorrectly judges a required precondition of the action \texttt{moveblock(o, c2)} as unsatisfied, stating that the orange block is not topmost in its column when it actually is. As a result, the symbolic planner rejects an otherwise valid action, leading to a dead-end state in which no valid plan can be generated and the episode terminates unsuccessfully.}
    \label{fig:vlm_grounder_failure_bw}
\end{figure*}

\begin{figure*}
    \centering
    \includegraphics[width=0.8\textwidth]{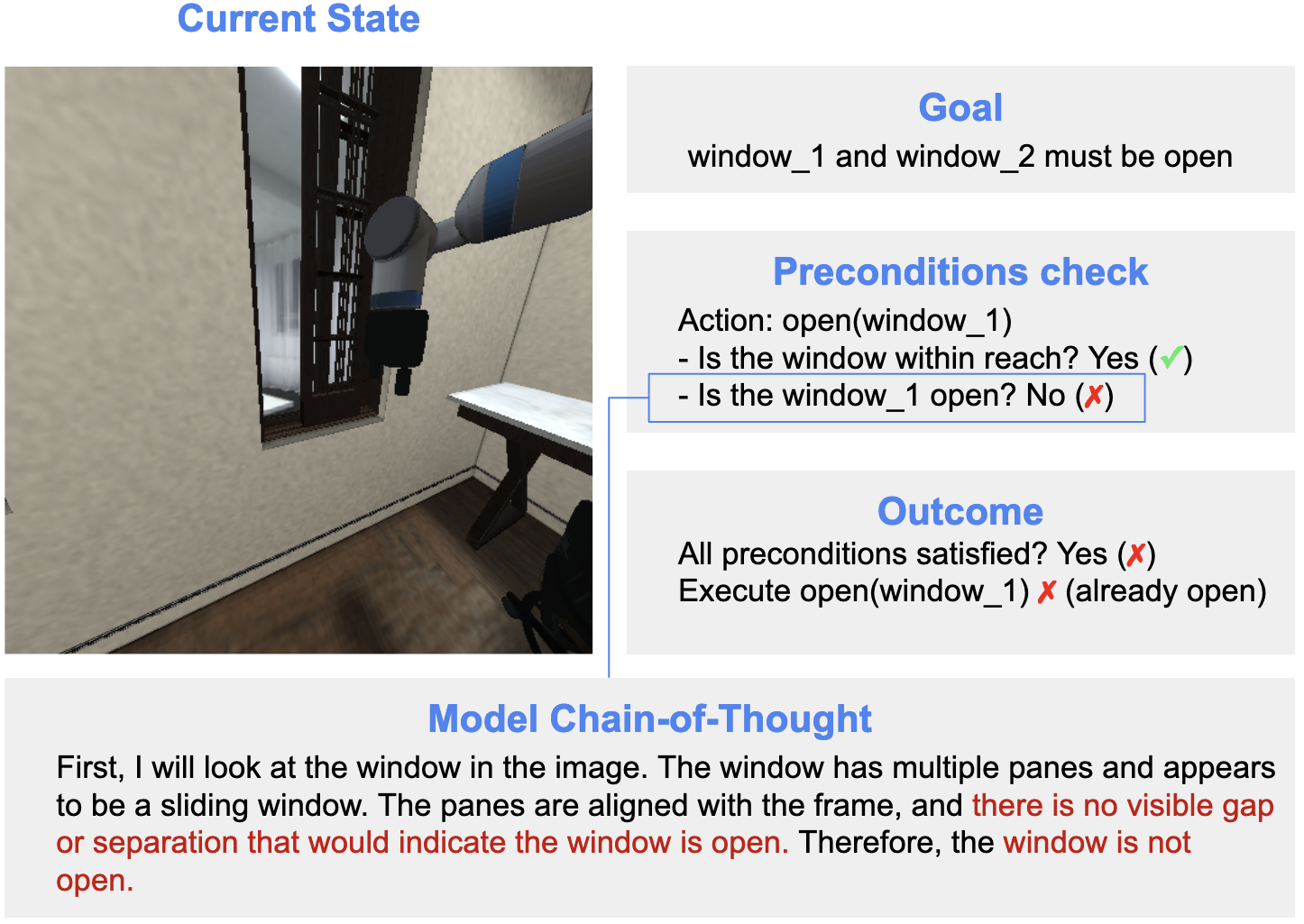}
    \caption{\textbf{Failure case of a VLM-as-grounder method in ViPlan-HH.} During precondition grounding, the VLM incorrectly judges a required precondition of the action \texttt{open(window\_1)} as satisfied, stating that the window is closed when it is already open. Consequently, the symbolic planner attempts an invalid action that cannot be executed in the environment. If uncorrected, this error can lead to repeated invalid action attempts and potential infinite loops.}
    \label{fig:vlm_grounder_failure_hh}
\end{figure*}

In VLM-as-grounder, there can be errors in assessing the preconditions or the effects of an action. We observe that episodes become irrecoverable for symbolic planners when preconditions are misjudged, entering one of two states:
\begin{enumerate}
    \item A precondition is erroneously judged as unsatisfied, thus preventing a valid action that is necessary to achieve the goal (see Figure~\ref{fig:vlm_grounder_failure_bw}). The precondition then remains unsatisfied also at future steps, leading to a dead-end state.
    \item A precondition is erroneously judged as satisfied, thus attempting an invalid action (see Figure~\ref{fig:vlm_grounder_failure_hh}). The precondition then remains satisfied also at future steps, leading to more attempts and potential infinite loops.
\end{enumerate}
Notice that an action can only be chosen if the estimated symbolic state satisfies the preconditions. Thus to enter the second case, some predicate must be wrong.

Errors in checking the effects of actions manifest in wrong assumptions about preconditions in the future steps and can thus lead to the second case above.

\section{Additional Related Work}
\label{app:additional_related_work}
The following is an expansion on Related Work (Section~3 of the main paper) with literature referring to Benchmarks for Vision-Language Models that evaluate their planning abilities and robustness to hallucinations.

\paragraph{Benchmarks for Vision-Language Models.} \label{sec:bench_for_VLM}

Our work closely aligns with other benchmarks, specifically in the area of planning, object hallucinations and relationship hallucinations with VLMs. For planning, previous research typically framed the problem as Visual Question Answering (VQA) \citep{antol2015vqa}, asking questions about which actions to take \citep{chen2023egoplan, majumdar2024openeqa}. Visual Spatial Planning (VSP) \citep{wu2024vsp} specifically focuses on spatial relationships, but without any notion of symbolic predicates. Our work is closer to PlanBench \citep{valmeekam2023planbench}, which investigates the combination of LLMs with symbolic planning, but without considering the visual component.

Hallucinations in foundation models are a widely studied phenomenon \citep{liu2024survey}. POPE \citep{li2023evaluating} is an established benchmark for object hallucination in VLMs, with adversarial examples constructed out of frequently co-occurring objects. This kind of evaluation was scaled up by DASH \citep{augustin2025dash} through smart retrieval from a bigger dataset. MMVP \citep{tong2024eyes} investigates both object and relationship hallucinations by looking at \emph{CLIP-blind} pairs of images, placing a large emphasis on the role of the vision encoder \citep{radford2021learning}. Benchmarks for relationship hallucination typically work in a similar way, by presenting pairs of images with adversarial captions \citep{thrush2022winoground, nikolaus2022vision}, with later advancements also focusing on the particular role of the vision encoder \citep{zeng2024investigating}.

\section{Model Information}
\label{app:model_information}

For clarity and readability, we report simplified model names throughout the main text and figures. In Table~\ref{tab:model-names} are the correspondance between the full names of the models and the simplified versions. For our experiments, GPT-5.2, GPT-4.1 and GPT-4.1 Nano were accessed via the OpenAI API\footnote{\url{https://openai.com/index/openai-api/}}, while all other models were downloaded from the Hugging Face Hub\footnote{\url{https://huggingface.co/docs/hub/index}}.

\begin{table*}[h!]
\centering
\caption{\textbf{Simplified Model Names.} Mapping between specific model identifiers and simplified model names used in the paper.}
\label{tab:model-names}
\resizebox{\linewidth}{!}{
\begin{tabular}{lll}
\toprule
\textbf{Identifier} & \textbf{Simplified Name} & \textbf{Reference}\\
\midrule
\texttt{CohereLabs/aya-vision-8b} & AyaVision 8B & \citep{cohere2025aya}\\
\texttt{CohereLabs/aya-vision-32b} & AyaVision 32B & \citep{cohere2025aya}\\
\texttt{nvidia/Cosmos-Reason2-8B} & Cosmos-Reason2 8B & \citep{cosmos_reason2_github} \\
\texttt{deepseek-ai/deepseek-vl2} & DeepSeek-VL2 & \citep{wu2024deepseek}\\
\texttt{gpt-5.2-2025-12-11} & GPT-5.2 & \citep{openai2025gpt5.2} \\
\texttt{gpt-4.1-nano-2025-04-14} & GPT-4.1 Nano & \citep{openai2025gpt4.1}\\
\texttt{gpt-4.1-2025-04-14} & GPT-4.1 & \citep{openai2025gpt4.1}\\
\texttt{google/gemma-3-12b-it} & Gemma-3 12B & \citep{team2025gemma}\\
\texttt{google/gemma-3-27b-it} & Gemma-3 27B & \citep{team2025gemma}\\
\texttt{OpenGVLab/InternVL3-8B} & InternVL3 8B & \citep{zhu2025internvl3}\\
\texttt{OpenGVLab/InternVL3-78B} & InternVL3 78B & \citep{zhu2025internvl3}\\
\texttt{OpenGVLab/InternVL3\_5-8B-HF} & InternVL3.5 8B & \citep{wang2025internvl35} \\
\texttt{OpenGVLab/InternVL3\_5-38B-HF} & InternVL3.5 38B & \citep{wang2025internvl35} \\
\texttt{OpenGVLab/InternVL3\_5-30B-A3B-HF} & InternVL3.5 30B A3B & \citep{wang2025internvl35} \\
\texttt{lmms-lab/llava-onevision-qwen2-7b-ov-hf} & LLaVA-Onevision 7B & \citep{li2025llavaonevision}\\
\texttt{lmms-lab/llava-onevision-qwen2-72b-ov-hf} & LLaVA-Onevision 72B & \citep{li2025llavaonevision}\\
\texttt{mistralai/Mistral-Small-3.1-24B-Instruct-2503} & Mistral-Small-3.1 24B & \citep{mistral2025small}\\
\texttt{allenai/Molmo-7B-D-0924} & Molmo 7B & \citep{deitke2024molmo}\\
\texttt{microsoft/Phi-4-multimodal-instruct} & Phi-4 Multimodal & \citep{abouelenin2025phi}\\
\texttt{Qwen/Qwen2.5-VL-7B-Instruct} & Qwen2.5-VL 7B & \citep{bai2025qwen2}\\
\texttt{Qwen/Qwen2.5-VL-72B-Instruct} & Qwen2.5-VL 72B & \citep{bai2025qwen2}\\
\texttt{Qwen/Qwen3-VL-8B-Instruct} & Qwen3-VL 8B & \citep{bai2025qwen3vltechnicalreport} \\
\texttt{Qwen/Qwen3-VL-32B-Instruct} & Qwen3-VL 32B & \citep{bai2025qwen3vltechnicalreport} \\
\texttt{Qwen/Qwen3-VL-30B-A3B-Instruct} & Qwen3-VL 30B A3B & \citep{bai2025qwen3vltechnicalreport} \\
\bottomrule
\end{tabular}}
\end{table*} 

\paragraph{Licensing.}

All models, environments, and software used in this work (Qwen-VL, InternVL, Gemma3, BEHAVIOR-1K, Fast Downward, Unified Planning, etc.) are released under permissive open-source licenses (Apache 2.0, MIT, or similar) that allow research use. Our use is consistent with their intended use. We release ViPlan under the MIT License.

\section{Error Computations}
\label{app:errors}

In this Section we outline how the errors are computed throughout the paper. 

To estimate the error of the mean on the success rate on a split, we assume each problem being a Bernoulli trial (i.e., drawn from a binomial distribution) with true success probability $p$. Our empirical estimate of $p$, denoted as $\hat{p}$, is used to estimate the standard error of the mean (SEM) as
\begin{equation}
\label{eq:binomial_error}
    \text{SEM}(\hat{p}) \approx \sqrt{(\hat{p}(1-\hat{p})/n}~,
\end{equation}
where $n$ is the number of problems in the split. This normal approximation to the binomial distribution is reasonable for $n=25$, which holds for all splits in our experiments.

In the Tables we further report the error of the average success rates across difficulty splits and occasionally even combined across the two domains of ViPlan. These are computed using standard error propagation for the average of independent estimates:
\begin{align}
    \text{SEM}(\bar{x}) &= \text{SEM}\left(\sum_{i=1}^m x_i/m \right)  \\&= \frac{1}{m}\sqrt{\sum_i \text{SEM}(x_i)^2} ~.
\end{align}

\section{Statistical Significance of Chain-of-Thought}
\label{app:statistics_cot}

We report in Table~\ref{tab:cot_errors} the numerical differences between the average success rate with and without CoT prompting, together with their errors. As a descriptive measure of effect magnitude relative to uncertainty, we also report the ratio between the absolute average difference and its SEM. Except for Plan in the medium split of both domains, there is no significant difference (measured in 3 standard deviations of the mean from 0) in employing CoT or not, and even in this case, the difference is positive in ViPlan-BW, but negative in ViPlan-HH. We conclude that there is no evidence that CoT helps in our benchmark when averaging across models.

\begin{table*}[h]
\centering
\caption{\textbf{Statistical significance of CoT difference.} We report the average difference between experiments with and without CoT prompting across models for each split and the Ground and Plan methods, using all models tested during pre-evaluation. We further report the standard error of the mean, and the ratio between the absolute value of the average and its error in parenthesis. We bold ratios below 3, indicating differences that lie within three SEMs of zero.}
\label{tab:cot_errors}
\begin{tabular}{l l c c c c c c}
\toprule
\multirow{2}{*}{\textbf{Method}} & \multirow{2}{*}{\textbf{Domain}} & \multicolumn{2}{c}{\textbf{Simple}} & \multicolumn{2}{c}{\textbf{Medium}} & \multicolumn{2}{c}{\textbf{Hard}} \\
 &  & \textbf{Average} & \textbf{Error} & \textbf{Average} & \textbf{Error} & \textbf{Average} & \textbf{Error} \\
\midrule
\multirow{2}{*}{\textbf{Ground}} & ViPlan-BW & -0.060 & 0.022 \scriptsize{(\textbf{2.8})} & -0.018 & 0.015 \scriptsize{(\textbf{1.2})} & -0.005 & 0.013 \scriptsize{(\textbf{0.4})} \\
 & ViPlan-HH & 0.010 & 0.016 \scriptsize{(\textbf{0.6})} & 0.000 & 0.005 \scriptsize{(\textbf{0.0})} & -0.003 & 0.005 \scriptsize{(\textbf{0.5})} \\
\midrule
\multirow{2}{*}{\textbf{Plan}} & ViPlan-BW & 0.052 & 0.018 \scriptsize{(\textbf{3.0})} & 0.033 & 0.009 \scriptsize{(3.6)} & 0.007 & 0.005 \scriptsize{(\textbf{1.4})} \\
 & ViPlan-HH & -0.068 & 0.024 \scriptsize{(\textbf{2.8})} & -0.065 & 0.018 \scriptsize{(3.5)} & -0.040 & 0.017 \scriptsize{(\textbf{2.4})} \\
\bottomrule
\end{tabular}
\end{table*}

\section{Chain-of-Thought Token Budget Analysis}
\label{app:cot_budget}

We use a generation budget of 1024 tokens for the main benchmark experiments. To assess whether this setting affects our conclusions about Chain-of-Thought prompting, we analyze output lengths and truncation across both domains and method classes. For VLM-as-grounder, none of the five selected models exhausts the token budget in either domain, with median CoT lengths of approximately 100 tokens in ViPlan-BW and 115 tokens in ViPlan-HH. For VLM-as-planner, four of the five selected models remain within the budget, with median CoT lengths of approximately 250--400 tokens. The exception is Qwen3-VL 32B, which can enter long repetitive reasoning chains and exhaust the budget.

We rerun Qwen3-VL 32B with Plan+CoT on ViPlan-HH using a 4096-token generation budget in order to mitigate the issue. Truncated outputs decrease from approximately 8\% to 2\%, while its task success rate increases from 20\% to 35\%. Substituting this result into the five-model aggregate raises Plan+CoT from 23\% to 26\%, which remains below the 39\% obtained by Plan without CoT. Therefore, increasing the generation budget improves a model-specific failure mode (as also shown in Appendix~\ref{app:qualit_analysis}, Figure~\ref{fig:cot_loop}) but does not change our conclusion that CoT provides no consistent improvement.

\section{Computational Resources}
\label{app:resources}
We ran all our experiments on a cluster equipped with three different GPU models from NVIDIA: A100 with 80GB VRAM, H100 with 80GB VRAM and H200 with 141GB VRAM.

The computational resources needed to run the full benchmark for 12 models fitting in a single 80GB GPU and 9 models fitting in two GPUs, totaling around 1800 GPU hours. While this is a considerable amount, for practitioners willing to benchmark a VLM of around 7-8B parameters, this would require roughly 39 hours for just the Ground and Plan methods (with and without CoT) and less than 15 if running only the experiments without Chain-of-Thought. Benchmarking a new method on the five selected models (assuming the GPU cost is similar to one of our CoT implementations) would instead cost around 100 GPU hours (assuming all selected models are ran on a double GPU setup, but this can be reduced if, for example, they are loaded on a single H200).

More in general, we estimate that all the experiments of the project required us around 4000 GPU hours, several of which employing smaller V100 GPUs to run the iGibson simulator.

Finally, the experiments with GPT-5.2, GPT-4.1 and GPT-4.1 Nano costed around 210\$.

\section{Domain Details}
\label{app:domains}

\subsection{ViPlan-BW}
\label{app:domains_Blocksworld}

We divided the tasks into three splits, simple, medium and hard, each composed of 25 different problems which were automatically generated and validated. The details for each split can be found in Table~\ref{tab:bw_task_splits}. 

\begin{table}[h]
  \caption{\textbf{ViPlan-BW Task Splits.} Task splits based on difficulty, number of blocks, columns, and plan length.}
  \label{tab:bw_task_splits}
  \centering
  \begin{tabular}{lccc}
    \toprule
    \textbf{Split} & \textbf{\# Blocks} & \textbf{\# Columns} & \textbf{Plan Length} \\
    \midrule
    Simple  & 3 & 4 & 3--5  \\
    Medium  & 5 & 5 & 5--10 \\
    Hard    & 6 & 4 & 8--15 \\
    \bottomrule
  \end{tabular}
\end{table}

While typical PDDL domains for Blocksworld also include a predicate for the agent holding a block in its hand, since our image observation space does not show the agent, we instead define a simplified version of the domain with only a single action that moves a block to the top of a specified column.

The full PDDL domain for ViPlan-BW can be seen in Figure~\ref{pddl:Blocksworld}. Note that the \texttt{rightOf} and \texttt{leftOf} predicates are never actually used in the \texttt{moveBlock} action; however they are still filled in by the VLM when enumerating all predicates, and can thus serve as a way of measuring the VLM's awareness of spatial directions such as left and right. The domain and problem files can be given to a classical planner to compute a plan.

\begin{figure*}
\centering
\begin{pddlcode}[title=ViPlan-BW Domain]
(define (domain Blocksworld)
  (:requirements :strips :typing :negative-preconditions :conditional-effects 
    :equality)
  (:types block column)

  (:predicates
    (on ?b1 - block ?b2 - block) ;; block b1 is on block b2
    (inColumn ?b - block ?c - column) ;; block b is in column c
    (clear ?b - block) ;; block b is clear (i.e., nothing is on top of it)
    (rightOf ?c1 - column ?c2 - column) ;; column c1 is to the right of column c2
    (leftOf ?c1 - column ?c2 - column) ;; column c1 is to the left of column c2
  )

  (:action moveBlock
    :parameters (?b1 - block ?c1 - column) ;; move block b1 to column c1
    :precondition (and (clear ?b1) (not (inColumn ?b1 ?c1))) ;; block b1 must be clear and not already in column c1
    :effect (and
      (forall
        (?b2 - block) ;; for all blocks b2
        (and
          (when
            (on ?b1 ?b2)
            (and (not (on ?b1 ?b2)) (clear ?b2)) ;; if block b1 was on block b2, then b1 is no longer on b2 and b2 is clear
          )
          (when
            (and (inColumn ?b2 ?c1) (clear ?b2) (not (= ?b2 ?b1)))
            (and (on ?b1 ?b2) (not (clear ?b2))) ;; if another block b2 was in the column c1 where b1 is moving and b2 was clear, then b1 is now on b2 and b2 is no longer clear
          )
        )
      )
      (forall
        (?c2 - column) ;; for all columns c2
        (when
          (inColumn ?b1 ?c2)
          (not (inColumn ?b1 ?c2))) ;; if block b1 was in column c2, then b1 is no longer in c2) 
      (inColumn ?b1 ?c1) ;; block b1 is now in column c1
      (clear ?b1) ;; block b1 is now clear (as it must be if it was moved)
    )
  )
)
\end{pddlcode}
\caption{\textbf{ViPlan-BW Domain.} PDDL domain for ViPlan-BW.}
\label{pddl:Blocksworld}
\end{figure*}

\begin{figure*}
\begin{pddlcode}[title=Example Problem for ViPlan-BW]
(define (problem simple_problem_0)
  (:domain Blocksworld)
  
  (:objects 
    Y P R - block
    C1 C2 C3 C4 - column
  )
  
  (:init

    (clear Y)
    (clear P)
    (clear R)

    (inColumn Y C2)
    (inColumn P C1)
    (inColumn R C4)

    (rightOf C2 C1)
    (rightOf C3 C2)
    (rightOf C4 C3)

    (leftOf C1 C2)
    (leftOf C2 C3)
    (leftOf C3 C4)
  )
  (:goal
    (and

      (clear Y)
      (clear P)
      (clear R)

      (inColumn Y C3)
      (inColumn P C4)
      (inColumn R C1)
    )
  )
)
\end{pddlcode}
\caption{\textbf{ViPlan-BW Problem.} Example of a problem for the ViPlan-BW domain.}
\label{pddl:Blocksworld_problem}
\end{figure*}

\subsection{ViPlan-HH}
\label{app:domains_iGibson}

\begin{table}[t]

\centering
\caption{\textbf{Task splits used in the ViPlan-HH domain of our benchmark.} Tasks appearing in multiple splits have been adjusted in terms of goal to achieve, to make them easier or harder, depending on the split. The two additional columns report the number of problem instances per task and the minimum number of actions required to complete each instance.}
\label{tab:HH_task_splits}
\resizebox{\linewidth}{!}{
\begin{tabular}{llcc}
\toprule
\textbf{Difficulty} & \textbf{Task} & \textbf{\# Instances} & \textbf{\# Actions} \\
\midrule
\multirow{5}{*}{Simple} 
    & cleaning out drawers        & 5 & 5 \\
    & locking every door          & 5  & 4  \\
    & locking every window        & 5  & 6  \\
    & packing food for work       & 5 & 5 \\
    & sorting books               & 5  & 4 \\
\midrule
\multirow{6}{*}{Medium} 
    & cleaning out drawers        & 2 & 10 \\
    & collect misplaced items     & 4 & 8 \\
    & packing food for work       & 4 & 10 \\
    & putting away toys           & 5  & 8 \\
    & sorting books               & 4 & 8 \\
    & sorting groceries           & 6 & 10 \\
\midrule
\multirow{5}{*}{Hard} 
    & cleaning out drawers        & 5 & 15 \\
    & organizing boxes in garage  & 5 & 11 \\
    & organizing file cabinet     & 4 & 14 \\
    & putting away toys           & 4 & 12 \\
    & sorting groceries           & 7 & 13 \\
\bottomrule
\end{tabular}}
\end{table}

ViPlan-HH is composed of a selection of Behavior Domain Definition Language (BDDL)\footnote{This is a simplified version of PDDL, where no domain is provided, only the problem.} problems from the BEHAVIOR-100 task suite \citep{srivastava2022behavior}. As BDDL problems are not enough for planning, but are used as constraints to generate the problem layout in the iGibson simulator, we translated them into PDDL and wrote an adequate domain, reported in Figure~\ref{pddl:iGibson}.
\paragraph{}
The iGibson simulator includes multiple scenes that are compatible with each problem, and each scene has multiple instances (e.g., different dispositions of objects or different looking 3D models for the same object). 
Thus, in order to reach 25 problems, we select 5 tasks for the simple split, 6 tasks for the medium split and 5 tasks for the hard split. For each tasks, we write a PDDL problem file (see e.g., Figure~\ref{pddl:iGibson_problem}) and then select multiple scenes and instances, to reach a total of 25 unique (PDDL problem, scene, scene instance) triplets, as reported in Table~\ref{tab:HH_task_splits}.
For each split, the problems are edited (e.g., by considering more or less objects) to ensure that length of the plan required to solve them matches the one used for ViPlan-BW (Table~\ref{tab:bw_task_splits}). Sample images from iGibson can be found in Figures~1 and 2 of the main text.
\paragraph{}
Note that a successor to iGibson, named OmniGibson \citep{li2024behavior}, was recently proposed. While it provides more realistic graphics, which would be an advantage for VLM evaluation, it requires a NVIDIA RTX GPU (NVIDIA RTX 2070+), causing significant accessibility issues. We thus opted for iGibson instead for ease of reuse. 

\subsubsection{ViPlan-HH Original Contributions}
While we build on top of iGibson and leverage BDDL problems from BEHAVIOR-100, adapting the environment to our benchmark required months of work. Our main contributions in this regard can be summarized as follows:
\begin{enumerate}
    \item Implement a \textbf{PDDL domain} and interface it with a symbolic planner
    \item Implement in iGibson each \textbf{predicate} and \textbf{high-level action} listed in the domain by leveraging privileged low-level access to the internal states of the simulator
    \item Add \textbf{3D bounding boxes with object labels} when multiple objects of the same type are in sight
    \item Building a \textbf{server-client framework} to enable containerization of the iGibson environment with GPU acceleration on a SLURM server
\end{enumerate}

\onecolumn
\begin{pddlcode}[title=ViPlan-HH Domain]
(define (domain igibson)
    (:requirements :strips :typing :negative-preconditions :conditional-effects :equality)

    (:types
        container movable - object
    )
    
    (:predicates
        ;; Agent predicates
        (reachable ?o - object)
        (holding ?m - movable)

        ;; Object attributes
        (open ?c - container)

        ;; Object relations
        (ontop ?o1 - object ?o2 - object) 
        (nextto ?o1 - object ?o2 - object) 

        ;; Only containers can contain objects
        (inside ?o - object ?c - container) 
    )
    
    (:action grasp
        :parameters (?m - movable)
        :precondition (and
            (reachable ?m)
            ;; Agent must not be holding anything
            (forall (?x - movable)
                (not (holding ?x))
            ) 
        )
        :effect (and
            (holding ?m)
            (forall (?y - object)
                (and
                    ;; If grasped object is on top of something, 
                    ;; it is no longer on top of it
                    (not (ontop ?m ?y)) 
                    
                    ;; Same for nextto
                    (not (nextto ?m ?y)))
            ) 
                    
            ;; If m was in a container, it's not anymore
            (forall (?c - container)
                (when (inside ?m ?c) (not (inside ?m ?c)))
            ) 
        )
    )

    (:action place-on
        :parameters (?m - movable ?o2 - object)
        :precondition (and
            (holding ?m)
            (reachable ?o2)
        )
        :effect (and
            (ontop ?m ?o2)
            (not (holding ?m))
        )
    )

    (:action place-next-to
        :parameters (?m - movable ?o2 - object)
        :precondition (and
            (holding ?m)
            (reachable ?o2)
        )
        :effect (and
            (nextto ?m ?o2) 
            (not (holding ?m))
        )
    )

    (:action place-inside
        :parameters (?m - movable ?c - container)
        :precondition (and
            (holding ?m)
            (reachable ?c)
            (open ?c)
        )
        :effect (and
            (inside ?m ?c)
            (not (holding ?m))
        )
    )

    (:action open-container
        :parameters (?c - container)
        :precondition (and
            (reachable ?c)
            (not (open ?c))
            ;; Agent must not be holding anything
            (forall (?x - movable)
                (not (holding ?x))
            ) 
        )
        :effect (and
            (open ?c)
            ;; All objects inside the container are reachable
            (forall (?o - object)
                (when (inside ?o ?c) (reachable ?o))
            ) 
        )
    )

    (:action close-container
        :parameters (?c - container)
        :precondition (and
            (reachable ?c)
            (open ?c)
        )
        :effect (and
            (not (open ?c))
            ;; All objects inside the container are unreachable
            (forall (?o - object)
                (when (inside ?o ?c) (not (reachable ?o))) 
            ) 
        )
    )

    (:action navigate-to
        :parameters (?o - object)
        :precondition (and
            (not (reachable ?o))
            ;; Do not navigate-to things hidden in a closed container
            (forall (?c - container)
                (or (not(inside ?o ?c)) (open ?c))
            )
        )
        :effect (and
            (reachable ?o) ;; make target object reachable

            ;; Make every other object unreachable
            (forall (?x - object)
                (when (not (= ?x ?o)) 
                    (not (reachable ?x)))) 

            ;; Also, if there exists a container which is ?o and that it's open,
            ;; set the objects inside as reachable
            (forall (?c - container ?x - object)
                (when (and (= ?c ?o) (open ?c) (inside ?x ?c))
                    (reachable ?x)))
        )
    )
)
\end{pddlcode}
\captionof{figure}{PDDL domain for the ViPlan-HH environment.}
\label{pddl:iGibson}
\twocolumn

\begin{figure*}
\begin{pddlcode}[title=Example Problem for ViPlan-HH]
(define (problem cleaning_out_drawers_0)
    (:domain igibson)

    (:objects
     	bowl_1 - movable
    	cabinet_1 - container
    	sink_1 - object
    )
    
    (:init 
        (inside bowl_1 cabinet_1) 
        (not (open cabinet_1))
    )
    
    (:goal 
        (and 
            (ontop bowl_1 sink_1) 
        )
    )
)
\end{pddlcode}
\caption{\textbf{ViPlan-HH Problem.} Example of a problem for the ViPlan-HH domain.}
\label{pddl:iGibson_problem}
\end{figure*}

\section{Prompts}
\label{app:prompts}
In the following, we report prompts for the Ground, Ground + CoT, Plan, and Plan + CoT methods used in our experiments. The prompts for Action and Action + CoT use the same format as Plan and Plan + CoT respectively, with the JSON only containing a single "action" element rather than a full plan (and the wording in the system prompt adjusted accordingly to ask for a single action rather than a plan). For the memory augmented Ground variants, the prompts are the same as the ones reported, with the memory context appended at the end of the user prompt. All possible memory augmentations are reported after the prompts.

\onecolumn
\begin{prompt}[title=Ground prompt for ViPlan-BW]
<system> 
You are tasked with replying to a question about the given image. You will be given a single question, defined after the keyword "Question:" and will need to answer it ONLY with Yes or No. Do not write anything else besides your answer. 

The image will be about colored blocks and how they relate to each other. In the environment, the blocks will be arranged in columns, spanning from left to right. Keep in mind that some of these columns can be empty with no blocks currently placed in them. Within a column one or multiple blocks of different colors can be stacked on top of each other. Your task is to correctly evaluate the question based on the image provided. 
</system> 
<user> 
{image}
</user>
\end{prompt}

\begin{prompt}[title=Ground + CoT prompt for ViPlan-BW]
<system>
You are tasked with replying to a question about the given image. You will be given a single question, defined after the keyword "Question:" and will need to first reason about it and then give a Yes or No answer. The reasoning for your answer should be written within the XML-style <explanation></explanation> tags. To write the final answer, you should write only "Yes" or "No" surrounded by <answer></answer> tags. Do not write anything else besides your step-by-step reasoning and your answer. 

Example output for a question about the an image:
```
Question: Is there a dog on top of a table?
<explanation>
First, I will look for a dog in the image. Then, I will check if the dog is on top of a table. In the image, there is a dog and there is a table, but the dog is not on top of the table. Therefore, the answer is "No".
</explanation>
<answer>
No
</answer>
```
</system>
<user>
The image will be about colored blocks and how they relate to each other. In the environment, the blocks will be arranged in columns, spanning from left to right. Keep in mind that some of these columns can be empty with no blocks currently placed in them. Within a column one or multiple blocks of different colors can be stacked on top of each other.

{image}
</user>
\end{prompt}

\begin{prompt}[title=Ground prompt for ViPlan-HH]
<system>
You are tasked with replying to a question about the given image. You will be given a single question, defined after the keyword "Question:" and will need to answer it ONLY with Yes or No. Do not write anything else besides your answer.
</system>
<user>
The environment is a virtual household simulator, with objects and furniture which can be interacted with. There is a robotic arm, which is the agent, that can hold objects.
{image}
</user>
\end{prompt}

\begin{prompt}[title=Ground + CoT prompt for ViPlan-HH]
<system>
You are tasked with replying to a question about the given image. You will be given a single question, defined after the keyword "Question:" and will need to first reason about it and then give a Yes or No answer. The reasoning for your answer should be written within the XML-style <explanation></explanation> tags. To write the final answer, you should write only "Yes" or "No" surrounded by <answer></answer> tags. Do not write anything else besides your step-by-step reasoning and your answer. 

Example output for a question about the an image:
```
Question: Is there a dog on top of a table?
<explanation>
First, I will look for a dog in the image. Then, I will check if the dog is on top of a table. In the image, there is a dog and there is a table, but the dog is not on top of the table. Therefore, the answer is "No".
</explanation>
<answer>
No
</answer>
```
</system>

<user>
The environment is a virtual household simulator, with objects and furniture which can be interacted with. There is a robotic arm, which is the agent, that can hold objects.

{image}
</user>
\end{prompt}

\begin{prompt}[title=Plan prompt for ViPlan-BW]
<system> 
You are an expert planning assistant. You will be given an image which represents the current state of the environment you are in, a natural language description of the goal that needs to be achieved and a set of actions that can be performed in the environment. 
Your task is to generate a plan that achieves the goal, in the form of a sequence of actions that need to be executed to reach the goal.
The format of your output should be a JSON object with the following structure:
```json
{
  "plan": [
    {
        "action": action_name,
        "parameters": {
            parameter_name: parameter_value
        }
    },
    ... other actions ...
    ]
}
```

You will also receive feedback of the previously taken actions, with a note showing if they failed or not. If an action failed, think about why that could be and then output a new plan accordingly.
</system>
<user>
## Description of the environment
The environment is about colored blocks and how they relate to each other. In the environment, the blocks will be arranged in columns, spanning from left to right. Keep in mind that some of these columns can be empty with no blocks currently placed in them. Within a column one or multiple blocks of different colors can be stacked on top of each other. Your task is to correctly evaluate the question based on the image provided.

## Available actions
You have only one action available, called `moveblock(block, column)`. This action allows you to move a block from its current column to the specified column. In order to perform this action, the block you want to move must be the topmost block of its column and must not already be in the target column. If the action is valid, the block will be moved to the specified column and will be placed on top of any blocks that are already in that column, if any.
To refer to the blocks, use lowercase letters for the colors: 'r' for red, 'g' for green, 'b' for blue, 'y' for yellow, 'p' for purple, 'o' for orange. To refer to the columns, use the labels provided in the image, 'c1', 'c2', 'c3', 'c4' and 'c5'.

## Goal
{goal_string}

## Previously taken actions
{previous_actions}

## Current environment state
{image}
</user>
\end{prompt}

\begin{prompt}[title=Plan + CoT prompt for ViPlan-BW]
<system> 
You are an expert planning assistant. You will be given an image which represents the current state of the environment you are in, a natural language description of the goal that needs to be achieved and a set of actions that can be performed in the environment. 
Your task is to generate a plan that achieves the goal, in the form of a sequence of actions that need to be executed to reach the goal.
Before answering with the plan, think carefully step by step about the actions you need to take and what the expected outcome of each action is. Write the reasoning behind the plan and justify each action you are going to take. Make sure that each action is possible, and if previous actions failed, reason about why this could be the case.

The format of your output should be a JSON object with the following structure. Make sure that the explanation is also written inside the json.
```json
{
  "explanation": <a detailed explanation of the plan>,
  "plan": [
    {
        "action": action_name,
        "parameters": {
            parameter_name: parameter_value
        }
    },
    ... other actions ...
    ]
}
```

You will also receive feedback of the previously taken actions, with a note showing if they failed or not. If an action failed, think about why that could be and then output a new plan accordingly.
</system>
<user>
## Description of the environment
The environment is about colored blocks and how they relate to each other. In the environment, the blocks will be arranged in columns, spanning from left to right. Keep in mind that some of these columns can be empty with no blocks currently placed in them. Within a column one or multiple blocks of different colors can be stacked on top of each other. Your task is to correctly evaluate the question based on the image provided.

## Available actions
You have only one action available, called `moveblock(block, column)`. This action allows you to move a block from its current column to the specified column. In order to perform this action, the block you want to move must be the topmost block of its column and must not already be in the target column. If the action is valid, the block will be moved to the specified column and will be placed on top of any blocks that are already in that column, if any.
To refer to the blocks, use lowercase letters for the colors: 'r' for red, 'g' for green, 'b' for blue, 'y' for yellow, 'p' for purple, 'o' for orange. To refer to the columns, use the labels provided in the image, 'c1', 'c2', 'c3', 'c4' and 'c5'.

## Goal
{goal_string}

## Previously taken actions
{previous_actions}

## Current environment state
{image}
</user>
\end{prompt}

\begin{prompt}[title=Plan prompt for ViPlan-HH]
<system> 
You are an expert planning assistant. You will be given an image which represents the current state of the environment you are in, a natural language description of the goal that needs to be achieved and a set of actions that can be performed in the environment. 
Your task is to generate a plan that achieves the goal, in the form of a sequence of actions that need to be executed to reach the goal.
The format of your output should be a JSON object with the following structure:
```json
{
  "plan": [
    {
        "action": action_name,
        "parameters": ['parameter1', 'parameter2', ...]
    },
    ... other actions ...
    ]
}
```

You will also receive feedback of the previously taken actions, with a note showing if they failed or not. If an action failed, think about why that could be and then output a new plan accordingly.
</system>
<user>

## Description of the environment
The environment is a virtual household simulator, with objects and furniture which can be interacted with. Keep in mind that some objects might not be visible or immediately reachable, in which case you need to navigate to them first. If after navigating to an object it is still not reachable, you might need to open a container.

## Additional information
{privileged_info}

## Available actions

- Action: grasp  
  - Parameters:  
    1. a movable object  
  - Preconditions:  
    - The object is within reach.  
    - The agent is not holding anything.  
  - Effects:  
    - The agent picks up that object.  
    - It is no longer on top of or next to any other object.  
    - If it was inside a container, it leaves the container.  

- Action: place-on  
  - Parameters:  
    1. the movable object being held  
    2. another object to serve as support  
  - Preconditions:  
    - The agent is holding the first object.  
    - The support object is within reach.  
  - Effects:  
    - The held object is placed on top of the support object.  
    - The agent's hands become free.  

- Action: place-next-to  
  - Parameters:  
    1. the movable object being held  
    2. another object to stand beside  
  - Preconditions:  
    - The agent is holding the first object.  
    - The other object is within reach.  
  - Effects:  
    - The held object is positioned next to the other object.  
    - The agent's hands become free.  

- Action: place-inside  
  - Parameters:  
    1. the movable object being held  
    2. an open container  
  - Preconditions:  
    - The agent is holding the object.  
    - The container is open and within reach.  
  - Effects:  
    - The object is placed inside the container.  
    - The agent's hands become free.  

- Action: open-container  
  - Parameters:  
    1. a closed container  
  - Preconditions:  
    - The container is within reach.  
    - The agent is not holding anything.  
  - Effects:  
    - The container becomes open.  
    - All objects inside it become reachable.  

- Action: close-container  
  - Parameters:  
    1. an open container  
  - Preconditions:  
    - The container is within reach.  
  - Effects:  
    - The container becomes closed.  
    - All objects inside it become unreachable.  

- Action: navigate-to  
  - Parameters:  
    1. any target object  
  - Preconditions:  
    - The target object is currently out of reach and not hidden in a closed container.  
  - Effects:  
    - The target object becomes reachable.  
    - All other objects become out of reach.  
    - If the target is an open container, everything inside it also becomes reachable.  

## Goal
{goal_string}

## Previously taken actions
{previous_actions}

## Current environment state
{image}
</user>
\end{prompt}

\begin{prompt}[title=Plan + CoT prompt for ViPlan-HH]
<system> 
You are an expert planning assistant. You will be given an image which represents the current state of the environment you are in, a natural language description of the goal that needs to be achieved and a set of actions that can be performed in the environment. 
Your task is to generate a plan that achieves the goal, in the form of a sequence of actions that need to be executed to reach the goal.
Before answering with the plan, think carefully step by step about the actions you need to take and what the expected outcome of each action is. Write the reasoning behind the plan and justify each action you are going to take. Make sure that each action is possible, and if previous actions failed, reason about why this could be the case.

The format of your output should be a JSON object with the following structure. Make sure that the explanation is also written inside the json.
```json
{
  "explanation": <a detailed explanation of the plan>,
  "plan": [
    {
        "action": action_name,
        "parameters": ['parameter1', 'parameter2', ...]
    },
    ... other actions ...
    ]
}
```

You will also receive feedback of the previously taken actions, with a note showing if they failed or not. If an action failed, think about why that could be and then output a new plan accordingly.
</system>
<user>

## Description of the environment
The environment is a virtual household simulator, with objects and furniture which can be interacted with. Keep in mind that some objects might not be visible or immediately reachable, in which case you need to navigate to them first. If after navigating to an object it is still not reachable, you might need to open a container.

## Additional information
{privileged_info}

## Available actions

- Action: grasp  
  - Parameters:  
    1. a movable object  
  - Preconditions:  
    - The object is within reach.  
    - The agent is not holding anything.  
  - Effects:  
    - The agent picks up that object.  
    - It is no longer on top of or next to any other object.  
    - If it was inside a container, it leaves the container.  

- Action: place-on  
  - Parameters:  
    1. the movable object being held  
    2. another object to serve as support  
  - Preconditions:  
    - The agent is holding the first object.  
    - The support object is within reach.  
  - Effects:  
    - The held object is placed on top of the support object.  
    - The agent's hands become free.  

- Action: place-next-to  
  - Parameters:  
    1. the movable object being held  
    2. another object to stand beside  
  - Preconditions:  
    - The agent is holding the first object.  
    - The other object is within reach.  
  - Effects:  
    - The held object is positioned next to the other object.  
    - The agent's hands become free.  

- Action: place-inside  
  - Parameters:  
    1. the movable object being held  
    2. an open container  
  - Preconditions:  
    - The agent is holding the object.  
    - The container is open and within reach.  
  - Effects:  
    - The object is placed inside the container.  
    - The agent's hands become free.  

- Action: open-container  
  - Parameters:  
    1. a closed container  
  - Preconditions:  
    - The container is within reach.  
    - The agent is not holding anything.  
  - Effects:  
    - The container becomes open.  
    - All objects inside it become reachable.  

- Action: close-container  
  - Parameters:  
    1. an open container  
  - Preconditions:  
    - The container is within reach.  
  - Effects:  
    - The container becomes closed.  
    - All objects inside it become unreachable.  

- Action: navigate-to  
  - Parameters:  
    1. any target object  
  - Preconditions:  
    - The target object is currently out of reach and not hidden in a closed container.  
  - Effects:  
    - The target object becomes reachable.  
    - All other objects become out of reach.  
    - If the target is an open container, everything inside it also becomes reachable.  

## Goal
{goal_string}

## Previously taken actions
{previous_actions}

## Current environment state
{image}
</user>
\end{prompt}

\begin{prompt}[title=Memory Context for Ground + Mem Variants]
Previously, in the same state (image) as the one shown, you were asked these questions and provided the following answers:
    - Q: {question}
      A: {answer}
      (repeated for all relevant Q/A pairs)

(the next line varies based on the previous outcome)
(if the previously selected action was illegal, but the VLM thought it was legal:)
Based on these answers, the action '{action_str}' was attempted, but something went wrong. This is very likely to have been caused by one or more incorrect answers above. Keep this in mind while answering the following question.

(if the previous action was deemed illegal by the VLM due to unsatisfied preconditions, but there is a chance it could have been legal:)
Based on these answers, the previously-planned action '{action_str}' was called off, as at least one of its preconditions was judged invalid. This might have been the correct choice or a mistake. Keep this in mind while answering the following question.

(if an unobserved precondition was not satisfied:)
Based on these answers, the action '{action_str}' was attempted, but something went wrong. This may have been due to an unobserved precondition not being met. Keep this in mind while answering the following question.
\end{prompt}

\twocolumn

\section{Complete Selection Results}
\label{app:complete_benchmark_results}

\begin{figure*}[h]
    \centering
    \includegraphics[width=\linewidth]{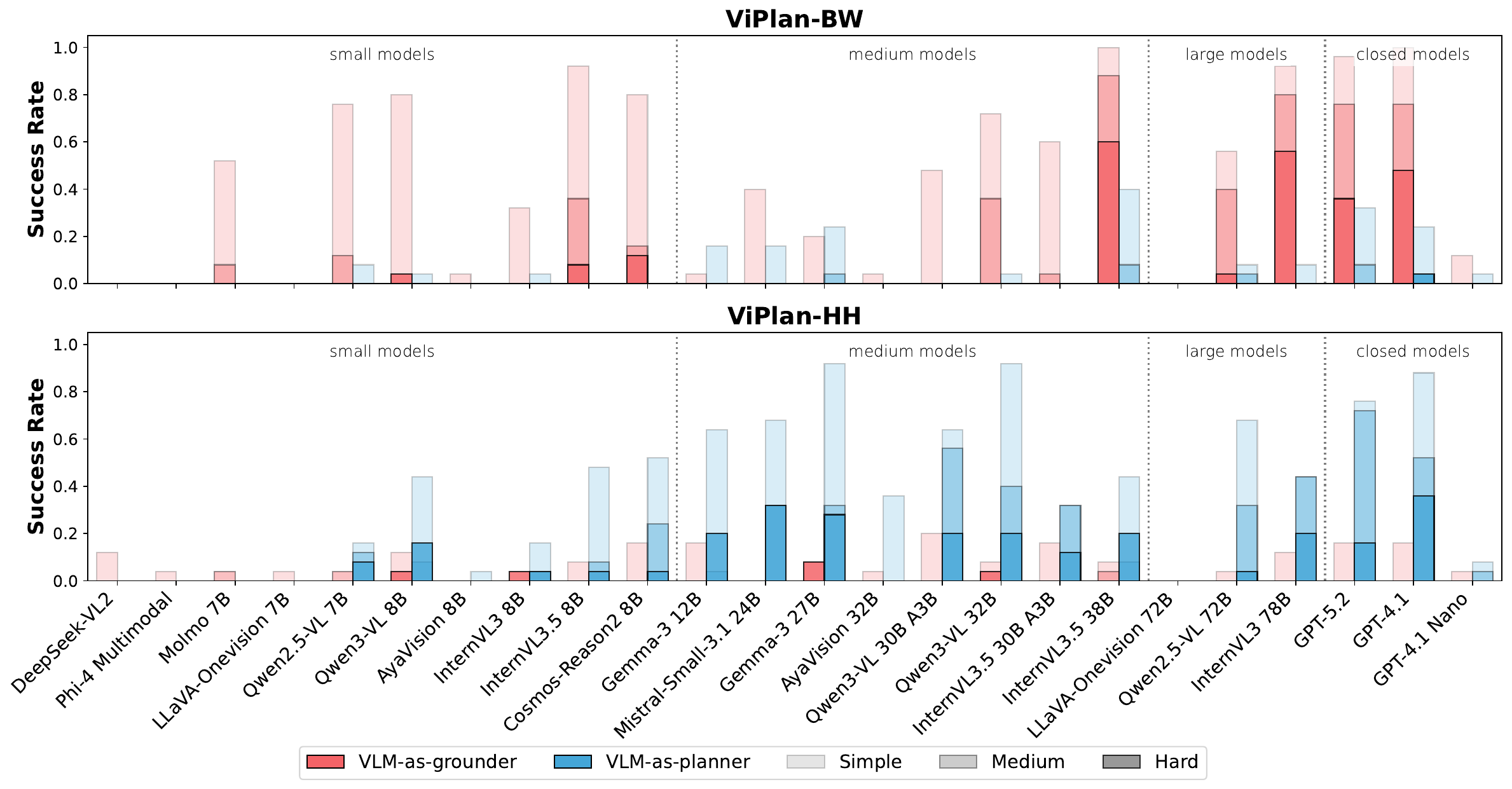}
        \caption{\textbf{Expanded Selection Results.} The \textcolor{myred}{Ground} approach excels in ViPlan-BW (top), where GPT-4.1, InternVL3 78B and InternVL 3.5 38B complete a significant fraction of tasks, matching the performance of GPT-5.2 and GPT-4.1. The \textcolor{myblue}{Plan} approach is instead better on ViPlan-HH (bottom), where medium, large and closed models perform generally better than with \textcolor{myred}{Ground}, with Gemma-3 27B and Qwen3-VL 32B standing out as medium-sized models.}
    \label{fig:pred_vs_vila}
\end{figure*}

\begin{figure*}[t]
    \centering
    \includegraphics[width=\linewidth]{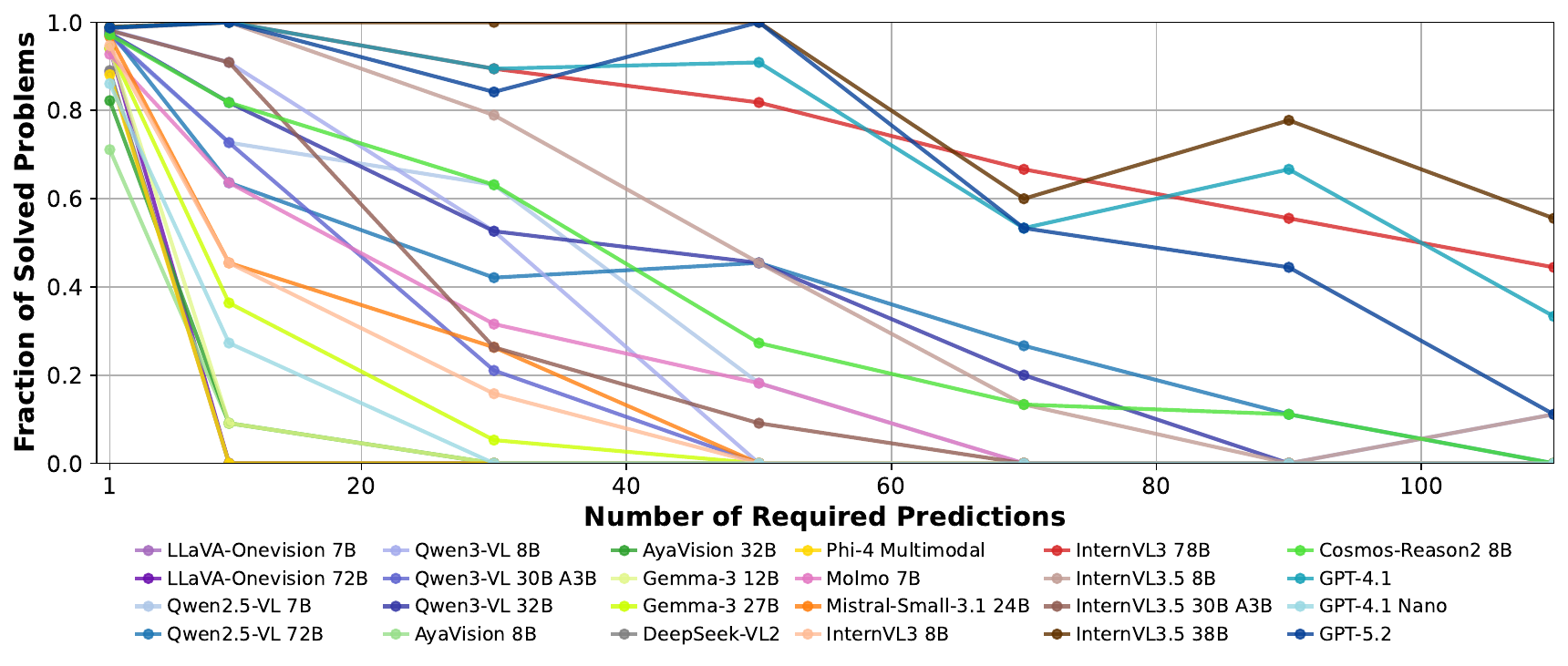}
    \caption{\textbf{Compounding Errors in Planning.} Analysis of the fractions of tasks solved in ViPlan-BW (all difficulties), based on the number of predictions a model would need to answer correctly to succeed for the Ground method. Benchmarks that ask independent questions to VLMs and measure their accuracy do not capture the effect that compounding errors have on solving a problem, and would correspond to measuring only one prediction. Most models score well on the single prediction, but deteriorate quickly as the errors compound.}
    \label{fig:predictions_solved}
\end{figure*}

We report the individual success rate and accuracy for each model, method, domain and split in Tables~\ref{tab:ViPLan-BW_pred_full}, \ref{tab:ViPLan-BW_history_full}, \ref{tab:ViPLan-HH_history_full}, \ref{tab:ViPLan-BW_vila_full}, \ref{tab:ViPLan-HH_pred_full}, \ref{tab:ViPLan-HH_vila_full}, \ref{tab:ViPLan-BW_act_react_full} and \ref{tab:ViPLan-HH_act_react_full}. The task success rates for the full selection results are also visualized in Figure~\ref{fig:pred_vs_vila}. Besides task success rate (which we use as our main metric), we further report predicate accuracy for Ground variants, computed as the fraction of yes-no questions that were answered correctly when compared to the ground truth environment. In practice, we observe that models generally show very high accuracy ($\geq 90\%$ for many model families); however, this does not always translate to high task success rate. This is due to compounding errors preventing successful completions: especially as the complexity grows, one episode requires up to 120 questions to be answered correctly, so even very accurate models can make a single mistake that compromises the task, as shown in Figure~\ref{fig:predictions_solved}.

\begin{table*}[h]
\caption{\textbf{Full Results for Ground and Ground + CoT on ViPlan-BW.} Success rate indicates the fraction of problems solved, accuracy is the fraction of correct predictions for a single predicate. The result is an average over the 25 problems in each split. Standard error of the mean is reported in parenthesis. The highest value in each column is bolded.}
\label{tab:ViPLan-BW_pred_full}
\centering
\resizebox{\linewidth}{!}{\scriptsize
  \begin{tabular}{l c c c c c c c}
\toprule
\multirow{2}{*}{\textbf{Model}} & \multirow{2}{*}{\textbf{CoT}} & \multicolumn{2}{c}{\textbf{Simple}} & \multicolumn{2}{c}{\textbf{Medium}} & \multicolumn{2}{c}{\textbf{Hard}} \\
 &  & \textbf{Success} & \textbf{Accuracy} & \textbf{Success} & \textbf{Accuracy} & \textbf{Success} & \textbf{Accuracy} \\
\midrule
AyaVision 32B & $\times$ & 0.04 \scriptsize{(0.04)} & 0.88 \scriptsize{(0.01)} & 0.00 \scriptsize{(0.00)} & 0.82 \scriptsize{(0.00)} & 0.00 \scriptsize{(0.00)} & 0.76 \scriptsize{(0.01)} \\
AyaVision 32B & $\checkmark$ & 0.00 \scriptsize{(0.00)} & 0.89 \scriptsize{(0.01)} & 0.04 \scriptsize{(0.04)} & 0.85 \scriptsize{(0.00)} & 0.00 \scriptsize{(0.00)} & 0.79 \scriptsize{(0.00)} \\
AyaVision 8B & $\times$ & 0.04 \scriptsize{(0.04)} & 0.68 \scriptsize{(0.01)} & 0.00 \scriptsize{(0.00)} & 0.69 \scriptsize{(0.01)} & 0.00 \scriptsize{(0.00)} & 0.76 \scriptsize{(0.01)} \\
AyaVision 8B & $\checkmark$ & 0.00 \scriptsize{(0.00)} & 0.51 \scriptsize{(0.01)} & 0.00 \scriptsize{(0.00)} & 0.53 \scriptsize{(0.01)} & 0.00 \scriptsize{(0.00)} & 0.60 \scriptsize{(0.01)} \\
Cosmos\text{-}Reason2 8B & $\times$ & 0.80 \scriptsize{(0.08)} & 0.99 \scriptsize{(0.00)} & 0.16 \scriptsize{(0.07)} & 0.97 \scriptsize{(0.00)} & 0.12 \scriptsize{(0.06)} & 0.96 \scriptsize{(0.00)} \\
Cosmos\text{-}Reason2 8B & $\checkmark$ & 0.44 \scriptsize{(0.10)} & 0.97 \scriptsize{(0.00)} & 0.12 \scriptsize{(0.06)} & 0.98 \scriptsize{(0.00)} & 0.04 \scriptsize{(0.04)} & 0.96 \scriptsize{(0.00)} \\
DeepSeek\text{-}VL2 & $\times$ & 0.00 \scriptsize{(0.00)} & 0.91 \scriptsize{(0.01)} & 0.00 \scriptsize{(0.00)} & 0.88 \scriptsize{(0.01)} & 0.00 \scriptsize{(0.00)} & 0.88 \scriptsize{(0.01)} \\
DeepSeek\text{-}VL2 & $\checkmark$ & 0.04 \scriptsize{(0.04)} & 0.91 \scriptsize{(0.00)} & 0.00 \scriptsize{(0.00)} & 0.85 \scriptsize{(0.00)} & 0.00 \scriptsize{(0.00)} & 0.86 \scriptsize{(0.00)} \\
GPT\text{-}4.1 & $\times$ & \textbf{1.00 \scriptsize{(0.00)}} & \textbf{1.00 \scriptsize{(0.00)}} & 0.76 \scriptsize{(0.09)} & \textbf{0.99 \scriptsize{(0.00)}} & 0.48 \scriptsize{(0.10)} & \textbf{0.98 \scriptsize{(0.00)}} \\
GPT\text{-}4.1 & $\checkmark$ & 0.92 \scriptsize{(0.05)} & 0.99 \scriptsize{(0.00)} & 0.76 \scriptsize{(0.09)} & 0.98 \scriptsize{(0.00)} & 0.44 \scriptsize{(0.10)} & \textbf{0.98 \scriptsize{(0.00)}} \\
GPT\text{-}4.1 Nano & $\times$ & 0.12 \scriptsize{(0.06)} & 0.89 \scriptsize{(0.01)} & 0.00 \scriptsize{(0.00)} & 0.85 \scriptsize{(0.00)} & 0.00 \scriptsize{(0.00)} & 0.84 \scriptsize{(0.00)} \\
GPT\text{-}4.1 Nano & $\checkmark$ & 0.36 \scriptsize{(0.10)} & 0.96 \scriptsize{(0.00)} & 0.00 \scriptsize{(0.00)} & 0.96 \scriptsize{(0.00)} & 0.00 \scriptsize{(0.00)} & 0.92 \scriptsize{(0.00)} \\
GPT\text{-}5.2 & $\times$ & 0.96 \scriptsize{(0.04)} & 0.99 \scriptsize{(0.00)} & 0.76 \scriptsize{(0.09)} & \textbf{0.99 \scriptsize{(0.00)}} & 0.36 \scriptsize{(0.10)} & \textbf{0.98 \scriptsize{(0.00)}} \\
Gemma\text{-}3 12B & $\times$ & 0.04 \scriptsize{(0.04)} & 0.96 \scriptsize{(0.01)} & 0.00 \scriptsize{(0.00)} & 0.95 \scriptsize{(0.00)} & 0.00 \scriptsize{(0.00)} & 0.91 \scriptsize{(0.00)} \\
Gemma\text{-}3 12B & $\checkmark$ & 0.16 \scriptsize{(0.07)} & 0.93 \scriptsize{(0.00)} & 0.00 \scriptsize{(0.00)} & 0.91 \scriptsize{(0.01)} & 0.00 \scriptsize{(0.00)} & 0.88 \scriptsize{(0.01)} \\
Gemma\text{-}3 27B & $\times$ & 0.20 \scriptsize{(0.08)} & 0.96 \scriptsize{(0.00)} & 0.00 \scriptsize{(0.00)} & 0.93 \scriptsize{(0.00)} & 0.00 \scriptsize{(0.00)} & 0.93 \scriptsize{(0.01)} \\
Gemma\text{-}3 27B & $\checkmark$ & 0.28 \scriptsize{(0.09)} & 0.96 \scriptsize{(0.00)} & 0.04 \scriptsize{(0.04)} & 0.94 \scriptsize{(0.00)} & 0.00 \scriptsize{(0.00)} & 0.91 \scriptsize{(0.00)} \\
InternVL3 78B & $\times$ & 0.92 \scriptsize{(0.05)} & \textbf{1.00 \scriptsize{(0.00)}} & 0.80 \scriptsize{(0.08)} & \textbf{0.99 \scriptsize{(0.00)}} & 0.56 \scriptsize{(0.10)} & \textbf{0.98 \scriptsize{(0.00)}} \\
InternVL3 78B & $\checkmark$ & 0.96 \scriptsize{(0.04)} & \textbf{1.00 \scriptsize{(0.00)}} & 0.80 \scriptsize{(0.08)} & \textbf{0.99 \scriptsize{(0.00)}} & 0.56 \scriptsize{(0.10)} & \textbf{0.98 \scriptsize{(0.00)}} \\
InternVL3 8B & $\times$ & 0.32 \scriptsize{(0.09)} & 0.97 \scriptsize{(0.00)} & 0.00 \scriptsize{(0.00)} & 0.95 \scriptsize{(0.00)} & 0.00 \scriptsize{(0.00)} & 0.92 \scriptsize{(0.00)} \\
InternVL3 8B & $\checkmark$ & 0.08 \scriptsize{(0.05)} & 0.95 \scriptsize{(0.00)} & 0.00 \scriptsize{(0.00)} & 0.92 \scriptsize{(0.00)} & 0.00 \scriptsize{(0.00)} & 0.86 \scriptsize{(0.00)} \\
InternVL3.5 30B A3B & $\times$ & 0.60 \scriptsize{(0.10)} & 0.99 \scriptsize{(0.00)} & 0.04 \scriptsize{(0.04)} & 0.98 \scriptsize{(0.00)} & 0.00 \scriptsize{(0.00)} & 0.97 \scriptsize{(0.00)} \\
InternVL3.5 30B A3B & $\checkmark$ & 0.80 \scriptsize{(0.08)} & 0.99 \scriptsize{(0.00)} & 0.24 \scriptsize{(0.09)} & 0.97 \scriptsize{(0.00)} & 0.00 \scriptsize{(0.00)} & 0.94 \scriptsize{(0.00)} \\
InternVL3.5 38B & $\times$ & \textbf{1.00 \scriptsize{(0.00)}} & \textbf{1.00 \scriptsize{(0.00)}} & \textbf{0.88 \scriptsize{(0.06)}} & \textbf{0.99 \scriptsize{(0.00)}} & \textbf{0.60 \scriptsize{(0.10)}} & \textbf{0.98 \scriptsize{(0.00)}} \\
InternVL3.5 38B & $\checkmark$ & \textbf{1.00 \scriptsize{(0.00)}} & \textbf{1.00 \scriptsize{(0.00)}} & \textbf{0.88 \scriptsize{(0.06)}} & 0.98 \scriptsize{(0.00)} & 0.36 \scriptsize{(0.10)} & 0.95 \scriptsize{(0.00)} \\
InternVL3.5 8B & $\times$ & 0.92 \scriptsize{(0.05)} & 0.99 \scriptsize{(0.00)} & 0.36 \scriptsize{(0.10)} & \textbf{0.99 \scriptsize{(0.00)}} & 0.08 \scriptsize{(0.05)} & \textbf{0.98 \scriptsize{(0.00)}} \\
InternVL3.5 8B & $\checkmark$ & 0.88 \scriptsize{(0.06)} & 0.99 \scriptsize{(0.00)} & 0.40 \scriptsize{(0.10)} & 0.98 \scriptsize{(0.00)} & 0.20 \scriptsize{(0.08)} & 0.97 \scriptsize{(0.00)} \\
LLaVA\text{-}Onevision 72B & $\times$ & 0.00 \scriptsize{(0.00)} & 0.95 \scriptsize{(0.01)} & 0.00 \scriptsize{(0.00)} & 0.94 \scriptsize{(0.00)} & 0.00 \scriptsize{(0.00)} & 0.93 \scriptsize{(0.01)} \\
LLaVA\text{-}Onevision 72B & $\checkmark$ & 0.04 \scriptsize{(0.04)} & 0.96 \scriptsize{(0.00)} & 0.00 \scriptsize{(0.00)} & 0.94 \scriptsize{(0.00)} & 0.00 \scriptsize{(0.00)} & 0.94 \scriptsize{(0.00)} \\
LLaVA\text{-}Onevision 7B & $\times$ & 0.00 \scriptsize{(0.00)} & 0.92 \scriptsize{(0.01)} & 0.00 \scriptsize{(0.00)} & 0.88 \scriptsize{(0.01)} & 0.00 \scriptsize{(0.00)} & 0.86 \scriptsize{(0.01)} \\
LLaVA\text{-}Onevision 7B & $\checkmark$ & 0.00 \scriptsize{(0.00)} & 0.93 \scriptsize{(0.01)} & 0.00 \scriptsize{(0.00)} & 0.88 \scriptsize{(0.01)} & 0.00 \scriptsize{(0.00)} & 0.89 \scriptsize{(0.01)} \\
Mistral\text{-}Small\text{-}3.1 24B & $\times$ & 0.40 \scriptsize{(0.10)} & 0.99 \scriptsize{(0.00)} & 0.00 \scriptsize{(0.00)} & 0.96 \scriptsize{(0.00)} & 0.00 \scriptsize{(0.00)} & 0.96 \scriptsize{(0.00)} \\
Mistral\text{-}Small\text{-}3.1 24B & $\checkmark$ & 0.56 \scriptsize{(0.10)} & 0.98 \scriptsize{(0.00)} & 0.04 \scriptsize{(0.04)} & 0.95 \scriptsize{(0.00)} & 0.00 \scriptsize{(0.00)} & 0.94 \scriptsize{(0.00)} \\
Molmo 7B & $\times$ & 0.52 \scriptsize{(0.10)} & 0.94 \scriptsize{(0.00)} & 0.08 \scriptsize{(0.05)} & 0.93 \scriptsize{(0.00)} & 0.00 \scriptsize{(0.00)} & 0.91 \scriptsize{(0.00)} \\
Molmo 7B & $\checkmark$ & 0.00 \scriptsize{(0.00)} & 0.77 \scriptsize{(0.01)} & 0.00 \scriptsize{(0.00)} & 0.75 \scriptsize{(0.01)} & 0.00 \scriptsize{(0.00)} & 0.74 \scriptsize{(0.01)} \\
Phi\text{-}4 Multimodal & $\times$ & 0.00 \scriptsize{(0.00)} & 0.92 \scriptsize{(0.01)} & 0.00 \scriptsize{(0.00)} & 0.89 \scriptsize{(0.01)} & 0.00 \scriptsize{(0.00)} & 0.83 \scriptsize{(0.01)} \\
Phi\text{-}4 Multimodal & $\checkmark$ & 0.00 \scriptsize{(0.00)} & 0.08 \scriptsize{(0.01)} & 0.00 \scriptsize{(0.00)} & 0.15 \scriptsize{(0.01)} & 0.00 \scriptsize{(0.00)} & 0.19 \scriptsize{(0.01)} \\
Qwen2.5\text{-}VL 72B & $\times$ & 0.56 \scriptsize{(0.10)} & 0.98 \scriptsize{(0.00)} & 0.40 \scriptsize{(0.10)} & \textbf{0.99 \scriptsize{(0.00)}} & 0.04 \scriptsize{(0.04)} & 0.97 \scriptsize{(0.00)} \\
Qwen2.5\text{-}VL 72B & $\checkmark$ & 0.48 \scriptsize{(0.10)} & 0.97 \scriptsize{(0.00)} & 0.20 \scriptsize{(0.08)} & \textbf{0.99 \scriptsize{(0.00)}} & 0.00 \scriptsize{(0.00)} & 0.96 \scriptsize{(0.00)} \\
Qwen2.5\text{-}VL 7B & $\times$ & 0.76 \scriptsize{(0.09)} & 0.99 \scriptsize{(0.00)} & 0.12 \scriptsize{(0.06)} & 0.97 \scriptsize{(0.00)} & 0.00 \scriptsize{(0.00)} & 0.94 \scriptsize{(0.00)} \\
Qwen2.5\text{-}VL 7B & $\checkmark$ & 0.08 \scriptsize{(0.05)} & 0.98 \scriptsize{(0.00)} & 0.00 \scriptsize{(0.00)} & 0.92 \scriptsize{(0.00)} & 0.00 \scriptsize{(0.00)} & 0.89 \scriptsize{(0.00)} \\
Qwen3\text{-}VL 30B A3B & $\times$ & 0.48 \scriptsize{(0.10)} & 0.98 \scriptsize{(0.00)} & 0.00 \scriptsize{(0.00)} & 0.97 \scriptsize{(0.00)} & 0.00 \scriptsize{(0.00)} & 0.96 \scriptsize{(0.00)} \\
Qwen3\text{-}VL 30B A3B & $\checkmark$ & 0.60 \scriptsize{(0.10)} & 0.97 \scriptsize{(0.00)} & 0.20 \scriptsize{(0.08)} & 0.97 \scriptsize{(0.00)} & 0.12 \scriptsize{(0.06)} & 0.97 \scriptsize{(0.00)} \\
Qwen3\text{-}VL 32B & $\times$ & 0.72 \scriptsize{(0.09)} & 0.98 \scriptsize{(0.00)} & 0.36 \scriptsize{(0.10)} & 0.98 \scriptsize{(0.00)} & 0.00 \scriptsize{(0.00)} & 0.96 \scriptsize{(0.00)} \\
Qwen3\text{-}VL 32B & $\checkmark$ & 0.56 \scriptsize{(0.10)} & 0.97 \scriptsize{(0.00)} & 0.44 \scriptsize{(0.10)} & 0.98 \scriptsize{(0.00)} & 0.04 \scriptsize{(0.04)} & 0.97 \scriptsize{(0.00)} \\
Qwen3\text{-}VL 8B & $\times$ & 0.80 \scriptsize{(0.08)} & 0.99 \scriptsize{(0.00)} & 0.00 \scriptsize{(0.00)} & 0.98 \scriptsize{(0.00)} & 0.04 \scriptsize{(0.04)} & \textbf{0.98 \scriptsize{(0.00)}} \\
Qwen3\text{-}VL 8B & $\checkmark$ & 0.80 \scriptsize{(0.08)} & 0.99 \scriptsize{(0.00)} & 0.40 \scriptsize{(0.10)} & 0.98 \scriptsize{(0.00)} & 0.40 \scriptsize{(0.10)} & 0.96 \scriptsize{(0.00)} \\
\bottomrule
\end{tabular}}
\end{table*}

\begin{table*}[h]
\caption{\textbf{Full Results for Ground + Mem and Ground + Mem + CoT on ViPlan-BW.} Success rate indicates the fraction of problems solved, accuracy is the fraction of correct predictions. The highest value in each column is bolded.}
\label{tab:ViPLan-BW_history_full}
\centering
\begin{tabular}{l c c c c c c c}
\toprule
\multirow{2}{*}{\textbf{Model}} & \multirow{2}{*}{\textbf{CoT}} & \multicolumn{2}{c}{\textbf{Simple}} & \multicolumn{2}{c}{\textbf{Medium}} & \multicolumn{2}{c}{\textbf{Hard}} \\
 &  & \textbf{Success} & \textbf{Accuracy} & \textbf{Success} & \textbf{Accuracy} & \textbf{Success} & \textbf{Accuracy} \\
\midrule
Gemma\text{-}3 27B & $\times$ & 0.24 \scriptsize{(0.09)} & 0.96 \scriptsize{(0.00)} & 0.00 \scriptsize{(0.00)} & 0.94 \scriptsize{(0.00)} & 0.00 \scriptsize{(0.00)} & 0.93 \scriptsize{(0.00)} \\
Gemma\text{-}3 27B & $\checkmark$ & 0.28 \scriptsize{(0.09)} & 0.96 \scriptsize{(0.00)} & 0.00 \scriptsize{(0.00)} & 0.93 \scriptsize{(0.00)} & 0.00 \scriptsize{(0.00)} & 0.92 \scriptsize{(0.00)} \\
InternVL3 78B & $\times$ & 0.92 \scriptsize{(0.05)} & \textbf{1.00 \scriptsize{(0.00)}} & 0.80 \scriptsize{(0.08)} & \textbf{0.99 \scriptsize{(0.00)}} & \textbf{0.68 \scriptsize{(0.09)}} & \textbf{0.98 \scriptsize{(0.00)}} \\
InternVL3 78B & $\checkmark$ & 0.96 \scriptsize{(0.04)} & 0.99 \scriptsize{(0.00)} & 0.76 \scriptsize{(0.09)} & \textbf{0.99 \scriptsize{(0.00)}} & 0.64 \scriptsize{(0.10)} & \textbf{0.98 \scriptsize{(0.00)}} \\
InternVL3.5 38B & $\times$ & \textbf{1.00 \scriptsize{(0.00)}} & \textbf{1.00 \scriptsize{(0.00)}} & \textbf{1.00 \scriptsize{(0.00)}} & \textbf{0.99 \scriptsize{(0.00)}} & 0.64 \scriptsize{(0.10)} & \textbf{0.98 \scriptsize{(0.00)}} \\
InternVL3.5 38B & $\checkmark$ & \textbf{1.00 \scriptsize{(0.00)}} & 0.99 \scriptsize{(0.00)} & 0.84 \scriptsize{(0.07)} & \textbf{0.99 \scriptsize{(0.00)}} & 0.48 \scriptsize{(0.10)} & 0.96 \scriptsize{(0.00)} \\
Qwen2.5\text{-}VL 72B & $\times$ & 0.60 \scriptsize{(0.10)} & 0.98 \scriptsize{(0.00)} & 0.20 \scriptsize{(0.08)} & 0.98 \scriptsize{(0.00)} & 0.00 \scriptsize{(0.00)} & 0.96 \scriptsize{(0.00)} \\
Qwen2.5\text{-}VL 72B & $\checkmark$ & 0.52 \scriptsize{(0.10)} & 0.97 \scriptsize{(0.00)} & 0.16 \scriptsize{(0.07)} & 0.98 \scriptsize{(0.00)} & 0.00 \scriptsize{(0.00)} & 0.96 \scriptsize{(0.00)} \\
Qwen3\text{-}VL 32B & $\times$ & 0.64 \scriptsize{(0.10)} & 0.99 \scriptsize{(0.00)} & 0.32 \scriptsize{(0.09)} & 0.98 \scriptsize{(0.00)} & 0.04 \scriptsize{(0.04)} & 0.97 \scriptsize{(0.00)} \\
Qwen3\text{-}VL 32B & $\checkmark$ & 0.56 \scriptsize{(0.10)} & 0.98 \scriptsize{(0.00)} & 0.48 \scriptsize{(0.10)} & 0.98 \scriptsize{(0.00)} & 0.08 \scriptsize{(0.05)} & 0.97 \scriptsize{(0.00)} \\
\bottomrule
\end{tabular}
\end{table*}

\begin{table*}[h]
\caption{\textbf{Full Results for Ground + Mem and Ground + Mem + CoT on ViPlan-HH.} Success rate indicates the fraction of problems solved, accuracy is the fraction of correct predictions. The highest value in each column is bolded.}
\label{tab:ViPLan-HH_history_full}
\centering
  \begin{tabular}{l c c c c c c c}
\toprule
\multirow{2}{*}{\textbf{Model}} & \multirow{2}{*}{\textbf{CoT}} & \multicolumn{2}{c}{\textbf{Simple}} & \multicolumn{2}{c}{\textbf{Medium}} & \multicolumn{2}{c}{\textbf{Hard}} \\
 &  & \textbf{Success} & \textbf{Accuracy} & \textbf{Success} & \textbf{Accuracy} & \textbf{Success} & \textbf{Accuracy} \\
\midrule
Gemma\text{-}3 27B & $\times$ & 0.00 \scriptsize{(0.00)} & 0.69 \scriptsize{(0.02)} & 0.00 \scriptsize{(0.00)} & \textbf{0.79 \scriptsize{(0.00)}} & 0.00 \scriptsize{(0.00)} & 0.76 \scriptsize{(0.00)} \\
Gemma\text{-}3 27B & $\checkmark$ & 0.12 \scriptsize{(0.06)} & 0.62 \scriptsize{(0.01)} & 0.00 \scriptsize{(0.00)} & 0.67 \scriptsize{(0.00)} & 0.00 \scriptsize{(0.00)} & 0.66 \scriptsize{(0.00)} \\
InternVL3 78B & $\times$ & 0.12 \scriptsize{(0.06)} & 0.66 \scriptsize{(0.01)} & \textbf{0.04 \scriptsize{(0.04)}} & 0.63 \scriptsize{(0.01)} & 0.04 \scriptsize{(0.04)} & 0.59 \scriptsize{(0.00)} \\
InternVL3 78B & $\checkmark$ & 0.16 \scriptsize{(0.07)} & 0.64 \scriptsize{(0.01)} & 0.00 \scriptsize{(0.00)} & 0.71 \scriptsize{(0.00)} & 0.00 \scriptsize{(0.00)} & 0.67 \scriptsize{(0.00)} \\
InternVL3.5 38B & $\times$ & \textbf{0.20 \scriptsize{(0.08)}} & 0.59 \scriptsize{(0.01)} & 0.00 \scriptsize{(0.00)} & 0.62 \scriptsize{(0.01)} & 0.04 \scriptsize{(0.04)} & 0.71 \scriptsize{(0.00)} \\
InternVL3.5 38B & $\checkmark$ & 0.12 \scriptsize{(0.06)} & 0.64 \scriptsize{(0.01)} & 0.00 \scriptsize{(0.00)} & 0.68 \scriptsize{(0.01)} & \textbf{0.08 \scriptsize{(0.05)}} & 0.68 \scriptsize{(0.00)} \\
Qwen2.5\text{-}VL 72B & $\times$ & 0.04 \scriptsize{(0.04)} & 0.66 \scriptsize{(0.01)} & 0.00 \scriptsize{(0.00)} & \textbf{0.79 \scriptsize{(0.00)}} & 0.00 \scriptsize{(0.00)} & \textbf{0.77 \scriptsize{(0.00)}} \\
Qwen2.5\text{-}VL 72B & $\checkmark$ & 0.04 \scriptsize{(0.04)} & \textbf{0.72 \scriptsize{(0.01)}} & 0.00 \scriptsize{(0.00)} & 0.77 \scriptsize{(0.00)} & 0.00 \scriptsize{(0.00)} & 0.71 \scriptsize{(0.00)} \\
Qwen3\text{-}VL 32B & $\times$ & \textbf{0.20 \scriptsize{(0.08)}} & \textbf{0.72 \scriptsize{(0.01)}} & \textbf{0.04 \scriptsize{(0.04)}} & 0.74 \scriptsize{(0.01)} & 0.00 \scriptsize{(0.00)} & 0.74 \scriptsize{(0.00)} \\
Qwen3\text{-}VL 32B & $\checkmark$ & 0.12 \scriptsize{(0.06)} & 0.68 \scriptsize{(0.01)} & 0.00 \scriptsize{(0.00)} & 0.70 \scriptsize{(0.01)} & 0.04 \scriptsize{(0.04)} & 0.64 \scriptsize{(0.00)} \\
\bottomrule
\end{tabular}
\end{table*}

\begin{table*}[h]
\caption{\textbf{Full Results for Plan and Plan + CoT. on ViPlan-BW.} Success rate indicates the fraction of problems solved. The result is an average over the 25 problems in each split. Standard error of the mean is reported in parenthesis. The highest value in each column is bolded.}
\label{tab:ViPLan-BW_vila_full}
\centering
  \begin{tabular}{l c c c c}
\toprule
\textbf{Model} & \textbf{CoT} & \textbf{Simple} & \textbf{Medium} & \textbf{Hard} \\
\midrule
AyaVision 32B & $\times$ & 0.00 \scriptsize{(0.00)} & 0.00 \scriptsize{(0.00)} & 0.00 \scriptsize{(0.00)} \\
AyaVision 32B & $\checkmark$ & 0.00 \scriptsize{(0.00)} & 0.00 \scriptsize{(0.00)} & 0.00 \scriptsize{(0.00)} \\
AyaVision 8B & $\times$ & 0.00 \scriptsize{(0.00)} & 0.00 \scriptsize{(0.00)} & 0.00 \scriptsize{(0.00)} \\
AyaVision 8B & $\checkmark$ & 0.04 \scriptsize{(0.04)} & 0.00 \scriptsize{(0.00)} & 0.00 \scriptsize{(0.00)} \\
Cosmos\text{-}Reason2 8B & $\times$ & 0.00 \scriptsize{(0.00)} & 0.00 \scriptsize{(0.00)} & 0.00 \scriptsize{(0.00)} \\
Cosmos\text{-}Reason2 8B & $\checkmark$ & 0.00 \scriptsize{(0.00)} & 0.00 \scriptsize{(0.00)} & 0.00 \scriptsize{(0.00)} \\
DeepSeek\text{-}VL2 & $\times$ & 0.00 \scriptsize{(0.00)} & 0.00 \scriptsize{(0.00)} & 0.00 \scriptsize{(0.00)} \\
DeepSeek\text{-}VL2 & $\checkmark$ & 0.00 \scriptsize{(0.00)} & 0.00 \scriptsize{(0.00)} & 0.00 \scriptsize{(0.00)} \\
GPT\text{-}4.1 & $\times$ & 0.24 \scriptsize{(0.09)} & 0.04 \scriptsize{(0.04)} & 0.04 \scriptsize{(0.04)} \\
GPT\text{-}4.1 & $\checkmark$ & \textbf{0.84 \scriptsize{(0.07)}} & \textbf{0.48 \scriptsize{(0.10)}} & \textbf{0.12 \scriptsize{(0.06)}} \\
GPT\text{-}4.1 Nano & $\times$ & 0.04 \scriptsize{(0.04)} & 0.00 \scriptsize{(0.00)} & 0.00 \scriptsize{(0.00)} \\
GPT\text{-}4.1 Nano & $\checkmark$ & 0.24 \scriptsize{(0.09)} & 0.00 \scriptsize{(0.00)} & 0.00 \scriptsize{(0.00)} \\
GPT\text{-}5.2 & $\times$ & 0.32 \scriptsize{(0.09)} & 0.08 \scriptsize{(0.05)} & 0.00 \scriptsize{(0.00)} \\
Gemma\text{-}3 12B & $\times$ & 0.16 \scriptsize{(0.07)} & 0.00 \scriptsize{(0.00)} & 0.00 \scriptsize{(0.00)} \\
Gemma\text{-}3 12B & $\checkmark$ & 0.28 \scriptsize{(0.09)} & 0.00 \scriptsize{(0.00)} & 0.00 \scriptsize{(0.00)} \\
Gemma\text{-}3 27B & $\times$ & 0.24 \scriptsize{(0.09)} & 0.04 \scriptsize{(0.04)} & 0.00 \scriptsize{(0.00)} \\
Gemma\text{-}3 27B & $\checkmark$ & 0.20 \scriptsize{(0.08)} & 0.12 \scriptsize{(0.06)} & 0.04 \scriptsize{(0.04)} \\
InternVL3 78B & $\times$ & 0.08 \scriptsize{(0.05)} & 0.00 \scriptsize{(0.00)} & 0.00 \scriptsize{(0.00)} \\
InternVL3 78B & $\checkmark$ & 0.12 \scriptsize{(0.06)} & 0.00 \scriptsize{(0.00)} & 0.00 \scriptsize{(0.00)} \\
InternVL3 8B & $\times$ & 0.04 \scriptsize{(0.04)} & 0.00 \scriptsize{(0.00)} & 0.00 \scriptsize{(0.00)} \\
InternVL3 8B & $\checkmark$ & 0.04 \scriptsize{(0.04)} & 0.00 \scriptsize{(0.00)} & 0.00 \scriptsize{(0.00)} \\
InternVL3.5 30B A3B & $\times$ & 0.00 \scriptsize{(0.00)} & 0.00 \scriptsize{(0.00)} & 0.00 \scriptsize{(0.00)} \\
InternVL3.5 30B A3B & $\checkmark$ & 0.08 \scriptsize{(0.05)} & 0.00 \scriptsize{(0.00)} & 0.00 \scriptsize{(0.00)} \\
InternVL3.5 38B & $\times$ & 0.40 \scriptsize{(0.10)} & 0.08 \scriptsize{(0.05)} & 0.00 \scriptsize{(0.00)} \\
InternVL3.5 38B & $\checkmark$ & 0.12 \scriptsize{(0.06)} & 0.00 \scriptsize{(0.00)} & 0.00 \scriptsize{(0.00)} \\
InternVL3.5 8B & $\times$ & 0.00 \scriptsize{(0.00)} & 0.00 \scriptsize{(0.00)} & 0.00 \scriptsize{(0.00)} \\
InternVL3.5 8B & $\checkmark$ & 0.00 \scriptsize{(0.00)} & 0.00 \scriptsize{(0.00)} & 0.00 \scriptsize{(0.00)} \\
LLaVA\text{-}Onevision 72B & $\times$ & 0.00 \scriptsize{(0.00)} & 0.00 \scriptsize{(0.00)} & 0.00 \scriptsize{(0.00)} \\
LLaVA\text{-}Onevision 72B & $\checkmark$ & 0.00 \scriptsize{(0.00)} & 0.00 \scriptsize{(0.00)} & 0.00 \scriptsize{(0.00)} \\
LLaVA\text{-}Onevision 7B & $\times$ & 0.00 \scriptsize{(0.00)} & 0.00 \scriptsize{(0.00)} & 0.00 \scriptsize{(0.00)} \\
LLaVA\text{-}Onevision 7B & $\checkmark$ & 0.00 \scriptsize{(0.00)} & 0.00 \scriptsize{(0.00)} & 0.00 \scriptsize{(0.00)} \\
Mistral\text{-}Small\text{-}3.1 24B & $\times$ & 0.16 \scriptsize{(0.07)} & 0.00 \scriptsize{(0.00)} & 0.00 \scriptsize{(0.00)} \\
Mistral\text{-}Small\text{-}3.1 24B & $\checkmark$ & 0.00 \scriptsize{(0.00)} & 0.00 \scriptsize{(0.00)} & 0.00 \scriptsize{(0.00)} \\
Molmo 7B & $\times$ & 0.00 \scriptsize{(0.00)} & 0.00 \scriptsize{(0.00)} & 0.00 \scriptsize{(0.00)} \\
Molmo 7B & $\checkmark$ & 0.00 \scriptsize{(0.00)} & 0.00 \scriptsize{(0.00)} & 0.00 \scriptsize{(0.00)} \\
Phi\text{-}4 Multimodal & $\times$ & 0.00 \scriptsize{(0.00)} & 0.00 \scriptsize{(0.00)} & 0.00 \scriptsize{(0.00)} \\
Phi\text{-}4 Multimodal & $\checkmark$ & 0.00 \scriptsize{(0.00)} & 0.00 \scriptsize{(0.00)} & 0.00 \scriptsize{(0.00)} \\
Qwen2.5\text{-}VL 72B & $\times$ & 0.08 \scriptsize{(0.05)} & 0.04 \scriptsize{(0.04)} & 0.00 \scriptsize{(0.00)} \\
Qwen2.5\text{-}VL 72B & $\checkmark$ & 0.16 \scriptsize{(0.07)} & 0.04 \scriptsize{(0.04)} & 0.00 \scriptsize{(0.00)} \\
Qwen2.5\text{-}VL 7B & $\times$ & 0.08 \scriptsize{(0.05)} & 0.00 \scriptsize{(0.00)} & 0.00 \scriptsize{(0.00)} \\
Qwen2.5\text{-}VL 7B & $\checkmark$ & 0.04 \scriptsize{(0.04)} & 0.00 \scriptsize{(0.00)} & 0.00 \scriptsize{(0.00)} \\
Qwen3\text{-}VL 30B A3B & $\times$ & 0.00 \scriptsize{(0.00)} & 0.00 \scriptsize{(0.00)} & 0.00 \scriptsize{(0.00)} \\
Qwen3\text{-}VL 30B A3B & $\checkmark$ & 0.20 \scriptsize{(0.08)} & 0.00 \scriptsize{(0.00)} & 0.00 \scriptsize{(0.00)} \\
Qwen3\text{-}VL 32B & $\times$ & 0.04 \scriptsize{(0.04)} & 0.00 \scriptsize{(0.00)} & 0.00 \scriptsize{(0.00)} \\
Qwen3\text{-}VL 32B & $\checkmark$ & 0.48 \scriptsize{(0.10)} & 0.20 \scriptsize{(0.08)} & 0.04 \scriptsize{(0.04)} \\
Qwen3\text{-}VL 8B & $\times$ & 0.04 \scriptsize{(0.04)} & 0.00 \scriptsize{(0.00)} & 0.00 \scriptsize{(0.00)} \\
Qwen3\text{-}VL 8B & $\checkmark$ & 0.00 \scriptsize{(0.00)} & 0.00 \scriptsize{(0.00)} & 0.00 \scriptsize{(0.00)} \\
\bottomrule
\end{tabular}
\end{table*}

\begin{table*}[h]
\caption{\textbf{Full Results for Ground and Ground + CoT on ViPlan-HH.} Success rate indicates the fraction of problems solved, accuracy is the fraction of correct predictions for a single predicate. The result is an average over the 25 problems in each split. Standard error of the mean is reported in parenthesis. The highest value in each column is bolded.}
\label{tab:ViPLan-HH_pred_full}
\centering
\resizebox{\linewidth}{!}{\scriptsize
  \begin{tabular}{l c c c c c c c}
\toprule
\multirow{2}{*}{\textbf{Model}} & \multirow{2}{*}{\textbf{CoT}} & \multicolumn{2}{c}{\textbf{Simple}} & \multicolumn{2}{c}{\textbf{Medium}} & \multicolumn{2}{c}{\textbf{Hard}} \\
 &  & \textbf{Success} & \textbf{Accuracy} & \textbf{Success} & \textbf{Accuracy} & \textbf{Success} & \textbf{Accuracy} \\
\midrule
AyaVision 32B & $\times$ & 0.04 \scriptsize{(0.04)} & 0.61 \scriptsize{(0.02)} & 0.00 \scriptsize{(0.00)} & 0.65 \scriptsize{(0.01)} & 0.00 \scriptsize{(0.00)} & 0.65 \scriptsize{(0.01)} \\
AyaVision 32B & $\checkmark$ & 0.04 \scriptsize{(0.04)} & 0.72 \scriptsize{(0.01)} & 0.00 \scriptsize{(0.00)} & 0.75 \scriptsize{(0.00)} & 0.00 \scriptsize{(0.00)} & 0.74 \scriptsize{(0.01)} \\
AyaVision 8B & $\times$ & 0.00 \scriptsize{(0.00)} & 0.46 \scriptsize{(0.02)} & 0.00 \scriptsize{(0.00)} & 0.44 \scriptsize{(0.01)} & 0.00 \scriptsize{(0.00)} & 0.44 \scriptsize{(0.01)} \\
AyaVision 8B & $\checkmark$ & 0.00 \scriptsize{(0.00)} & 0.67 \scriptsize{(0.01)} & 0.00 \scriptsize{(0.00)} & 0.70 \scriptsize{(0.01)} & 0.00 \scriptsize{(0.00)} & 0.69 \scriptsize{(0.00)} \\
Cosmos\text{-}Reason2 8B & $\times$ & 0.16 \scriptsize{(0.07)} & 0.57 \scriptsize{(0.01)} & 0.00 \scriptsize{(0.00)} & 0.65 \scriptsize{(0.01)} & 0.00 \scriptsize{(0.00)} & 0.50 \scriptsize{(0.01)} \\
Cosmos\text{-}Reason2 8B & $\checkmark$ & 0.00 \scriptsize{(0.00)} & 0.69 \scriptsize{(0.01)} & 0.00 \scriptsize{(0.00)} & 0.72 \scriptsize{(0.01)} & 0.00 \scriptsize{(0.00)} & 0.73 \scriptsize{(0.01)} \\
DeepSeek\text{-}VL2 & $\times$ & 0.12 \scriptsize{(0.06)} & 0.71 \scriptsize{(0.01)} & 0.00 \scriptsize{(0.00)} & 0.75 \scriptsize{(0.00)} & 0.00 \scriptsize{(0.00)} & 0.75 \scriptsize{(0.01)} \\
DeepSeek\text{-}VL2 & $\checkmark$ & 0.00 \scriptsize{(0.00)} & 0.69 \scriptsize{(0.01)} & 0.00 \scriptsize{(0.00)} & 0.70 \scriptsize{(0.00)} & 0.00 \scriptsize{(0.00)} & 0.68 \scriptsize{(0.01)} \\
GPT\text{-}4.1 & $\times$ & 0.16 \scriptsize{(0.07)} & 0.68 \scriptsize{(0.01)} & 0.00 \scriptsize{(0.00)} & 0.74 \scriptsize{(0.01)} & 0.00 \scriptsize{(0.00)} & 0.70 \scriptsize{(0.01)} \\
GPT\text{-}4.1 & $\checkmark$ & 0.20 \scriptsize{(0.08)} & 0.69 \scriptsize{(0.01)} & 0.04 \scriptsize{(0.04)} & 0.72 \scriptsize{(0.01)} & 0.00 \scriptsize{(0.00)} & 0.71 \scriptsize{(0.00)} \\
GPT\text{-}4.1 Nano & $\times$ & 0.04 \scriptsize{(0.04)} & 0.54 \scriptsize{(0.01)} & 0.00 \scriptsize{(0.00)} & 0.60 \scriptsize{(0.01)} & 0.00 \scriptsize{(0.00)} & 0.53 \scriptsize{(0.01)} \\
GPT\text{-}4.1 Nano & $\checkmark$ & 0.04 \scriptsize{(0.04)} & 0.67 \scriptsize{(0.01)} & 0.00 \scriptsize{(0.00)} & 0.72 \scriptsize{(0.01)} & 0.00 \scriptsize{(0.00)} & 0.66 \scriptsize{(0.01)} \\
GPT\text{-}5.2 & $\times$ & 0.16 \scriptsize{(0.07)} & 0.65 \scriptsize{(0.01)} & 0.00 \scriptsize{(0.00)} & 0.76 \scriptsize{(0.00)} & 0.00 \scriptsize{(0.00)} & 0.82 \scriptsize{(0.00)} \\
Gemma\text{-}3 12B & $\times$ & 0.16 \scriptsize{(0.07)} & 0.52 \scriptsize{(0.02)} & 0.00 \scriptsize{(0.00)} & 0.68 \scriptsize{(0.01)} & 0.00 \scriptsize{(0.00)} & 0.59 \scriptsize{(0.01)} \\
Gemma\text{-}3 12B & $\checkmark$ & 0.12 \scriptsize{(0.06)} & 0.65 \scriptsize{(0.01)} & 0.00 \scriptsize{(0.00)} & 0.58 \scriptsize{(0.01)} & 0.00 \scriptsize{(0.00)} & 0.60 \scriptsize{(0.01)} \\
Gemma\text{-}3 27B & $\times$ & 0.08 \scriptsize{(0.05)} & 0.67 \scriptsize{(0.01)} & 0.00 \scriptsize{(0.00)} & 0.77 \scriptsize{(0.00)} & 0.08 \scriptsize{(0.05)} & 0.75 \scriptsize{(0.00)} \\
Gemma\text{-}3 27B & $\checkmark$ & 0.12 \scriptsize{(0.06)} & 0.70 \scriptsize{(0.01)} & 0.00 \scriptsize{(0.00)} & 0.68 \scriptsize{(0.01)} & 0.00 \scriptsize{(0.00)} & 0.66 \scriptsize{(0.00)} \\
InternVL3 78B & $\times$ & 0.12 \scriptsize{(0.06)} & 0.71 \scriptsize{(0.01)} & 0.00 \scriptsize{(0.00)} & 0.68 \scriptsize{(0.01)} & 0.00 \scriptsize{(0.00)} & 0.65 \scriptsize{(0.01)} \\
InternVL3 78B & $\checkmark$ & \textbf{0.28 \scriptsize{(0.09)}} & 0.66 \scriptsize{(0.01)} & 0.00 \scriptsize{(0.00)} & 0.69 \scriptsize{(0.01)} & 0.04 \scriptsize{(0.04)} & 0.70 \scriptsize{(0.01)} \\
InternVL3 8B & $\times$ & 0.00 \scriptsize{(0.00)} & 0.73 \scriptsize{(0.01)} & 0.00 \scriptsize{(0.00)} & 0.78 \scriptsize{(0.00)} & 0.04 \scriptsize{(0.04)} & 0.78 \scriptsize{(0.01)} \\
InternVL3 8B & $\checkmark$ & 0.08 \scriptsize{(0.05)} & 0.62 \scriptsize{(0.01)} & 0.00 \scriptsize{(0.00)} & 0.59 \scriptsize{(0.01)} & 0.00 \scriptsize{(0.00)} & 0.56 \scriptsize{(0.01)} \\
InternVL3.5 30B A3B & $\times$ & 0.16 \scriptsize{(0.07)} & 0.68 \scriptsize{(0.01)} & 0.00 \scriptsize{(0.00)} & 0.73 \scriptsize{(0.01)} & 0.00 \scriptsize{(0.00)} & 0.68 \scriptsize{(0.01)} \\
InternVL3.5 30B A3B & $\checkmark$ & 0.04 \scriptsize{(0.04)} & 0.72 \scriptsize{(0.01)} & 0.00 \scriptsize{(0.00)} & 0.75 \scriptsize{(0.01)} & 0.00 \scriptsize{(0.00)} & 0.76 \scriptsize{(0.01)} \\
InternVL3.5 38B & $\times$ & 0.08 \scriptsize{(0.05)} & 0.52 \scriptsize{(0.01)} & 0.04 \scriptsize{(0.04)} & 0.55 \scriptsize{(0.01)} & 0.00 \scriptsize{(0.00)} & 0.46 \scriptsize{(0.01)} \\
InternVL3.5 38B & $\checkmark$ & 0.08 \scriptsize{(0.05)} & 0.68 \scriptsize{(0.01)} & 0.00 \scriptsize{(0.00)} & 0.66 \scriptsize{(0.01)} & \textbf{0.12 \scriptsize{(0.06)}} & 0.69 \scriptsize{(0.00)} \\
InternVL3.5 8B & $\times$ & 0.08 \scriptsize{(0.05)} & 0.69 \scriptsize{(0.01)} & 0.00 \scriptsize{(0.00)} & 0.72 \scriptsize{(0.01)} & 0.00 \scriptsize{(0.00)} & 0.68 \scriptsize{(0.01)} \\
InternVL3.5 8B & $\checkmark$ & 0.04 \scriptsize{(0.04)} & 0.65 \scriptsize{(0.01)} & 0.00 \scriptsize{(0.00)} & 0.70 \scriptsize{(0.01)} & 0.00 \scriptsize{(0.00)} & 0.71 \scriptsize{(0.01)} \\
LLaVA\text{-}Onevision 72B & $\times$ & 0.00 \scriptsize{(0.00)} & 0.64 \scriptsize{(0.02)} & 0.00 \scriptsize{(0.00)} & 0.80 \scriptsize{(0.01)} & 0.00 \scriptsize{(0.00)} & 0.75 \scriptsize{(0.01)} \\
LLaVA\text{-}Onevision 72B & $\checkmark$ & 0.00 \scriptsize{(0.00)} & 0.69 \scriptsize{(0.02)} & 0.04 \scriptsize{(0.04)} & 0.79 \scriptsize{(0.01)} & 0.00 \scriptsize{(0.00)} & 0.78 \scriptsize{(0.02)} \\
LLaVA\text{-}Onevision 7B & $\times$ & 0.04 \scriptsize{(0.04)} & 0.59 \scriptsize{(0.01)} & 0.00 \scriptsize{(0.00)} & 0.68 \scriptsize{(0.01)} & 0.00 \scriptsize{(0.00)} & 0.69 \scriptsize{(0.01)} \\
LLaVA\text{-}Onevision 7B & $\checkmark$ & 0.04 \scriptsize{(0.04)} & 0.58 \scriptsize{(0.01)} & 0.00 \scriptsize{(0.00)} & 0.65 \scriptsize{(0.01)} & 0.00 \scriptsize{(0.00)} & 0.70 \scriptsize{(0.01)} \\
Mistral\text{-}Small\text{-}3.1 24B & $\times$ & 0.00 \scriptsize{(0.00)} & 0.69 \scriptsize{(0.01)} & 0.00 \scriptsize{(0.00)} & \textbf{0.83 \scriptsize{(0.00)}} & 0.00 \scriptsize{(0.00)} & \textbf{0.83 \scriptsize{(0.00)}} \\
Mistral\text{-}Small\text{-}3.1 24B & $\checkmark$ & 0.04 \scriptsize{(0.04)} & 0.60 \scriptsize{(0.02)} & 0.00 \scriptsize{(0.00)} & 0.64 \scriptsize{(0.01)} & 0.00 \scriptsize{(0.00)} & 0.67 \scriptsize{(0.01)} \\
Molmo 7B & $\times$ & 0.00 \scriptsize{(0.00)} & 0.60 \scriptsize{(0.02)} & 0.04 \scriptsize{(0.04)} & 0.65 \scriptsize{(0.01)} & 0.00 \scriptsize{(0.00)} & 0.67 \scriptsize{(0.00)} \\
Molmo 7B & $\checkmark$ & 0.00 \scriptsize{(0.00)} & 0.68 \scriptsize{(0.01)} & 0.00 \scriptsize{(0.00)} & 0.71 \scriptsize{(0.01)} & 0.00 \scriptsize{(0.00)} & 0.78 \scriptsize{(0.00)} \\
Phi\text{-}4 Multimodal & $\times$ & 0.04 \scriptsize{(0.04)} & 0.31 \scriptsize{(0.02)} & 0.00 \scriptsize{(0.00)} & 0.22 \scriptsize{(0.01)} & 0.00 \scriptsize{(0.00)} & 0.33 \scriptsize{(0.01)} \\
Phi\text{-}4 Multimodal & $\checkmark$ & 0.00 \scriptsize{(0.00)} & 0.05 \scriptsize{(0.03)} & 0.00 \scriptsize{(0.00)} & 0.02 \scriptsize{(0.00)} & 0.00 \scriptsize{(0.00)} & 0.17 \scriptsize{(0.02)} \\
Qwen2.5\text{-}VL 72B & $\times$ & 0.04 \scriptsize{(0.04)} & 0.72 \scriptsize{(0.01)} & 0.00 \scriptsize{(0.00)} & 0.80 \scriptsize{(0.00)} & 0.00 \scriptsize{(0.00)} & 0.78 \scriptsize{(0.00)} \\
Qwen2.5\text{-}VL 72B & $\checkmark$ & 0.00 \scriptsize{(0.00)} & 0.70 \scriptsize{(0.01)} & 0.00 \scriptsize{(0.00)} & 0.75 \scriptsize{(0.01)} & 0.00 \scriptsize{(0.00)} & 0.76 \scriptsize{(0.00)} \\
Qwen2.5\text{-}VL 7B & $\times$ & 0.00 \scriptsize{(0.00)} & 0.73 \scriptsize{(0.01)} & 0.04 \scriptsize{(0.04)} & 0.81 \scriptsize{(0.00)} & 0.00 \scriptsize{(0.00)} & 0.82 \scriptsize{(0.00)} \\
Qwen2.5\text{-}VL 7B & $\checkmark$ & 0.04 \scriptsize{(0.04)} & 0.68 \scriptsize{(0.01)} & 0.00 \scriptsize{(0.00)} & 0.73 \scriptsize{(0.00)} & 0.04 \scriptsize{(0.04)} & 0.62 \scriptsize{(0.00)} \\
Qwen3\text{-}VL 30B A3B & $\times$ & 0.20 \scriptsize{(0.08)} & 0.75 \scriptsize{(0.01)} & 0.00 \scriptsize{(0.00)} & 0.72 \scriptsize{(0.01)} & 0.00 \scriptsize{(0.00)} & 0.71 \scriptsize{(0.01)} \\
Qwen3\text{-}VL 30B A3B & $\checkmark$ & 0.08 \scriptsize{(0.05)} & 0.68 \scriptsize{(0.01)} & 0.00 \scriptsize{(0.00)} & 0.74 \scriptsize{(0.01)} & 0.00 \scriptsize{(0.00)} & 0.73 \scriptsize{(0.00)} \\
Qwen3\text{-}VL 32B & $\times$ & 0.08 \scriptsize{(0.05)} & 0.71 \scriptsize{(0.01)} & 0.00 \scriptsize{(0.00)} & 0.71 \scriptsize{(0.01)} & 0.04 \scriptsize{(0.04)} & 0.66 \scriptsize{(0.01)} \\
Qwen3\text{-}VL 32B & $\checkmark$ & 0.12 \scriptsize{(0.06)} & 0.66 \scriptsize{(0.01)} & \textbf{0.08 \scriptsize{(0.05)}} & 0.71 \scriptsize{(0.01)} & 0.04 \scriptsize{(0.04)} & 0.69 \scriptsize{(0.00)} \\
Qwen3\text{-}VL 8B & $\times$ & 0.12 \scriptsize{(0.06)} & 0.66 \scriptsize{(0.01)} & 0.00 \scriptsize{(0.00)} & 0.73 \scriptsize{(0.01)} & 0.04 \scriptsize{(0.04)} & 0.72 \scriptsize{(0.01)} \\
Qwen3\text{-}VL 8B & $\checkmark$ & 0.04 \scriptsize{(0.04)} & \textbf{0.76 \scriptsize{(0.01)}} & 0.00 \scriptsize{(0.00)} & 0.72 \scriptsize{(0.01)} & 0.00 \scriptsize{(0.00)} & 0.74 \scriptsize{(0.00)} \\
\bottomrule
\end{tabular}}
\end{table*}

\begin{table*}[h]
\caption{\textbf{Full Results for Plan and Plan + CoT on ViPlan-HH.} Success rate indicates the fraction of problems solved. The result is an average over the 25 problems in each split. Standard error of the mean is reported in parenthesis. The highest value in each column is bolded.}
\label{tab:ViPLan-HH_vila_full}
\centering
  \begin{tabular}{l c c c c}
\toprule
\textbf{Model} & \textbf{CoT} & \textbf{Simple} & \textbf{Medium} & \textbf{Hard} \\
\midrule
AyaVision 32B & $\times$ & 0.36 \scriptsize{(0.10)} & 0.00 \scriptsize{(0.00)} & 0.00 \scriptsize{(0.00)} \\
AyaVision 32B & $\checkmark$ & 0.04 \scriptsize{(0.04)} & 0.00 \scriptsize{(0.00)} & 0.00 \scriptsize{(0.00)} \\
AyaVision 8B & $\times$ & 0.04 \scriptsize{(0.04)} & 0.00 \scriptsize{(0.00)} & 0.00 \scriptsize{(0.00)} \\
AyaVision 8B & $\checkmark$ & 0.00 \scriptsize{(0.00)} & 0.00 \scriptsize{(0.00)} & 0.04 \scriptsize{(0.04)} \\
Cosmos\text{-}Reason2 8B & $\times$ & 0.52 \scriptsize{(0.10)} & 0.24 \scriptsize{(0.09)} & 0.04 \scriptsize{(0.04)} \\
Cosmos\text{-}Reason2 8B & $\checkmark$ & 0.44 \scriptsize{(0.10)} & 0.12 \scriptsize{(0.06)} & 0.24 \scriptsize{(0.09)} \\
DeepSeek\text{-}VL2 & $\times$ & 0.00 \scriptsize{(0.00)} & 0.00 \scriptsize{(0.00)} & 0.00 \scriptsize{(0.00)} \\
DeepSeek\text{-}VL2 & $\checkmark$ & 0.00 \scriptsize{(0.00)} & 0.00 \scriptsize{(0.00)} & 0.00 \scriptsize{(0.00)} \\
GPT\text{-}4.1 & $\times$ & 0.88 \scriptsize{(0.06)} & 0.52 \scriptsize{(0.10)} & \textbf{0.36 \scriptsize{(0.10)}} \\
GPT\text{-}4.1 & $\checkmark$ & 0.60 \scriptsize{(0.10)} & 0.44 \scriptsize{(0.10)} & 0.32 \scriptsize{(0.09)} \\
GPT\text{-}4.1 Nano & $\times$ & 0.08 \scriptsize{(0.05)} & 0.04 \scriptsize{(0.04)} & 0.00 \scriptsize{(0.00)} \\
GPT\text{-}4.1 Nano & $\checkmark$ & 0.16 \scriptsize{(0.07)} & 0.04 \scriptsize{(0.04)} & 0.04 \scriptsize{(0.04)} \\
GPT\text{-}5.2 & $\times$ & 0.76 \scriptsize{(0.09)} & \textbf{0.72 \scriptsize{(0.09)}} & 0.16 \scriptsize{(0.07)} \\
Gemma\text{-}3 12B & $\times$ & 0.64 \scriptsize{(0.10)} & 0.04 \scriptsize{(0.04)} & 0.20 \scriptsize{(0.08)} \\
Gemma\text{-}3 12B & $\checkmark$ & 0.48 \scriptsize{(0.10)} & 0.00 \scriptsize{(0.00)} & 0.00 \scriptsize{(0.00)} \\
Gemma\text{-}3 27B & $\times$ & \textbf{0.92 \scriptsize{(0.05)}} & 0.32 \scriptsize{(0.09)} & 0.28 \scriptsize{(0.09)} \\
Gemma\text{-}3 27B & $\checkmark$ & 0.76 \scriptsize{(0.09)} & 0.16 \scriptsize{(0.07)} & 0.04 \scriptsize{(0.04)} \\
InternVL3 78B & $\times$ & 0.44 \scriptsize{(0.10)} & 0.44 \scriptsize{(0.10)} & 0.20 \scriptsize{(0.08)} \\
InternVL3 78B & $\checkmark$ & 0.36 \scriptsize{(0.10)} & 0.04 \scriptsize{(0.04)} & 0.08 \scriptsize{(0.05)} \\
InternVL3 8B & $\times$ & 0.16 \scriptsize{(0.07)} & 0.04 \scriptsize{(0.04)} & 0.04 \scriptsize{(0.04)} \\
InternVL3 8B & $\checkmark$ & 0.44 \scriptsize{(0.10)} & 0.00 \scriptsize{(0.00)} & 0.00 \scriptsize{(0.00)} \\
InternVL3.5 30B A3B & $\times$ & 0.32 \scriptsize{(0.09)} & 0.32 \scriptsize{(0.09)} & 0.12 \scriptsize{(0.06)} \\
InternVL3.5 30B A3B & $\checkmark$ & 0.12 \scriptsize{(0.06)} & 0.00 \scriptsize{(0.00)} & 0.00 \scriptsize{(0.00)} \\
InternVL3.5 38B & $\times$ & 0.44 \scriptsize{(0.10)} & 0.08 \scriptsize{(0.05)} & 0.20 \scriptsize{(0.08)} \\
InternVL3.5 38B & $\checkmark$ & 0.32 \scriptsize{(0.09)} & 0.12 \scriptsize{(0.06)} & 0.00 \scriptsize{(0.00)} \\
InternVL3.5 8B & $\times$ & 0.48 \scriptsize{(0.10)} & 0.08 \scriptsize{(0.05)} & 0.04 \scriptsize{(0.04)} \\
InternVL3.5 8B & $\checkmark$ & 0.36 \scriptsize{(0.10)} & 0.00 \scriptsize{(0.00)} & 0.00 \scriptsize{(0.00)} \\
LLaVA\text{-}Onevision 72B & $\times$ & 0.00 \scriptsize{(0.00)} & 0.00 \scriptsize{(0.00)} & 0.00 \scriptsize{(0.00)} \\
LLaVA\text{-}Onevision 72B & $\checkmark$ & 0.00 \scriptsize{(0.00)} & 0.00 \scriptsize{(0.00)} & 0.00 \scriptsize{(0.00)} \\
LLaVA\text{-}Onevision 7B & $\times$ & 0.00 \scriptsize{(0.00)} & 0.00 \scriptsize{(0.00)} & 0.00 \scriptsize{(0.00)} \\
LLaVA\text{-}Onevision 7B & $\checkmark$ & 0.00 \scriptsize{(0.00)} & 0.00 \scriptsize{(0.00)} & 0.00 \scriptsize{(0.00)} \\
Mistral\text{-}Small\text{-}3.1 24B & $\times$ & 0.68 \scriptsize{(0.09)} & 0.32 \scriptsize{(0.09)} & 0.32 \scriptsize{(0.09)} \\
Mistral\text{-}Small\text{-}3.1 24B & $\checkmark$ & 0.44 \scriptsize{(0.10)} & 0.28 \scriptsize{(0.09)} & 0.04 \scriptsize{(0.04)} \\
Molmo 7B & $\times$ & 0.00 \scriptsize{(0.00)} & 0.00 \scriptsize{(0.00)} & 0.00 \scriptsize{(0.00)} \\
Molmo 7B & $\checkmark$ & 0.00 \scriptsize{(0.00)} & 0.00 \scriptsize{(0.00)} & 0.00 \scriptsize{(0.00)} \\
Phi\text{-}4 Multimodal & $\times$ & 0.00 \scriptsize{(0.00)} & 0.00 \scriptsize{(0.00)} & 0.00 \scriptsize{(0.00)} \\
Phi\text{-}4 Multimodal & $\checkmark$ & 0.00 \scriptsize{(0.00)} & 0.00 \scriptsize{(0.00)} & 0.00 \scriptsize{(0.00)} \\
Qwen2.5\text{-}VL 72B & $\times$ & 0.68 \scriptsize{(0.09)} & 0.32 \scriptsize{(0.09)} & 0.04 \scriptsize{(0.04)} \\
Qwen2.5\text{-}VL 72B & $\checkmark$ & 0.44 \scriptsize{(0.10)} & 0.16 \scriptsize{(0.07)} & 0.32 \scriptsize{(0.09)} \\
Qwen2.5\text{-}VL 7B & $\times$ & 0.16 \scriptsize{(0.07)} & 0.12 \scriptsize{(0.06)} & 0.08 \scriptsize{(0.05)} \\
Qwen2.5\text{-}VL 7B & $\checkmark$ & 0.24 \scriptsize{(0.09)} & 0.00 \scriptsize{(0.00)} & 0.00 \scriptsize{(0.00)} \\
Qwen3\text{-}VL 30B A3B & $\times$ & 0.64 \scriptsize{(0.10)} & 0.56 \scriptsize{(0.10)} & 0.20 \scriptsize{(0.08)} \\
Qwen3\text{-}VL 30B A3B & $\checkmark$ & 0.40 \scriptsize{(0.10)} & 0.32 \scriptsize{(0.09)} & 0.32 \scriptsize{(0.09)} \\
Qwen3\text{-}VL 32B & $\times$ & \textbf{0.92 \scriptsize{(0.05)}} & 0.40 \scriptsize{(0.10)} & 0.20 \scriptsize{(0.08)} \\
Qwen3\text{-}VL 32B & $\checkmark$ & 0.40 \scriptsize{(0.10)} & 0.12 \scriptsize{(0.06)} & 0.08 \scriptsize{(0.05)} \\
Qwen3\text{-}VL 8B & $\times$ & 0.44 \scriptsize{(0.10)} & 0.08 \scriptsize{(0.05)} & 0.16 \scriptsize{(0.07)} \\
Qwen3\text{-}VL 8B & $\checkmark$ & 0.20 \scriptsize{(0.08)} & 0.16 \scriptsize{(0.07)} & 0.08 \scriptsize{(0.05)} \\
\bottomrule
\end{tabular}
\end{table*}

\begin{table*}[h]
\caption{\textbf{Full Results for Action and Action + CoT on ViPlan-BW.} Success rate indicates the fraction of problems solved. The highest value in each column is bolded.}
\label{tab:ViPLan-BW_act_react_full}
\centering
  \begin{tabular}{l c c c c}
\toprule
\textbf{Model} & \textbf{CoT} & \textbf{Simple} & \textbf{Medium} & \textbf{Hard} \\
\midrule
Gemma\text{-}3 27B & $\times$ & 0.28 \scriptsize{(0.09)} & 0.08 \scriptsize{(0.05)} & 0.00 \scriptsize{(0.00)} \\
Gemma\text{-}3 27B & $\checkmark$ & 0.28 \scriptsize{(0.09)} & 0.08 \scriptsize{(0.05)} & \textbf{0.04 \scriptsize{(0.04)}} \\
InternVL3 78B & $\times$ & 0.08 \scriptsize{(0.05)} & 0.00 \scriptsize{(0.00)} & 0.00 \scriptsize{(0.00)} \\
InternVL3 78B & $\checkmark$ & \textbf{0.56 \scriptsize{(0.10)}} & 0.08 \scriptsize{(0.05)} & \textbf{0.04 \scriptsize{(0.04)}} \\
InternVL3.5 38B & $\times$ & 0.48 \scriptsize{(0.10)} & 0.04 \scriptsize{(0.04)} & 0.00 \scriptsize{(0.00)} \\
InternVL3.5 38B & $\checkmark$ & 0.12 \scriptsize{(0.06)} & 0.04 \scriptsize{(0.04)} & 0.00 \scriptsize{(0.00)} \\
Qwen2.5\text{-}VL 72B & $\times$ & 0.04 \scriptsize{(0.04)} & 0.00 \scriptsize{(0.00)} & 0.00 \scriptsize{(0.00)} \\
Qwen2.5\text{-}VL 72B & $\checkmark$ & 0.12 \scriptsize{(0.06)} & 0.00 \scriptsize{(0.00)} & \textbf{0.04 \scriptsize{(0.04)}} \\
Qwen3\text{-}VL 32B & $\times$ & 0.04 \scriptsize{(0.04)} & 0.00 \scriptsize{(0.00)} & 0.00 \scriptsize{(0.00)} \\
Qwen3\text{-}VL 32B & $\checkmark$ & 0.40 \scriptsize{(0.10)} & \textbf{0.32 \scriptsize{(0.09)}} & 0.00 \scriptsize{(0.00)} \\
\bottomrule
\end{tabular}
\end{table*}

\begin{table*}[h]
\caption{\textbf{Full Results for Action and Action + CoT on ViPlan-HH.} Success rate indicates the fraction of problems solved. The highest value in each column is bolded.}
\label{tab:ViPLan-HH_act_react_full}
\centering
  \begin{tabular}{l c c c c}
\toprule
\textbf{Model} & \textbf{CoT} & \textbf{Simple} & \textbf{Medium} & \textbf{Hard} \\
\midrule
Gemma\text{-}3 27B & $\times$ & \textbf{0.80 \scriptsize{(0.08)}} & 0.20 \scriptsize{(0.08)} & 0.32 \scriptsize{(0.09)} \\
Gemma\text{-}3 27B & $\checkmark$ & 0.72 \scriptsize{(0.09)} & 0.28 \scriptsize{(0.09)} & 0.16 \scriptsize{(0.07)} \\
InternVL3 78B & $\times$ & 0.44 \scriptsize{(0.10)} & \textbf{0.48 \scriptsize{(0.10)}} & 0.32 \scriptsize{(0.09)} \\
InternVL3 78B & $\checkmark$ & 0.36 \scriptsize{(0.10)} & 0.32 \scriptsize{(0.09)} & 0.16 \scriptsize{(0.07)} \\
InternVL3.5 38B & $\times$ & 0.48 \scriptsize{(0.10)} & 0.32 \scriptsize{(0.09)} & 0.12 \scriptsize{(0.06)} \\
InternVL3.5 38B & $\checkmark$ & 0.44 \scriptsize{(0.10)} & 0.16 \scriptsize{(0.07)} & 0.08 \scriptsize{(0.05)} \\
Qwen2.5\text{-}VL 72B & $\times$ & 0.56 \scriptsize{(0.10)} & 0.40 \scriptsize{(0.10)} & 0.24 \scriptsize{(0.09)} \\
Qwen2.5\text{-}VL 72B & $\checkmark$ & 0.60 \scriptsize{(0.10)} & 0.36 \scriptsize{(0.10)} & \textbf{0.36 \scriptsize{(0.10)}} \\
Qwen3\text{-}VL 32B & $\times$ & 0.76 \scriptsize{(0.09)} & 0.44 \scriptsize{(0.10)} & 0.28 \scriptsize{(0.09)} \\
Qwen3\text{-}VL 32B & $\checkmark$ & \textbf{0.80 \scriptsize{(0.08)}} & 0.32 \scriptsize{(0.09)} & 0.12 \scriptsize{(0.06)} \\
\bottomrule
\end{tabular}
\end{table*}

\section{Action Failures on ViPlan-BW}
\label{app:action_failures}

While the main ViPlan-BW experiments were performed without action failures, the possibility of actions going wrong in the real world is a motivating example particularly for the VLM-as-grounder evaluation setting, as verifying preconditions and effects is a natural way of counteracting action failures. We opted to exclude this from the main results as, with action failures, a task can fail either due to the VLMs' poor performance or simply due to poor luck, which in turn makes comparing different models difficult. However, to test the robustness to action failures of the approach, we report the results with a probability of action failure $p_f = 0.1$, which we implement into our simulator, in Table~\ref{tab:ViPLan-BW_fail0_1_pred_full} for Ground and in Table~\ref{tab:ViPLan-BW_fail0_1_vila_full} for Plan.

Overall, we observe similar trends as the version with no failures (Table~\ref{tab:ViPLan-BW_pred_full}), showing the resilience of the Ground setting, and validating the no-failure choice for the main benchmark. 

Note that, as these experiments were performed in a preliminary version of the benchmark, some VLMs are missing from the results. Nevertheless, the method is the same, so the results are directly comparable to the ones in the main benchmark.

\begin{table*}[h]
\caption{\textbf{Full Results for Ground on ViPlan-BW with action failures.} Actions have a 10\% chance of failing. Success rate indicates the fraction of problems solved, accuracy is the fraction of correct predictions for a single predicate. The result is an average over the 25 problems in each split. Standard error of the mean is reported in parenthesis. The highest value in each column is bolded.}
\label{tab:ViPLan-BW_fail0_1_pred_full}
\centering
\resizebox{\linewidth}{!}{
  \begin{tabular}{l c c c c c c c}
\toprule
\multirow{2}{*}{\textbf{Model}} & \multirow{2}{*}{\textbf{CoT}} & \multicolumn{2}{c}{\textbf{Simple}} & \multicolumn{2}{c}{\textbf{Medium}} & \multicolumn{2}{c}{\textbf{Hard}} \\
 &  & \textbf{Success} & \textbf{Accuracy} & \textbf{Success} & \textbf{Accuracy} & \textbf{Success} & \textbf{Accuracy} \\
\midrule
AyaVision 32B & $\times$ & 0.04 \scriptsize{(0.04)} & 0.88 \scriptsize{(0.01)} & 0.00 \scriptsize{(0.00)} & 0.84 \scriptsize{(0.00)} & 0.00 \scriptsize{(0.00)} & 0.76 \scriptsize{(0.00)} \\
AyaVision 8B & $\times$ & 0.04 \scriptsize{(0.04)} & 0.68 \scriptsize{(0.01)} & 0.00 \scriptsize{(0.00)} & 0.70 \scriptsize{(0.01)} & 0.00 \scriptsize{(0.00)} & 0.76 \scriptsize{(0.01)} \\
Cosmos\text{-}Reason2 8B & $\times$ & 0.84 \scriptsize{(0.07)} & 0.99 \scriptsize{(0.00)} & 0.00 \scriptsize{(0.00)} & 0.97 \scriptsize{(0.00)} & 0.08 \scriptsize{(0.05)} & 0.96 \scriptsize{(0.00)} \\
DeepSeek\text{-}VL2 & $\times$ & 0.00 \scriptsize{(0.00)} & 0.91 \scriptsize{(0.01)} & 0.00 \scriptsize{(0.00)} & 0.88 \scriptsize{(0.01)} & 0.00 \scriptsize{(0.00)} & 0.88 \scriptsize{(0.01)} \\
GPT\text{-}4.1 Nano & $\times$ & 0.16 \scriptsize{(0.07)} & 0.89 \scriptsize{(0.01)} & 0.00 \scriptsize{(0.00)} & 0.84 \scriptsize{(0.00)} & 0.00 \scriptsize{(0.00)} & 0.84 \scriptsize{(0.00)} \\
Gemma\text{-}3 12B & $\times$ & 0.04 \scriptsize{(0.04)} & 0.96 \scriptsize{(0.00)} & 0.00 \scriptsize{(0.00)} & 0.95 \scriptsize{(0.00)} & 0.00 \scriptsize{(0.00)} & 0.91 \scriptsize{(0.00)} \\
Gemma\text{-}3 27B & $\times$ & 0.20 \scriptsize{(0.08)} & 0.96 \scriptsize{(0.00)} & 0.00 \scriptsize{(0.00)} & 0.94 \scriptsize{(0.00)} & 0.00 \scriptsize{(0.00)} & 0.92 \scriptsize{(0.01)} \\
InternVL3 78B & $\times$ & 0.92 \scriptsize{(0.05)} & 0.99 \scriptsize{(0.00)} & 0.84 \scriptsize{(0.07)} & \textbf{0.99 \scriptsize{(0.00)}} & \textbf{0.76 \scriptsize{(0.09)}} & \textbf{0.98 \scriptsize{(0.00)}} \\
InternVL3 8B & $\times$ & 0.32 \scriptsize{(0.09)} & 0.97 \scriptsize{(0.00)} & 0.00 \scriptsize{(0.00)} & 0.95 \scriptsize{(0.00)} & 0.00 \scriptsize{(0.00)} & 0.92 \scriptsize{(0.00)} \\
InternVL3.5 30B A3B & $\times$ & 0.44 \scriptsize{(0.10)} & 0.99 \scriptsize{(0.00)} & 0.08 \scriptsize{(0.05)} & \textbf{0.99 \scriptsize{(0.00)}} & 0.00 \scriptsize{(0.00)} & 0.97 \scriptsize{(0.00)} \\
InternVL3.5 38B & $\times$ & \textbf{1.00 \scriptsize{(0.00)}} & \textbf{1.00 \scriptsize{(0.00)}} & \textbf{0.88 \scriptsize{(0.06)}} & \textbf{0.99 \scriptsize{(0.00)}} & 0.68 \scriptsize{(0.09)} & \textbf{0.98 \scriptsize{(0.00)}} \\
InternVL3.5 8B & $\times$ & 0.92 \scriptsize{(0.05)} & 0.99 \scriptsize{(0.00)} & 0.32 \scriptsize{(0.09)} & \textbf{0.99 \scriptsize{(0.00)}} & 0.16 \scriptsize{(0.07)} & \textbf{0.98 \scriptsize{(0.00)}} \\
LLaVA\text{-}Onevision 72B & $\times$ & 0.00 \scriptsize{(0.00)} & 0.95 \scriptsize{(0.01)} & 0.00 \scriptsize{(0.00)} & 0.94 \scriptsize{(0.00)} & 0.00 \scriptsize{(0.00)} & 0.93 \scriptsize{(0.00)} \\
LLaVA\text{-}Onevision 7B & $\times$ & 0.00 \scriptsize{(0.00)} & 0.92 \scriptsize{(0.01)} & 0.00 \scriptsize{(0.00)} & 0.88 \scriptsize{(0.01)} & 0.00 \scriptsize{(0.00)} & 0.86 \scriptsize{(0.01)} \\
Mistral\text{-}Small\text{-}3.1 24B & $\times$ & 0.36 \scriptsize{(0.10)} & 0.99 \scriptsize{(0.00)} & 0.00 \scriptsize{(0.00)} & 0.96 \scriptsize{(0.00)} & 0.00 \scriptsize{(0.00)} & 0.95 \scriptsize{(0.00)} \\
Molmo 7B & $\times$ & 0.44 \scriptsize{(0.10)} & 0.94 \scriptsize{(0.00)} & 0.04 \scriptsize{(0.04)} & 0.93 \scriptsize{(0.00)} & 0.08 \scriptsize{(0.05)} & 0.92 \scriptsize{(0.00)} \\
Phi\text{-}4 Multimodal & $\times$ & 0.00 \scriptsize{(0.00)} & 0.92 \scriptsize{(0.01)} & 0.00 \scriptsize{(0.00)} & 0.90 \scriptsize{(0.00)} & 0.00 \scriptsize{(0.00)} & 0.83 \scriptsize{(0.01)} \\
Qwen2.5\text{-}VL 72B & $\times$ & 0.52 \scriptsize{(0.10)} & 0.98 \scriptsize{(0.00)} & 0.44 \scriptsize{(0.10)} & \textbf{0.99 \scriptsize{(0.00)}} & 0.08 \scriptsize{(0.05)} & 0.97 \scriptsize{(0.00)} \\
Qwen2.5\text{-}VL 7B & $\times$ & 0.76 \scriptsize{(0.09)} & 0.98 \scriptsize{(0.00)} & 0.12 \scriptsize{(0.06)} & 0.97 \scriptsize{(0.00)} & 0.00 \scriptsize{(0.00)} & 0.95 \scriptsize{(0.00)} \\
Qwen3\text{-}VL 30B A3B & $\times$ & 0.32 \scriptsize{(0.09)} & 0.98 \scriptsize{(0.00)} & 0.00 \scriptsize{(0.00)} & 0.98 \scriptsize{(0.00)} & 0.00 \scriptsize{(0.00)} & 0.97 \scriptsize{(0.00)} \\
Qwen3\text{-}VL 8B & $\times$ & 0.76 \scriptsize{(0.09)} & 0.99 \scriptsize{(0.00)} & 0.00 \scriptsize{(0.00)} & 0.98 \scriptsize{(0.00)} & 0.08 \scriptsize{(0.05)} & 0.97 \scriptsize{(0.00)} \\
\bottomrule
\end{tabular}}
\end{table*}

\begin{table*}[h]
\caption{\textbf{Full Results for Plan on ViPlan-BW with action failures.} Actions have a 10\% chance of failing. Success rate indicates the fraction of problems solved. The result is an average over the 25 problems in each split. Standard error of the mean is reported in parenthesis. The highest value in each column is bolded.}
\label{tab:ViPLan-BW_fail0_1_vila_full}
\centering
  \begin{tabular}{l c c c c}
\toprule
\textbf{Model} & \textbf{CoT} & \textbf{Simple} & \textbf{Medium} & \textbf{Hard} \\
\midrule
AyaVision 32B & $\times$ & 0.00 \scriptsize{(0.00)} & \textbf{0.00 \scriptsize{(0.00)}} & 0.00 \scriptsize{(0.00)} \\
AyaVision 8B & $\times$ & 0.00 \scriptsize{(0.00)} & \textbf{0.00 \scriptsize{(0.00)}} & 0.00 \scriptsize{(0.00)} \\
Cosmos\text{-}Reason2 8B & $\times$ & 0.00 \scriptsize{(0.00)} & \textbf{0.00 \scriptsize{(0.00)}} & 0.00 \scriptsize{(0.00)} \\
DeepSeek\text{-}VL2 & $\times$ & 0.00 \scriptsize{(0.00)} & \textbf{0.00 \scriptsize{(0.00)}} & 0.00 \scriptsize{(0.00)} \\
GPT\text{-}4.1 Nano & $\times$ & 0.08 \scriptsize{(0.05)} & \textbf{0.00 \scriptsize{(0.00)}} & 0.00 \scriptsize{(0.00)} \\
Gemma\text{-}3 12B & $\times$ & 0.20 \scriptsize{(0.08)} & \textbf{0.00 \scriptsize{(0.00)}} & 0.00 \scriptsize{(0.00)} \\
Gemma\text{-}3 27B & $\times$ & 0.24 \scriptsize{(0.09)} & \textbf{0.00 \scriptsize{(0.00)}} & 0.00 \scriptsize{(0.00)} \\
InternVL3 78B & $\times$ & 0.08 \scriptsize{(0.05)} & \textbf{0.00 \scriptsize{(0.00)}} & 0.00 \scriptsize{(0.00)} \\
InternVL3 8B & $\times$ & 0.04 \scriptsize{(0.04)} & \textbf{0.00 \scriptsize{(0.00)}} & 0.00 \scriptsize{(0.00)} \\
InternVL3.5 30B A3B & $\times$ & 0.00 \scriptsize{(0.00)} & \textbf{0.00 \scriptsize{(0.00)}} & 0.00 \scriptsize{(0.00)} \\
InternVL3.5 38B & $\times$ & \textbf{0.32 \scriptsize{(0.09)}} & \textbf{0.00 \scriptsize{(0.00)}} & 0.00 \scriptsize{(0.00)} \\
InternVL3.5 8B & $\times$ & 0.00 \scriptsize{(0.00)} & \textbf{0.00 \scriptsize{(0.00)}} & \textbf{0.04 \scriptsize{(0.04)}} \\
LLaVA\text{-}Onevision 72B & $\times$ & 0.00 \scriptsize{(0.00)} & \textbf{0.00 \scriptsize{(0.00)}} & 0.00 \scriptsize{(0.00)} \\
LLaVA\text{-}Onevision 7B & $\times$ & 0.00 \scriptsize{(0.00)} & \textbf{0.00 \scriptsize{(0.00)}} & 0.00 \scriptsize{(0.00)} \\
Mistral\text{-}Small\text{-}3.1 24B & $\times$ & 0.20 \scriptsize{(0.08)} & \textbf{0.00 \scriptsize{(0.00)}} & 0.00 \scriptsize{(0.00)} \\
Molmo 7B & $\times$ & 0.00 \scriptsize{(0.00)} & \textbf{0.00 \scriptsize{(0.00)}} & 0.00 \scriptsize{(0.00)} \\
Phi\text{-}4 Multimodal & $\times$ & 0.00 \scriptsize{(0.00)} & \textbf{0.00 \scriptsize{(0.00)}} & 0.00 \scriptsize{(0.00)} \\
Qwen2.5\text{-}VL 72B & $\times$ & 0.08 \scriptsize{(0.05)} & \textbf{0.00 \scriptsize{(0.00)}} & 0.00 \scriptsize{(0.00)} \\
Qwen2.5\text{-}VL 7B & $\times$ & 0.00 \scriptsize{(0.00)} & \textbf{0.00 \scriptsize{(0.00)}} & 0.00 \scriptsize{(0.00)} \\
Qwen3\text{-}VL 30B A3B & $\times$ & 0.00 \scriptsize{(0.00)} & \textbf{0.00 \scriptsize{(0.00)}} & \textbf{0.04 \scriptsize{(0.04)}} \\
Qwen3\text{-}VL 8B & $\times$ & 0.04 \scriptsize{(0.04)} & \textbf{0.00 \scriptsize{(0.00)}} & 0.00 \scriptsize{(0.00)} \\
\bottomrule
\end{tabular}
\end{table*}

\section{Blank-Image Ablation}
\label{app:blank_image}

To assess how strongly the two paradigms depend on visual information, we rerun Plan and Ground with InternVL3.5 38B while replacing the environment observation with a blank image and leaving all textual inputs unchanged. In ViPlan-HH, these textual inputs include the goal, action descriptions, previous-action history, and the privileged information described in Section~\ref{sec:methods}.

\begin{table*}[t]
\centering
\caption{\textbf{Blank-image ablation with InternVL3.5 38B.} Success rates with the original visual observation and after replacing it with a blank image, while keeping all textual inputs unchanged. Bold indicates the higher score within each pair. Ground collapses to 0\% success without visual input in both domains, whereas Plan degrades in ViPlan-BW and not in ViPlan-HH.}
\label{tab:blank_image}
\begin{tabular}{llccc|ccc}
\toprule
& & \multicolumn{3}{c|}{\textbf{ViPlan-BW}} & \multicolumn{3}{c}{\textbf{ViPlan-HH}} \\
\textbf{Method} & \textbf{Input} & \textbf{Simple} & \textbf{Medium} & \textbf{Hard} & \textbf{Simple} & \textbf{Medium} & \textbf{Hard} \\
\midrule
Plan &
Original &
\textbf{0.40} & \textbf{0.08} & 0.00 &
0.44 & 0.08 & \textbf{0.20} \\

Plan &
Blank &
0.12 & 0.00 & 0.00 &
\textbf{0.64} & 0.08 & 0.00 \\
\midrule
Ground &
Original &
\textbf{1.00} & \textbf{0.88} & \textbf{0.60} &
\textbf{0.08} & \textbf{0.04} & 0.00 \\

Ground &
Blank &
0.00 & 0.00 & 0.00 &
0.00 & 0.00 & 0.00 \\
\bottomrule
\end{tabular}
\end{table*}

Plan drops from an average success rate of 16\% to 4\% in ViPlan-BW, whereas its average ViPlan-HH success remains 24\%, even improving from 44\% to 64\% when not given the image in the simple split. Ground instead falls to 0\% in both domains, including from 83\% to 0\% on average in ViPlan-BW. These results are consistent with direct VLM planning being able to exploit non-visual task information in the household domain, while symbolic grounding necessarily depends on accurately recovering the visual state.

\section{Predicate Accuracy Results}
\label{app:predicate_accuracy_results}

We report the accuracy for individual predicates Q/A pairs for methods in the VLM-as-grounder class on ViPlan-BW in Tables~\ref{tab:ViPLan-BW_pred_no_cot} (No CoT) and~\ref{tab:ViPLan-BW_pred_cot} (CoT), and on ViPlan-HH in Tables~\ref{tab:ViPLan-HH_pred_no_cot} and~\ref{tab:ViPLan-HH_pred_cot} (CoT).

Some predicates are clearly harder than others to predict: in ViPlan-BW, the \texttt{clear} predicate tends to show much worse performance compared to the others, with some models even approaching random chance (0.50). \texttt{clear} indicates a block that can be moved, and is translated to "Is the block x the topmost of its column?", which is a harder question compared to the other predicates, which are more straight-forward. This is an example of what is sometimes known as a \emph{derived predicate}, as it could also be obtained with a combination of \texttt{forall} and \texttt{on}, which confirms that predicates encoding higher-level relationships can be more challenging also for VLMs. The accuracy also tends to worsen as the splits increase in difficulty, which suggests that having more objects in the image (as harder splits have more blocks) results in a harder task for the VLM.

The performance is overall worse for ViPlan-HH, where many predicates, such as \texttt{nextto}, \texttt{open} and \texttt{reachable} fail to reach 90\% accuracy even for the best-performing models. This can be attributed to the ambiguity of the domain: if in ViPlan-BW all predicates are distinctly and correctly identifiable, asking if an object is reachable by the agent or next to another object can require a degree of interpretation, which is challenging for VLMs. More well-defined predicates, such as \texttt{holding} and \texttt{inside} show a higher accuracy, confirming this hypothesis.

Furthermore, our experiments also measure the individual performance by ground truth answer (yes-no), which we omit from this manuscript for brevity. This reveals that some models, like Molmo and the AyaVision family, have a strong bias towards answering "yes", while others like DeepSeek-VL2 tend to always answer "no", leading to poor performance.

\begin{table*}[h]
\centering
\caption{\textbf{Individual Predicate Accuracy for Ground on ViPlan-BW.} The table shows the accuracy for each predicate in each split. Bolded values show the best accuracy for each predicate and split. Standard error of the mean is reported in parenthesis.}
\label{tab:ViPLan-BW_pred_no_cot}
\resizebox{!}{0.46\textheight}{
\begin{tabular}{l l c c c c c}
\toprule
\textbf{Model} & \textbf{Split} & \texttt{clear} & \texttt{incolumn} & \texttt{leftof} & \texttt{on} & \texttt{rightof} \\
\midrule
\multirow{3}{*}{AyaVision 32B} & Simple & 0.70 \scriptsize{(0.03)} & 0.77 \scriptsize{(0.02)} & \textbf{1.00 \scriptsize{(0.00)}} & 0.69 \scriptsize{(0.02)} & \textbf{1.00 \scriptsize{(0.00)}} \\
 & Medium & 0.74 \scriptsize{(0.02)} & 0.76 \scriptsize{(0.01)} & 0.98 \scriptsize{(0.00)} & 0.57 \scriptsize{(0.01)} & \textbf{1.00 \scriptsize{(0.00)}} \\
 & Hard & 0.59 \scriptsize{(0.02)} & 0.79 \scriptsize{(0.01)} & \textbf{1.00 \scriptsize{(0.00)}} & 0.55 \scriptsize{(0.01)} & \textbf{1.00 \scriptsize{(0.00)}} \\
\midrule
\multirow{3}{*}{AyaVision 8B} & Simple & 0.78 \scriptsize{(0.02)} & 0.72 \scriptsize{(0.01)} & 0.70 \scriptsize{(0.01)} & 0.91 \scriptsize{(0.01)} & 0.48 \scriptsize{(0.01)} \\
 & Medium & 0.62 \scriptsize{(0.04)} & 0.75 \scriptsize{(0.01)} & 0.62 \scriptsize{(0.02)} & 0.89 \scriptsize{(0.01)} & 0.53 \scriptsize{(0.02)} \\
 & Hard & 0.57 \scriptsize{(0.03)} & 0.81 \scriptsize{(0.01)} & 0.66 \scriptsize{(0.02)} & 0.91 \scriptsize{(0.01)} & 0.53 \scriptsize{(0.02)} \\
\midrule
\multirow{3}{*}{Cosmos-Reason2 8B} & Simple & 0.96 \scriptsize{(0.02)} & 0.96 \scriptsize{(0.01)} & \textbf{1.00 \scriptsize{(0.00)}} & 0.98 \scriptsize{(0.01)} & \textbf{1.00 \scriptsize{(0.00)}} \\
 & Medium & 0.83 \scriptsize{(0.02)} & 0.98 \scriptsize{(0.00)} & \textbf{1.00 \scriptsize{(0.00)}} & 0.93 \scriptsize{(0.00)} & \textbf{1.00 \scriptsize{(0.00)}} \\
 & Hard & 0.85 \scriptsize{(0.02)} & 0.97 \scriptsize{(0.00)} & \textbf{1.00 \scriptsize{(0.00)}} & 0.93 \scriptsize{(0.00)} & \textbf{1.00 \scriptsize{(0.00)}} \\
\midrule
\multirow{3}{*}{DeepSeek-VL2} & Simple & 0.23 \scriptsize{(0.05)} & 0.95 \scriptsize{(0.01)} & 0.92 \scriptsize{(0.01)} & 0.93 \scriptsize{(0.02)} & \textbf{1.00 \scriptsize{(0.00)}} \\
 & Medium & 0.33 \scriptsize{(0.04)} & 0.95 \scriptsize{(0.01)} & 0.88 \scriptsize{(0.01)} & 0.81 \scriptsize{(0.02)} & 0.99 \scriptsize{(0.00)} \\
 & Hard & 0.54 \scriptsize{(0.04)} & 0.98 \scriptsize{(0.01)} & 0.88 \scriptsize{(0.02)} & 0.82 \scriptsize{(0.01)} & \textbf{1.00 \scriptsize{(0.00)}} \\
\midrule
\multirow{3}{*}{GPT-4.1} & Simple & 0.99 \scriptsize{(0.01)} & 0.99 \scriptsize{(0.01)} & \textbf{1.00 \scriptsize{(0.00)}} & 0.99 \scriptsize{(0.01)} & \textbf{1.00 \scriptsize{(0.00)}} \\
 & Medium & 0.93 \scriptsize{(0.02)} & 0.98 \scriptsize{(0.00)} & \textbf{1.00 \scriptsize{(0.00)}} & 0.98 \scriptsize{(0.00)} & \textbf{1.00 \scriptsize{(0.00)}} \\
 & Hard & 0.95 \scriptsize{(0.01)} & 0.97 \scriptsize{(0.00)} & \textbf{1.00 \scriptsize{(0.00)}} & \textbf{0.98 \scriptsize{(0.00)}} & \textbf{1.00 \scriptsize{(0.00)}} \\
\midrule
\multirow{3}{*}{GPT-4.1 Nano} & Simple & 0.67 \scriptsize{(0.05)} & 0.97 \scriptsize{(0.01)} & 0.90 \scriptsize{(0.01)} & 0.91 \scriptsize{(0.02)} & 0.86 \scriptsize{(0.02)} \\
 & Medium & 0.70 \scriptsize{(0.02)} & 0.91 \scriptsize{(0.01)} & 0.79 \scriptsize{(0.01)} & 0.87 \scriptsize{(0.01)} & 0.86 \scriptsize{(0.01)} \\
 & Hard & 0.54 \scriptsize{(0.02)} & 0.94 \scriptsize{(0.01)} & 0.89 \scriptsize{(0.01)} & 0.79 \scriptsize{(0.01)} & 0.86 \scriptsize{(0.01)} \\
\midrule
\multirow{3}{*}{GPT-5.2} & Simple & 0.98 \scriptsize{(0.02)} & 0.98 \scriptsize{(0.01)} & \textbf{1.00 \scriptsize{(0.00)}} & 0.98 \scriptsize{(0.01)} & \textbf{1.00 \scriptsize{(0.00)}} \\
 & Medium & 0.95 \scriptsize{(0.02)} & 0.98 \scriptsize{(0.00)} & \textbf{1.00 \scriptsize{(0.00)}} & 0.98 \scriptsize{(0.00)} & \textbf{1.00 \scriptsize{(0.00)}} \\
 & Hard & 0.95 \scriptsize{(0.01)} & 0.97 \scriptsize{(0.00)} & \textbf{1.00 \scriptsize{(0.00)}} & \textbf{0.98 \scriptsize{(0.00)}} & \textbf{1.00 \scriptsize{(0.00)}} \\
\midrule
\multirow{3}{*}{Gemma-3 12B} & Simple & 0.57 \scriptsize{(0.06)} & 0.94 \scriptsize{(0.01)} & \textbf{1.00 \scriptsize{(0.00)}} & 0.96 \scriptsize{(0.01)} & \textbf{1.00 \scriptsize{(0.00)}} \\
 & Medium & 0.64 \scriptsize{(0.04)} & 0.94 \scriptsize{(0.01)} & \textbf{1.00 \scriptsize{(0.00)}} & 0.93 \scriptsize{(0.01)} & \textbf{1.00 \scriptsize{(0.00)}} \\
 & Hard & 0.69 \scriptsize{(0.03)} & 0.91 \scriptsize{(0.01)} & \textbf{1.00 \scriptsize{(0.00)}} & 0.88 \scriptsize{(0.01)} & \textbf{1.00 \scriptsize{(0.00)}} \\
\midrule
\multirow{3}{*}{Gemma-3 27B} & Simple & 0.85 \scriptsize{(0.03)} & 0.96 \scriptsize{(0.01)} & 0.95 \scriptsize{(0.01)} & 0.96 \scriptsize{(0.01)} & \textbf{1.00 \scriptsize{(0.00)}} \\
 & Medium & 0.62 \scriptsize{(0.04)} & 0.96 \scriptsize{(0.01)} & 0.93 \scriptsize{(0.01)} & 0.92 \scriptsize{(0.01)} & \textbf{1.00 \scriptsize{(0.00)}} \\
 & Hard & 0.69 \scriptsize{(0.04)} & 0.94 \scriptsize{(0.01)} & 0.97 \scriptsize{(0.01)} & 0.90 \scriptsize{(0.01)} & \textbf{1.00 \scriptsize{(0.00)}} \\
\midrule
\multirow{3}{*}{InternVL3 78B} & Simple & \textbf{1.00 \scriptsize{(0.00)}} & 0.99 \scriptsize{(0.01)} & \textbf{1.00 \scriptsize{(0.00)}} & \textbf{1.00 \scriptsize{(0.00)}} & \textbf{1.00 \scriptsize{(0.00)}} \\
 & Medium & 0.99 \scriptsize{(0.01)} & \textbf{1.00 \scriptsize{(0.00)}} & \textbf{1.00 \scriptsize{(0.00)}} & 0.98 \scriptsize{(0.01)} & 0.98 \scriptsize{(0.01)} \\
 & Hard & 0.96 \scriptsize{(0.01)} & 0.99 \scriptsize{(0.00)} & \textbf{1.00 \scriptsize{(0.00)}} & 0.96 \scriptsize{(0.00)} & 0.99 \scriptsize{(0.00)} \\
\midrule
\multirow{3}{*}{InternVL3 8B} & Simple & 0.73 \scriptsize{(0.03)} & 0.97 \scriptsize{(0.01)} & \textbf{1.00 \scriptsize{(0.00)}} & 0.93 \scriptsize{(0.01)} & \textbf{1.00 \scriptsize{(0.00)}} \\
 & Medium & 0.66 \scriptsize{(0.03)} & 0.95 \scriptsize{(0.01)} & \textbf{1.00 \scriptsize{(0.00)}} & 0.92 \scriptsize{(0.01)} & \textbf{1.00 \scriptsize{(0.00)}} \\
 & Hard & 0.58 \scriptsize{(0.03)} & 0.97 \scriptsize{(0.00)} & \textbf{1.00 \scriptsize{(0.00)}} & 0.88 \scriptsize{(0.01)} & 0.99 \scriptsize{(0.00)} \\
\midrule
\multirow{3}{*}{InternVL3.5 30B A3B} & Simple & 0.90 \scriptsize{(0.03)} & 0.99 \scriptsize{(0.00)} & \textbf{1.00 \scriptsize{(0.00)}} & 0.99 \scriptsize{(0.01)} & \textbf{1.00 \scriptsize{(0.00)}} \\
 & Medium & 0.79 \scriptsize{(0.03)} & 0.98 \scriptsize{(0.00)} & \textbf{1.00 \scriptsize{(0.00)}} & 0.98 \scriptsize{(0.00)} & \textbf{1.00 \scriptsize{(0.00)}} \\
 & Hard & 0.88 \scriptsize{(0.02)} & 0.98 \scriptsize{(0.00)} & \textbf{1.00 \scriptsize{(0.00)}} & 0.96 \scriptsize{(0.00)} & \textbf{1.00 \scriptsize{(0.00)}} \\
\midrule
\multirow{3}{*}{InternVL3.5 38B} & Simple & \textbf{1.00 \scriptsize{(0.00)}} & \textbf{1.00 \scriptsize{(0.00)}} & \textbf{1.00 \scriptsize{(0.00)}} & \textbf{1.00 \scriptsize{(0.00)}} & \textbf{1.00 \scriptsize{(0.00)}} \\
 & Medium & \textbf{1.00 \scriptsize{(0.00)}} & \textbf{1.00 \scriptsize{(0.00)}} & \textbf{1.00 \scriptsize{(0.00)}} & 0.95 \scriptsize{(0.01)} & \textbf{1.00 \scriptsize{(0.00)}} \\
 & Hard & \textbf{1.00 \scriptsize{(0.00)}} & 0.99 \scriptsize{(0.00)} & \textbf{1.00 \scriptsize{(0.00)}} & 0.95 \scriptsize{(0.00)} & \textbf{1.00 \scriptsize{(0.00)}} \\
\midrule
\multirow{3}{*}{InternVL3.5 8B} & Simple & 0.93 \scriptsize{(0.02)} & 0.99 \scriptsize{(0.00)} & \textbf{1.00 \scriptsize{(0.00)}} & \textbf{1.00 \scriptsize{(0.00)}} & \textbf{1.00 \scriptsize{(0.00)}} \\
 & Medium & 0.82 \scriptsize{(0.02)} & \textbf{1.00 \scriptsize{(0.00)}} & \textbf{1.00 \scriptsize{(0.00)}} & \textbf{0.99 \scriptsize{(0.00)}} & \textbf{1.00 \scriptsize{(0.00)}} \\
 & Hard & 0.81 \scriptsize{(0.02)} & \textbf{1.00 \scriptsize{(0.00)}} & \textbf{1.00 \scriptsize{(0.00)}} & \textbf{0.98 \scriptsize{(0.00)}} & \textbf{1.00 \scriptsize{(0.00)}} \\
\midrule
\multirow{3}{*}{LLaVA-Onevision 72B} & Simple & 0.49 \scriptsize{(0.06)} & 0.94 \scriptsize{(0.01)} & \textbf{1.00 \scriptsize{(0.00)}} & 0.95 \scriptsize{(0.01)} & \textbf{1.00 \scriptsize{(0.00)}} \\
 & Medium & 0.47 \scriptsize{(0.05)} & 0.91 \scriptsize{(0.01)} & \textbf{1.00 \scriptsize{(0.00)}} & 0.93 \scriptsize{(0.01)} & \textbf{1.00 \scriptsize{(0.00)}} \\
 & Hard & 0.68 \scriptsize{(0.04)} & 0.94 \scriptsize{(0.01)} & \textbf{1.00 \scriptsize{(0.00)}} & 0.91 \scriptsize{(0.01)} & \textbf{1.00 \scriptsize{(0.00)}} \\
\midrule
\multirow{3}{*}{LLaVA-Onevision 7B} & Simple & 0.40 \scriptsize{(0.06)} & 0.90 \scriptsize{(0.02)} & 0.97 \scriptsize{(0.01)} & 0.96 \scriptsize{(0.01)} & 0.95 \scriptsize{(0.01)} \\
 & Medium & 0.36 \scriptsize{(0.04)} & 0.90 \scriptsize{(0.01)} & 0.96 \scriptsize{(0.01)} & 0.83 \scriptsize{(0.01)} & 0.95 \scriptsize{(0.01)} \\
 & Hard & 0.52 \scriptsize{(0.03)} & 0.94 \scriptsize{(0.01)} & 0.96 \scriptsize{(0.01)} & 0.79 \scriptsize{(0.01)} & 0.93 \scriptsize{(0.01)} \\
\midrule
\multirow{3}{*}{Mistral-Small-3.1 24B} & Simple & 0.89 \scriptsize{(0.03)} & 0.98 \scriptsize{(0.01)} & \textbf{1.00 \scriptsize{(0.00)}} & 0.99 \scriptsize{(0.01)} & \textbf{1.00 \scriptsize{(0.00)}} \\
 & Medium & 0.72 \scriptsize{(0.04)} & 0.96 \scriptsize{(0.01)} & 0.98 \scriptsize{(0.01)} & 0.95 \scriptsize{(0.01)} & 0.99 \scriptsize{(0.00)} \\
 & Hard & 0.75 \scriptsize{(0.03)} & 0.98 \scriptsize{(0.01)} & \textbf{1.00 \scriptsize{(0.00)}} & 0.94 \scriptsize{(0.01)} & 0.99 \scriptsize{(0.00)} \\
\midrule
\multirow{3}{*}{Molmo 7B} & Simple & 0.77 \scriptsize{(0.02)} & 0.98 \scriptsize{(0.00)} & 0.95 \scriptsize{(0.01)} & 0.88 \scriptsize{(0.01)} & 0.97 \scriptsize{(0.00)} \\
 & Medium & 0.81 \scriptsize{(0.01)} & 0.97 \scriptsize{(0.00)} & 0.96 \scriptsize{(0.00)} & 0.86 \scriptsize{(0.01)} & 0.97 \scriptsize{(0.00)} \\
 & Hard & 0.77 \scriptsize{(0.01)} & 0.95 \scriptsize{(0.00)} & 0.96 \scriptsize{(0.00)} & 0.88 \scriptsize{(0.00)} & 0.92 \scriptsize{(0.00)} \\
\midrule
\multirow{3}{*}{Phi-4 Multimodal} & Simple & 0.44 \scriptsize{(0.05)} & 0.96 \scriptsize{(0.01)} & \textbf{1.00 \scriptsize{(0.00)}} & 0.77 \scriptsize{(0.02)} & \textbf{1.00 \scriptsize{(0.00)}} \\
 & Medium & 0.36 \scriptsize{(0.04)} & 0.97 \scriptsize{(0.01)} & \textbf{1.00 \scriptsize{(0.00)}} & 0.71 \scriptsize{(0.02)} & \textbf{1.00 \scriptsize{(0.00)}} \\
 & Hard & 0.54 \scriptsize{(0.03)} & 0.95 \scriptsize{(0.01)} & \textbf{1.00 \scriptsize{(0.00)}} & 0.64 \scriptsize{(0.01)} & \textbf{1.00 \scriptsize{(0.00)}} \\
\midrule
\multirow{3}{*}{Qwen2.5-VL 72B} & Simple & 0.95 \scriptsize{(0.02)} & 0.95 \scriptsize{(0.01)} & \textbf{1.00 \scriptsize{(0.00)}} & 0.98 \scriptsize{(0.01)} & \textbf{1.00 \scriptsize{(0.00)}} \\
 & Medium & 0.97 \scriptsize{(0.01)} & 0.97 \scriptsize{(0.01)} & \textbf{1.00 \scriptsize{(0.00)}} & 0.98 \scriptsize{(0.01)} & \textbf{1.00 \scriptsize{(0.00)}} \\
 & Hard & 0.94 \scriptsize{(0.02)} & 0.96 \scriptsize{(0.01)} & \textbf{1.00 \scriptsize{(0.00)}} & 0.96 \scriptsize{(0.01)} & \textbf{1.00 \scriptsize{(0.00)}} \\
\midrule
\multirow{3}{*}{Qwen2.5-VL 7B} & Simple & 0.95 \scriptsize{(0.02)} & \textbf{1.00 \scriptsize{(0.00)}} & 0.98 \scriptsize{(0.01)} & \textbf{1.00 \scriptsize{(0.00)}} & 0.99 \scriptsize{(0.00)} \\
 & Medium & 0.87 \scriptsize{(0.03)} & 0.98 \scriptsize{(0.00)} & 0.98 \scriptsize{(0.00)} & 0.98 \scriptsize{(0.00)} & 0.97 \scriptsize{(0.01)} \\
 & Hard & 0.83 \scriptsize{(0.03)} & 0.97 \scriptsize{(0.01)} & 0.90 \scriptsize{(0.01)} & 0.96 \scriptsize{(0.01)} & 0.94 \scriptsize{(0.01)} \\
\midrule
\multirow{3}{*}{Qwen3-VL 30B A3B} & Simple & 0.91 \scriptsize{(0.02)} & 0.95 \scriptsize{(0.01)} & 0.99 \scriptsize{(0.00)} & 0.99 \scriptsize{(0.00)} & \textbf{1.00 \scriptsize{(0.00)}} \\
 & Medium & 0.74 \scriptsize{(0.03)} & 0.97 \scriptsize{(0.01)} & 0.99 \scriptsize{(0.00)} & 0.97 \scriptsize{(0.01)} & \textbf{1.00 \scriptsize{(0.00)}} \\
 & Hard & 0.85 \scriptsize{(0.02)} & 0.97 \scriptsize{(0.01)} & \textbf{1.00 \scriptsize{(0.00)}} & 0.95 \scriptsize{(0.01)} & \textbf{1.00 \scriptsize{(0.00)}} \\
\midrule
\multirow{3}{*}{Qwen3-VL 32B} & Simple & 0.94 \scriptsize{(0.02)} & 0.96 \scriptsize{(0.01)} & \textbf{1.00 \scriptsize{(0.00)}} & 0.98 \scriptsize{(0.01)} & \textbf{1.00 \scriptsize{(0.00)}} \\
 & Medium & 0.89 \scriptsize{(0.02)} & 0.97 \scriptsize{(0.00)} & \textbf{1.00 \scriptsize{(0.00)}} & 0.97 \scriptsize{(0.00)} & \textbf{1.00 \scriptsize{(0.00)}} \\
 & Hard & 0.86 \scriptsize{(0.02)} & 0.96 \scriptsize{(0.00)} & \textbf{1.00 \scriptsize{(0.00)}} & 0.95 \scriptsize{(0.00)} & \textbf{1.00 \scriptsize{(0.00)}} \\
\midrule
\multirow{3}{*}{Qwen3-VL 8B} & Simple & 0.93 \scriptsize{(0.03)} & 0.98 \scriptsize{(0.01)} & \textbf{1.00 \scriptsize{(0.00)}} & 0.97 \scriptsize{(0.01)} & \textbf{1.00 \scriptsize{(0.00)}} \\
 & Medium & 0.84 \scriptsize{(0.02)} & 0.98 \scriptsize{(0.00)} & \textbf{1.00 \scriptsize{(0.00)}} & 0.98 \scriptsize{(0.00)} & \textbf{1.00 \scriptsize{(0.00)}} \\
 & Hard & 0.89 \scriptsize{(0.02)} & 0.99 \scriptsize{(0.00)} & \textbf{1.00 \scriptsize{(0.00)}} & 0.97 \scriptsize{(0.00)} & \textbf{1.00 \scriptsize{(0.00)}} \\
\bottomrule
\end{tabular}}
\end{table*}

\begin{table*}[h]
\centering
\caption{\textbf{Individual Predicate Accuracy for Ground + CoT on ViPlan-BW.} The table shows the accuracy for each predicate in each split. Bolded values show the best accuracy for each predicate and split. Standard error of the mean is reported in parenthesis.}
\label{tab:ViPLan-BW_pred_cot}
\resizebox{!}{0.46\textheight}{
\begin{tabular}{l l c c c c c}
\toprule
\textbf{Model} & \textbf{Split} & \texttt{clear} & \texttt{incolumn} & \texttt{leftof} & \texttt{on} & \texttt{rightof} \\
\midrule
\multirow{3}{*}{AyaVision 32B} & Simple & 0.73 \scriptsize{(0.03)} & 0.78 \scriptsize{(0.02)} & \textbf{1.00 \scriptsize{(0.00)}} & 0.74 \scriptsize{(0.02)} & 0.98 \scriptsize{(0.00)} \\
 & Medium & 0.69 \scriptsize{(0.02)} & 0.80 \scriptsize{(0.01)} & \textbf{1.00 \scriptsize{(0.00)}} & 0.66 \scriptsize{(0.01)} & 0.98 \scriptsize{(0.00)} \\
 & Hard & 0.53 \scriptsize{(0.02)} & 0.84 \scriptsize{(0.01)} & \textbf{1.00 \scriptsize{(0.00)}} & 0.65 \scriptsize{(0.01)} & 0.96 \scriptsize{(0.01)} \\
\midrule
\multirow{3}{*}{AyaVision 8B} & Simple & 0.52 \scriptsize{(0.06)} & 0.58 \scriptsize{(0.03)} & 0.46 \scriptsize{(0.02)} & 0.85 \scriptsize{(0.02)} & 0.30 \scriptsize{(0.02)} \\
 & Medium & 0.45 \scriptsize{(0.05)} & 0.58 \scriptsize{(0.02)} & 0.44 \scriptsize{(0.02)} & 0.80 \scriptsize{(0.02)} & 0.33 \scriptsize{(0.02)} \\
 & Hard & 0.53 \scriptsize{(0.04)} & 0.65 \scriptsize{(0.02)} & 0.43 \scriptsize{(0.02)} & 0.76 \scriptsize{(0.01)} & 0.34 \scriptsize{(0.02)} \\
\midrule
\multirow{3}{*}{Cosmos-Reason2 8B} & Simple & 0.93 \scriptsize{(0.02)} & 0.91 \scriptsize{(0.01)} & \textbf{1.00 \scriptsize{(0.00)}} & 0.98 \scriptsize{(0.01)} & \textbf{1.00 \scriptsize{(0.00)}} \\
 & Medium & 0.86 \scriptsize{(0.02)} & 0.96 \scriptsize{(0.00)} & \textbf{1.00 \scriptsize{(0.00)}} & 0.96 \scriptsize{(0.00)} & \textbf{1.00 \scriptsize{(0.00)}} \\
 & Hard & 0.92 \scriptsize{(0.01)} & 0.92 \scriptsize{(0.01)} & \textbf{1.00 \scriptsize{(0.00)}} & 0.95 \scriptsize{(0.00)} & \textbf{1.00 \scriptsize{(0.00)}} \\
\midrule
\multirow{3}{*}{DeepSeek-VL2} & Simple & 0.71 \scriptsize{(0.03)} & 0.96 \scriptsize{(0.01)} & 0.92 \scriptsize{(0.01)} & 0.77 \scriptsize{(0.02)} & 0.96 \scriptsize{(0.01)} \\
 & Medium & 0.64 \scriptsize{(0.03)} & 0.87 \scriptsize{(0.01)} & 0.91 \scriptsize{(0.01)} & 0.74 \scriptsize{(0.01)} & 0.94 \scriptsize{(0.01)} \\
 & Hard & 0.60 \scriptsize{(0.03)} & 0.95 \scriptsize{(0.01)} & 0.94 \scriptsize{(0.01)} & 0.76 \scriptsize{(0.01)} & 0.96 \scriptsize{(0.01)} \\
\midrule
\multirow{3}{*}{GPT-4.1} & Simple & \textbf{1.00 \scriptsize{(0.00)}} & 0.97 \scriptsize{(0.01)} & \textbf{1.00 \scriptsize{(0.00)}} & 0.99 \scriptsize{(0.01)} & \textbf{1.00 \scriptsize{(0.00)}} \\
 & Medium & 0.95 \scriptsize{(0.01)} & 0.97 \scriptsize{(0.00)} & \textbf{1.00 \scriptsize{(0.00)}} & 0.97 \scriptsize{(0.00)} & \textbf{1.00 \scriptsize{(0.00)}} \\
 & Hard & 0.96 \scriptsize{(0.01)} & 0.96 \scriptsize{(0.00)} & \textbf{1.00 \scriptsize{(0.00)}} & \textbf{0.97 \scriptsize{(0.00)}} & \textbf{1.00 \scriptsize{(0.00)}} \\
\midrule
\multirow{3}{*}{GPT-4.1 Nano} & Simple & 0.88 \scriptsize{(0.03)} & 0.91 \scriptsize{(0.01)} & \textbf{1.00 \scriptsize{(0.00)}} & 0.91 \scriptsize{(0.01)} & \textbf{1.00 \scriptsize{(0.00)}} \\
 & Medium & 0.84 \scriptsize{(0.02)} & 0.93 \scriptsize{(0.01)} & \textbf{1.00 \scriptsize{(0.00)}} & 0.94 \scriptsize{(0.01)} & \textbf{1.00 \scriptsize{(0.00)}} \\
 & Hard & 0.84 \scriptsize{(0.02)} & 0.92 \scriptsize{(0.01)} & \textbf{1.00 \scriptsize{(0.00)}} & 0.88 \scriptsize{(0.01)} & \textbf{1.00 \scriptsize{(0.00)}} \\
\midrule
\multirow{3}{*}{Gemma-3 12B} & Simple & 0.83 \scriptsize{(0.03)} & 0.88 \scriptsize{(0.01)} & 0.97 \scriptsize{(0.01)} & 0.84 \scriptsize{(0.02)} & 0.99 \scriptsize{(0.00)} \\
 & Medium & 0.68 \scriptsize{(0.04)} & 0.92 \scriptsize{(0.01)} & 0.95 \scriptsize{(0.01)} & 0.81 \scriptsize{(0.02)} & 0.98 \scriptsize{(0.01)} \\
 & Hard & 0.72 \scriptsize{(0.03)} & 0.91 \scriptsize{(0.01)} & 0.94 \scriptsize{(0.01)} & 0.81 \scriptsize{(0.01)} & 0.99 \scriptsize{(0.00)} \\
\midrule
\multirow{3}{*}{Gemma-3 27B} & Simple & 0.86 \scriptsize{(0.03)} & 0.94 \scriptsize{(0.01)} & \textbf{1.00 \scriptsize{(0.00)}} & 0.89 \scriptsize{(0.02)} & \textbf{1.00 \scriptsize{(0.00)}} \\
 & Medium & 0.80 \scriptsize{(0.02)} & 0.95 \scriptsize{(0.01)} & 0.99 \scriptsize{(0.00)} & 0.83 \scriptsize{(0.01)} & \textbf{1.00 \scriptsize{(0.00)}} \\
 & Hard & 0.84 \scriptsize{(0.02)} & 0.94 \scriptsize{(0.01)} & \textbf{1.00 \scriptsize{(0.00)}} & 0.84 \scriptsize{(0.01)} & \textbf{1.00 \scriptsize{(0.00)}} \\
\midrule
\multirow{3}{*}{InternVL3 78B} & Simple & 0.98 \scriptsize{(0.02)} & 0.99 \scriptsize{(0.01)} & \textbf{1.00 \scriptsize{(0.00)}} & \textbf{1.00 \scriptsize{(0.00)}} & \textbf{1.00 \scriptsize{(0.00)}} \\
 & Medium & \textbf{1.00 \scriptsize{(0.00)}} & 0.98 \scriptsize{(0.00)} & \textbf{1.00 \scriptsize{(0.00)}} & \textbf{0.98 \scriptsize{(0.00)}} & \textbf{1.00 \scriptsize{(0.00)}} \\
 & Hard & \textbf{0.99 \scriptsize{(0.01)}} & \textbf{0.98 \scriptsize{(0.00)}} & \textbf{1.00 \scriptsize{(0.00)}} & 0.95 \scriptsize{(0.00)} & \textbf{1.00 \scriptsize{(0.00)}} \\
\midrule
\multirow{3}{*}{InternVL3 8B} & Simple & 0.77 \scriptsize{(0.04)} & 0.98 \scriptsize{(0.01)} & 0.99 \scriptsize{(0.00)} & 0.82 \scriptsize{(0.02)} & 0.99 \scriptsize{(0.00)} \\
 & Medium & 0.73 \scriptsize{(0.03)} & 0.95 \scriptsize{(0.01)} & 0.99 \scriptsize{(0.00)} & 0.80 \scriptsize{(0.01)} & 0.99 \scriptsize{(0.00)} \\
 & Hard & 0.61 \scriptsize{(0.03)} & 0.96 \scriptsize{(0.01)} & \textbf{1.00 \scriptsize{(0.00)}} & 0.71 \scriptsize{(0.01)} & \textbf{1.00 \scriptsize{(0.00)}} \\
\midrule
\multirow{3}{*}{InternVL3.5 30B A3B} & Simple & 0.91 \scriptsize{(0.03)} & 0.99 \scriptsize{(0.01)} & \textbf{1.00 \scriptsize{(0.00)}} & 0.97 \scriptsize{(0.01)} & \textbf{1.00 \scriptsize{(0.00)}} \\
 & Medium & 0.85 \scriptsize{(0.02)} & 0.98 \scriptsize{(0.00)} & \textbf{1.00 \scriptsize{(0.00)}} & 0.94 \scriptsize{(0.01)} & \textbf{1.00 \scriptsize{(0.00)}} \\
 & Hard & 0.76 \scriptsize{(0.02)} & 0.96 \scriptsize{(0.01)} & \textbf{1.00 \scriptsize{(0.00)}} & 0.90 \scriptsize{(0.01)} & \textbf{1.00 \scriptsize{(0.00)}} \\
\midrule
\multirow{3}{*}{InternVL3.5 38B} & Simple & \textbf{1.00 \scriptsize{(0.00)}} & \textbf{1.00 \scriptsize{(0.00)}} & \textbf{1.00 \scriptsize{(0.00)}} & 0.98 \scriptsize{(0.01)} & \textbf{1.00 \scriptsize{(0.00)}} \\
 & Medium & \textbf{1.00 \scriptsize{(0.00)}} & \textbf{0.99 \scriptsize{(0.00)}} & \textbf{1.00 \scriptsize{(0.00)}} & 0.94 \scriptsize{(0.01)} & \textbf{1.00 \scriptsize{(0.00)}} \\
 & Hard & 0.98 \scriptsize{(0.01)} & 0.97 \scriptsize{(0.00)} & \textbf{1.00 \scriptsize{(0.00)}} & 0.89 \scriptsize{(0.01)} & \textbf{1.00 \scriptsize{(0.00)}} \\
\midrule
\multirow{3}{*}{InternVL3.5 8B} & Simple & 0.98 \scriptsize{(0.01)} & 0.97 \scriptsize{(0.01)} & \textbf{1.00 \scriptsize{(0.00)}} & 0.98 \scriptsize{(0.01)} & \textbf{1.00 \scriptsize{(0.00)}} \\
 & Medium & 0.88 \scriptsize{(0.02)} & 0.97 \scriptsize{(0.00)} & \textbf{1.00 \scriptsize{(0.00)}} & \textbf{0.98 \scriptsize{(0.00)}} & \textbf{1.00 \scriptsize{(0.00)}} \\
 & Hard & 0.87 \scriptsize{(0.01)} & \textbf{0.98 \scriptsize{(0.00)}} & \textbf{1.00 \scriptsize{(0.00)}} & 0.96 \scriptsize{(0.00)} & \textbf{1.00 \scriptsize{(0.00)}} \\
\midrule
\multirow{3}{*}{LLaVA-Onevision 72B} & Simple & 0.76 \scriptsize{(0.04)} & 0.92 \scriptsize{(0.01)} & \textbf{1.00 \scriptsize{(0.00)}} & 0.93 \scriptsize{(0.01)} & \textbf{1.00 \scriptsize{(0.00)}} \\
 & Medium & 0.54 \scriptsize{(0.04)} & 0.92 \scriptsize{(0.01)} & 0.99 \scriptsize{(0.00)} & 0.95 \scriptsize{(0.01)} & \textbf{1.00 \scriptsize{(0.00)}} \\
 & Hard & 0.73 \scriptsize{(0.04)} & 0.93 \scriptsize{(0.01)} & \textbf{1.00 \scriptsize{(0.00)}} & 0.93 \scriptsize{(0.01)} & \textbf{1.00 \scriptsize{(0.00)}} \\
\midrule
\multirow{3}{*}{LLaVA-Onevision 7B} & Simple & 0.42 \scriptsize{(0.06)} & 0.89 \scriptsize{(0.02)} & \textbf{1.00 \scriptsize{(0.00)}} & 0.93 \scriptsize{(0.02)} & \textbf{1.00 \scriptsize{(0.00)}} \\
 & Medium & 0.41 \scriptsize{(0.04)} & 0.82 \scriptsize{(0.02)} & 0.96 \scriptsize{(0.01)} & 0.84 \scriptsize{(0.01)} & 0.98 \scriptsize{(0.01)} \\
 & Hard & 0.63 \scriptsize{(0.04)} & 0.92 \scriptsize{(0.01)} & 0.99 \scriptsize{(0.00)} & 0.83 \scriptsize{(0.01)} & 0.98 \scriptsize{(0.01)} \\
\midrule
\multirow{3}{*}{Mistral-Small-3.1 24B} & Simple & 0.96 \scriptsize{(0.02)} & 0.96 \scriptsize{(0.01)} & \textbf{1.00 \scriptsize{(0.00)}} & 0.95 \scriptsize{(0.01)} & \textbf{1.00 \scriptsize{(0.00)}} \\
 & Medium & 0.90 \scriptsize{(0.02)} & 0.95 \scriptsize{(0.01)} & \textbf{1.00 \scriptsize{(0.00)}} & 0.88 \scriptsize{(0.01)} & \textbf{1.00 \scriptsize{(0.00)}} \\
 & Hard & 0.87 \scriptsize{(0.02)} & 0.97 \scriptsize{(0.01)} & \textbf{1.00 \scriptsize{(0.00)}} & 0.87 \scriptsize{(0.01)} & \textbf{1.00 \scriptsize{(0.00)}} \\
\midrule
\multirow{3}{*}{Molmo 7B} & Simple & 0.28 \scriptsize{(0.05)} & 0.83 \scriptsize{(0.02)} & 0.78 \scriptsize{(0.02)} & 0.93 \scriptsize{(0.02)} & 0.71 \scriptsize{(0.02)} \\
 & Medium & 0.37 \scriptsize{(0.04)} & 0.87 \scriptsize{(0.01)} & 0.76 \scriptsize{(0.02)} & 0.80 \scriptsize{(0.02)} & 0.64 \scriptsize{(0.02)} \\
 & Hard & 0.49 \scriptsize{(0.04)} & 0.80 \scriptsize{(0.02)} & 0.69 \scriptsize{(0.02)} & 0.81 \scriptsize{(0.01)} & 0.61 \scriptsize{(0.02)} \\
\midrule
\multirow{3}{*}{Phi-4 Multimodal} & Simple & 0.09 \scriptsize{(0.03)} & 0.12 \scriptsize{(0.02)} & 0.04 \scriptsize{(0.01)} & 0.19 \scriptsize{(0.03)} & 0.04 \scriptsize{(0.01)} \\
 & Medium & 0.12 \scriptsize{(0.03)} & 0.29 \scriptsize{(0.02)} & 0.03 \scriptsize{(0.01)} & 0.26 \scriptsize{(0.02)} & 0.05 \scriptsize{(0.01)} \\
 & Hard & 0.11 \scriptsize{(0.03)} & 0.35 \scriptsize{(0.02)} & 0.06 \scriptsize{(0.01)} & 0.22 \scriptsize{(0.01)} & 0.05 \scriptsize{(0.01)} \\
\midrule
\multirow{3}{*}{Qwen2.5-VL 72B} & Simple & 0.92 \scriptsize{(0.03)} & 0.92 \scriptsize{(0.01)} & \textbf{1.00 \scriptsize{(0.00)}} & 0.97 \scriptsize{(0.01)} & \textbf{1.00 \scriptsize{(0.00)}} \\
 & Medium & 0.95 \scriptsize{(0.02)} & 0.98 \scriptsize{(0.00)} & \textbf{1.00 \scriptsize{(0.00)}} & \textbf{0.98 \scriptsize{(0.00)}} & \textbf{1.00 \scriptsize{(0.00)}} \\
 & Hard & 0.93 \scriptsize{(0.02)} & 0.94 \scriptsize{(0.01)} & \textbf{1.00 \scriptsize{(0.00)}} & 0.94 \scriptsize{(0.01)} & \textbf{1.00 \scriptsize{(0.00)}} \\
\midrule
\multirow{3}{*}{Qwen2.5-VL 7B} & Simple & 0.78 \scriptsize{(0.04)} & 0.98 \scriptsize{(0.01)} & \textbf{1.00 \scriptsize{(0.00)}} & 0.95 \scriptsize{(0.01)} & \textbf{1.00 \scriptsize{(0.00)}} \\
 & Medium & 0.72 \scriptsize{(0.03)} & 0.95 \scriptsize{(0.01)} & \textbf{1.00 \scriptsize{(0.00)}} & 0.77 \scriptsize{(0.01)} & \textbf{1.00 \scriptsize{(0.00)}} \\
 & Hard & 0.76 \scriptsize{(0.02)} & 0.97 \scriptsize{(0.00)} & \textbf{1.00 \scriptsize{(0.00)}} & 0.77 \scriptsize{(0.01)} & \textbf{1.00 \scriptsize{(0.00)}} \\
\midrule
\multirow{3}{*}{Qwen3-VL 30B A3B} & Simple & 0.96 \scriptsize{(0.02)} & 0.91 \scriptsize{(0.01)} & \textbf{1.00 \scriptsize{(0.00)}} & 0.96 \scriptsize{(0.01)} & \textbf{1.00 \scriptsize{(0.00)}} \\
 & Medium & 0.93 \scriptsize{(0.01)} & 0.93 \scriptsize{(0.01)} & \textbf{1.00 \scriptsize{(0.00)}} & \textbf{0.98 \scriptsize{(0.00)}} & \textbf{1.00 \scriptsize{(0.00)}} \\
 & Hard & 0.95 \scriptsize{(0.01)} & 0.92 \scriptsize{(0.01)} & \textbf{1.00 \scriptsize{(0.00)}} & \textbf{0.97 \scriptsize{(0.00)}} & \textbf{1.00 \scriptsize{(0.00)}} \\
\midrule
\multirow{3}{*}{Qwen3-VL 32B} & Simple & 0.90 \scriptsize{(0.03)} & 0.93 \scriptsize{(0.01)} & \textbf{1.00 \scriptsize{(0.00)}} & 0.97 \scriptsize{(0.01)} & \textbf{1.00 \scriptsize{(0.00)}} \\
 & Medium & 0.93 \scriptsize{(0.02)} & 0.96 \scriptsize{(0.01)} & \textbf{1.00 \scriptsize{(0.00)}} & 0.97 \scriptsize{(0.01)} & \textbf{1.00 \scriptsize{(0.00)}} \\
 & Hard & 0.92 \scriptsize{(0.02)} & 0.94 \scriptsize{(0.01)} & \textbf{1.00 \scriptsize{(0.00)}} & 0.96 \scriptsize{(0.01)} & \textbf{1.00 \scriptsize{(0.00)}} \\
\midrule
\multirow{3}{*}{Qwen3-VL 8B} & Simple & 0.98 \scriptsize{(0.01)} & 0.97 \scriptsize{(0.01)} & \textbf{1.00 \scriptsize{(0.00)}} & 0.97 \scriptsize{(0.01)} & \textbf{1.00 \scriptsize{(0.00)}} \\
 & Medium & 0.97 \scriptsize{(0.01)} & 0.95 \scriptsize{(0.01)} & \textbf{1.00 \scriptsize{(0.00)}} & 0.96 \scriptsize{(0.00)} & \textbf{1.00 \scriptsize{(0.00)}} \\
 & Hard & 0.96 \scriptsize{(0.01)} & 0.95 \scriptsize{(0.00)} & \textbf{1.00 \scriptsize{(0.00)}} & 0.94 \scriptsize{(0.00)} & \textbf{1.00 \scriptsize{(0.00)}} \\
\bottomrule
\end{tabular}}
\end{table*}

\begin{table*}[h]
\centering
\caption{\textbf{Individual Predicate Accuracy for Ground on ViPlan-HH.} The table shows the accuracy for each predicate in each split. Bolded values show the best accuracy for each predicate and split. Standard error of the mean is reported in parenthesis.}
\label{tab:ViPLan-HH_pred_no_cot}
\resizebox{!}{0.46\textheight}{
\begin{tabular}{l l c c c c c c}
\toprule
\textbf{Model} & \textbf{Split} & \texttt{holding} & \texttt{inside} & \texttt{nextto} & \texttt{ontop} & \texttt{open} & \texttt{reachable} \\
\midrule
\multirow{3}{*}{AyaVision 32B} & Simple & 0.79 \scriptsize{(0.04)} & 0.79 \scriptsize{(0.04)} & 0.34 \scriptsize{(0.03)} & 0.78 \scriptsize{(0.03)} & 0.30 \scriptsize{(0.05)} & 0.71 \scriptsize{(0.03)} \\
 & Medium & 0.86 \scriptsize{(0.03)} & 0.76 \scriptsize{(0.03)} & 0.40 \scriptsize{(0.01)} & 0.87 \scriptsize{(0.01)} & 0.47 \scriptsize{(0.07)} & 0.64 \scriptsize{(0.03)} \\
 & Hard & 0.92 \scriptsize{(0.02)} & 0.57 \scriptsize{(0.03)} & 0.55 \scriptsize{(0.02)} & 0.78 \scriptsize{(0.01)} & 0.49 \scriptsize{(0.04)} & 0.58 \scriptsize{(0.02)} \\
\midrule
\multirow{3}{*}{AyaVision 8B} & Simple & 0.59 \scriptsize{(0.05)} & 0.61 \scriptsize{(0.05)} & 0.12 \scriptsize{(0.02)} & 0.57 \scriptsize{(0.03)} & 0.39 \scriptsize{(0.06)} & 0.68 \scriptsize{(0.04)} \\
 & Medium & 0.64 \scriptsize{(0.03)} & 0.52 \scriptsize{(0.03)} & 0.14 \scriptsize{(0.01)} & 0.67 \scriptsize{(0.01)} & 0.33 \scriptsize{(0.05)} & 0.63 \scriptsize{(0.03)} \\
 & Hard & 0.57 \scriptsize{(0.04)} & 0.42 \scriptsize{(0.03)} & 0.19 \scriptsize{(0.02)} & 0.63 \scriptsize{(0.02)} & 0.79 \scriptsize{(0.04)} & 0.38 \scriptsize{(0.03)} \\
\midrule
\multirow{3}{*}{Cosmos-Reason2 8B} & Simple & 0.75 \scriptsize{(0.04)} & 0.63 \scriptsize{(0.05)} & 0.34 \scriptsize{(0.03)} & 0.61 \scriptsize{(0.03)} & 0.51 \scriptsize{(0.05)} & 0.76 \scriptsize{(0.03)} \\
 & Medium & 0.78 \scriptsize{(0.02)} & 0.88 \scriptsize{(0.02)} & 0.41 \scriptsize{(0.01)} & 0.83 \scriptsize{(0.01)} & \textbf{0.79 \scriptsize{(0.05)}} & 0.65 \scriptsize{(0.02)} \\
 & Hard & 0.85 \scriptsize{(0.03)} & 0.71 \scriptsize{(0.03)} & 0.29 \scriptsize{(0.02)} & 0.55 \scriptsize{(0.02)} & 0.76 \scriptsize{(0.04)} & 0.47 \scriptsize{(0.03)} \\
\midrule
\multirow{3}{*}{DeepSeek-VL2} & Simple & 0.91 \scriptsize{(0.02)} & 0.81 \scriptsize{(0.03)} & 0.54 \scriptsize{(0.02)} & 0.85 \scriptsize{(0.02)} & 0.46 \scriptsize{(0.05)} & 0.69 \scriptsize{(0.03)} \\
 & Medium & 0.98 \scriptsize{(0.00)} & 0.90 \scriptsize{(0.02)} & 0.59 \scriptsize{(0.00)} & 0.92 \scriptsize{(0.00)} & 0.77 \scriptsize{(0.04)} & 0.50 \scriptsize{(0.01)} \\
 & Hard & 0.91 \scriptsize{(0.01)} & 0.93 \scriptsize{(0.01)} & 0.45 \scriptsize{(0.01)} & 0.89 \scriptsize{(0.01)} & \textbf{0.88 \scriptsize{(0.02)}} & \textbf{0.83 \scriptsize{(0.01)}} \\
\midrule
\multirow{3}{*}{GPT-4.1} & Simple & 0.79 \scriptsize{(0.03)} & 0.80 \scriptsize{(0.03)} & 0.52 \scriptsize{(0.02)} & 0.83 \scriptsize{(0.02)} & 0.50 \scriptsize{(0.04)} & 0.66 \scriptsize{(0.03)} \\
 & Medium & 0.81 \scriptsize{(0.02)} & 0.94 \scriptsize{(0.02)} & 0.55 \scriptsize{(0.01)} & 0.89 \scriptsize{(0.01)} & 0.74 \scriptsize{(0.04)} & 0.72 \scriptsize{(0.02)} \\
 & Hard & 0.86 \scriptsize{(0.02)} & 0.79 \scriptsize{(0.03)} & 0.56 \scriptsize{(0.02)} & 0.86 \scriptsize{(0.01)} & 0.77 \scriptsize{(0.04)} & 0.49 \scriptsize{(0.03)} \\
\midrule
\multirow{3}{*}{GPT-4.1 Nano} & Simple & 0.85 \scriptsize{(0.04)} & 0.66 \scriptsize{(0.05)} & 0.18 \scriptsize{(0.02)} & 0.73 \scriptsize{(0.02)} & 0.49 \scriptsize{(0.06)} & 0.67 \scriptsize{(0.04)} \\
 & Medium & 0.89 \scriptsize{(0.02)} & 0.83 \scriptsize{(0.03)} & 0.30 \scriptsize{(0.01)} & 0.84 \scriptsize{(0.01)} & 0.62 \scriptsize{(0.06)} & 0.55 \scriptsize{(0.03)} \\
 & Hard & 0.86 \scriptsize{(0.04)} & 0.74 \scriptsize{(0.03)} & 0.21 \scriptsize{(0.02)} & 0.69 \scriptsize{(0.02)} & 0.64 \scriptsize{(0.06)} & 0.53 \scriptsize{(0.04)} \\
\midrule
\multirow{3}{*}{GPT-5.2} & Simple & 0.86 \scriptsize{(0.03)} & 0.67 \scriptsize{(0.04)} & 0.52 \scriptsize{(0.02)} & 0.84 \scriptsize{(0.02)} & 0.59 \scriptsize{(0.04)} & 0.49 \scriptsize{(0.03)} \\
 & Medium & 0.95 \scriptsize{(0.01)} & 0.94 \scriptsize{(0.01)} & 0.64 \scriptsize{(0.01)} & 0.94 \scriptsize{(0.00)} & 0.53 \scriptsize{(0.03)} & 0.42 \scriptsize{(0.01)} \\
 & Hard & 0.98 \scriptsize{(0.00)} & 0.88 \scriptsize{(0.01)} & 0.73 \scriptsize{(0.00)} & \textbf{0.94 \scriptsize{(0.00)}} & 0.53 \scriptsize{(0.02)} & 0.63 \scriptsize{(0.01)} \\
\midrule
\multirow{3}{*}{Gemma-3 12B} & Simple & 0.63 \scriptsize{(0.05)} & 0.71 \scriptsize{(0.07)} & 0.23 \scriptsize{(0.03)} & 0.65 \scriptsize{(0.03)} & 0.27 \scriptsize{(0.05)} & 0.70 \scriptsize{(0.04)} \\
 & Medium & 0.71 \scriptsize{(0.03)} & 0.79 \scriptsize{(0.03)} & 0.61 \scriptsize{(0.01)} & 0.76 \scriptsize{(0.01)} & 0.44 \scriptsize{(0.06)} & 0.64 \scriptsize{(0.02)} \\
 & Hard & 0.63 \scriptsize{(0.04)} & 0.76 \scriptsize{(0.03)} & 0.39 \scriptsize{(0.02)} & 0.76 \scriptsize{(0.02)} & 0.68 \scriptsize{(0.05)} & 0.43 \scriptsize{(0.03)} \\
\midrule
\multirow{3}{*}{Gemma-3 27B} & Simple & 0.80 \scriptsize{(0.03)} & 0.80 \scriptsize{(0.04)} & 0.52 \scriptsize{(0.03)} & \textbf{0.92 \scriptsize{(0.01)}} & 0.46 \scriptsize{(0.05)} & 0.50 \scriptsize{(0.03)} \\
 & Medium & 0.86 \scriptsize{(0.01)} & 0.95 \scriptsize{(0.01)} & 0.63 \scriptsize{(0.01)} & 0.94 \scriptsize{(0.00)} & 0.54 \scriptsize{(0.03)} & 0.55 \scriptsize{(0.02)} \\
 & Hard & 0.87 \scriptsize{(0.01)} & 0.85 \scriptsize{(0.01)} & 0.63 \scriptsize{(0.01)} & 0.93 \scriptsize{(0.00)} & 0.38 \scriptsize{(0.03)} & 0.50 \scriptsize{(0.01)} \\
\midrule
\multirow{3}{*}{InternVL3 78B} & Simple & 0.89 \scriptsize{(0.03)} & 0.82 \scriptsize{(0.03)} & 0.59 \scriptsize{(0.02)} & 0.77 \scriptsize{(0.02)} & 0.38 \scriptsize{(0.04)} & 0.76 \scriptsize{(0.03)} \\
 & Medium & 0.86 \scriptsize{(0.02)} & 0.87 \scriptsize{(0.02)} & 0.48 \scriptsize{(0.01)} & 0.83 \scriptsize{(0.01)} & 0.37 \scriptsize{(0.04)} & 0.74 \scriptsize{(0.02)} \\
 & Hard & 0.83 \scriptsize{(0.03)} & 0.83 \scriptsize{(0.03)} & 0.43 \scriptsize{(0.02)} & 0.83 \scriptsize{(0.02)} & 0.79 \scriptsize{(0.05)} & 0.42 \scriptsize{(0.03)} \\
\midrule
\multirow{3}{*}{InternVL3 8B} & Simple & 0.91 \scriptsize{(0.02)} & 0.83 \scriptsize{(0.02)} & 0.61 \scriptsize{(0.02)} & 0.84 \scriptsize{(0.01)} & 0.36 \scriptsize{(0.04)} & 0.76 \scriptsize{(0.02)} \\
 & Medium & 0.91 \scriptsize{(0.01)} & 0.89 \scriptsize{(0.01)} & 0.65 \scriptsize{(0.01)} & 0.91 \scriptsize{(0.01)} & 0.55 \scriptsize{(0.03)} & 0.71 \scriptsize{(0.02)} \\
 & Hard & 0.90 \scriptsize{(0.02)} & 0.93 \scriptsize{(0.02)} & 0.71 \scriptsize{(0.01)} & 0.90 \scriptsize{(0.01)} & 0.62 \scriptsize{(0.05)} & 0.44 \scriptsize{(0.02)} \\
\midrule
\multirow{3}{*}{InternVL3.5 30B A3B} & Simple & 0.93 \scriptsize{(0.03)} & 0.87 \scriptsize{(0.03)} & 0.58 \scriptsize{(0.03)} & 0.76 \scriptsize{(0.03)} & 0.39 \scriptsize{(0.05)} & 0.63 \scriptsize{(0.03)} \\
 & Medium & 0.92 \scriptsize{(0.01)} & 0.90 \scriptsize{(0.02)} & 0.50 \scriptsize{(0.01)} & 0.92 \scriptsize{(0.01)} & 0.56 \scriptsize{(0.04)} & 0.67 \scriptsize{(0.02)} \\
 & Hard & 0.87 \scriptsize{(0.02)} & 0.59 \scriptsize{(0.03)} & 0.58 \scriptsize{(0.01)} & 0.81 \scriptsize{(0.01)} & 0.44 \scriptsize{(0.04)} & 0.51 \scriptsize{(0.02)} \\
\midrule
\multirow{3}{*}{InternVL3.5 38B} & Simple & 0.86 \scriptsize{(0.03)} & 0.56 \scriptsize{(0.06)} & 0.32 \scriptsize{(0.03)} & 0.55 \scriptsize{(0.03)} & 0.40 \scriptsize{(0.04)} & 0.60 \scriptsize{(0.03)} \\
 & Medium & 0.86 \scriptsize{(0.02)} & 0.78 \scriptsize{(0.02)} & 0.30 \scriptsize{(0.01)} & 0.64 \scriptsize{(0.01)} & 0.68 \scriptsize{(0.05)} & 0.68 \scriptsize{(0.02)} \\
 & Hard & 0.66 \scriptsize{(0.03)} & 0.61 \scriptsize{(0.04)} & 0.29 \scriptsize{(0.01)} & 0.54 \scriptsize{(0.01)} & 0.74 \scriptsize{(0.05)} & 0.63 \scriptsize{(0.03)} \\
\midrule
\multirow{3}{*}{InternVL3.5 8B} & Simple & 0.81 \scriptsize{(0.04)} & 0.73 \scriptsize{(0.04)} & 0.46 \scriptsize{(0.03)} & 0.87 \scriptsize{(0.02)} & 0.53 \scriptsize{(0.06)} & \textbf{0.79 \scriptsize{(0.03)}} \\
 & Medium & 0.91 \scriptsize{(0.02)} & 0.95 \scriptsize{(0.01)} & 0.48 \scriptsize{(0.01)} & 0.91 \scriptsize{(0.01)} & 0.73 \scriptsize{(0.04)} & 0.67 \scriptsize{(0.02)} \\
 & Hard & 0.94 \scriptsize{(0.01)} & 0.84 \scriptsize{(0.03)} & 0.39 \scriptsize{(0.01)} & 0.92 \scriptsize{(0.01)} & 0.43 \scriptsize{(0.05)} & 0.71 \scriptsize{(0.02)} \\
\midrule
\multirow{3}{*}{LLaVA-Onevision 72B} & Simple & 0.87 \scriptsize{(0.04)} & 0.48 \scriptsize{(0.09)} & 0.50 \scriptsize{(0.04)} & 0.84 \scriptsize{(0.03)} & 0.20 \scriptsize{(0.06)} & 0.64 \scriptsize{(0.04)} \\
 & Medium & 0.94 \scriptsize{(0.01)} & 0.91 \scriptsize{(0.01)} & 0.68 \scriptsize{(0.01)} & 0.94 \scriptsize{(0.01)} & 0.60 \scriptsize{(0.03)} & 0.63 \scriptsize{(0.02)} \\
 & Hard & 0.96 \scriptsize{(0.01)} & 0.81 \scriptsize{(0.03)} & 0.59 \scriptsize{(0.01)} & 0.88 \scriptsize{(0.01)} & 0.60 \scriptsize{(0.06)} & 0.68 \scriptsize{(0.03)} \\
\midrule
\multirow{3}{*}{LLaVA-Onevision 7B} & Simple & 0.87 \scriptsize{(0.03)} & 0.82 \scriptsize{(0.03)} & 0.28 \scriptsize{(0.02)} & 0.81 \scriptsize{(0.02)} & 0.64 \scriptsize{(0.04)} & 0.44 \scriptsize{(0.03)} \\
 & Medium & 0.89 \scriptsize{(0.02)} & 0.96 \scriptsize{(0.01)} & 0.41 \scriptsize{(0.01)} & 0.88 \scriptsize{(0.01)} & 0.73 \scriptsize{(0.03)} & 0.52 \scriptsize{(0.02)} \\
 & Hard & 0.94 \scriptsize{(0.01)} & 0.88 \scriptsize{(0.03)} & 0.45 \scriptsize{(0.01)} & \textbf{0.94 \scriptsize{(0.01)}} & 0.84 \scriptsize{(0.04)} & 0.45 \scriptsize{(0.03)} \\
\midrule
\multirow{3}{*}{Mistral-Small-3.1 24B} & Simple & \textbf{0.98 \scriptsize{(0.01)}} & 0.84 \scriptsize{(0.04)} & 0.72 \scriptsize{(0.02)} & 0.87 \scriptsize{(0.01)} & 0.67 \scriptsize{(0.03)} & 0.28 \scriptsize{(0.02)} \\
 & Medium & \textbf{1.00 \scriptsize{(0.00)}} & \textbf{1.00 \scriptsize{(0.00)}} & 0.79 \scriptsize{(0.01)} & 0.94 \scriptsize{(0.00)} & 0.73 \scriptsize{(0.02)} & 0.40 \scriptsize{(0.01)} \\
 & Hard & \textbf{0.99 \scriptsize{(0.00)}} & \textbf{0.96 \scriptsize{(0.00)}} & 0.78 \scriptsize{(0.00)} & 0.93 \scriptsize{(0.00)} & 0.72 \scriptsize{(0.01)} & 0.40 \scriptsize{(0.01)} \\
\midrule
\multirow{3}{*}{Molmo 7B} & Simple & 0.88 \scriptsize{(0.03)} & 0.75 \scriptsize{(0.04)} & 0.51 \scriptsize{(0.03)} & 0.58 \scriptsize{(0.03)} & 0.29 \scriptsize{(0.05)} & 0.65 \scriptsize{(0.04)} \\
 & Medium & 0.86 \scriptsize{(0.02)} & 0.60 \scriptsize{(0.03)} & 0.55 \scriptsize{(0.01)} & 0.68 \scriptsize{(0.01)} & 0.54 \scriptsize{(0.04)} & \textbf{0.76 \scriptsize{(0.02)}} \\
 & Hard & 0.90 \scriptsize{(0.01)} & 0.69 \scriptsize{(0.01)} & 0.62 \scriptsize{(0.00)} & 0.70 \scriptsize{(0.00)} & 0.61 \scriptsize{(0.03)} & 0.71 \scriptsize{(0.01)} \\
\midrule
\multirow{3}{*}{Phi-4 Multimodal} & Simple & 0.51 \scriptsize{(0.06)} & 0.39 \scriptsize{(0.05)} & 0.20 \scriptsize{(0.02)} & 0.14 \scriptsize{(0.02)} & 0.33 \scriptsize{(0.06)} & 0.63 \scriptsize{(0.04)} \\
 & Medium & 0.69 \scriptsize{(0.03)} & 0.38 \scriptsize{(0.03)} & 0.11 \scriptsize{(0.01)} & 0.18 \scriptsize{(0.01)} & 0.26 \scriptsize{(0.04)} & 0.57 \scriptsize{(0.02)} \\
 & Hard & 0.56 \scriptsize{(0.03)} & 0.43 \scriptsize{(0.03)} & 0.24 \scriptsize{(0.01)} & 0.18 \scriptsize{(0.01)} & 0.59 \scriptsize{(0.04)} & 0.63 \scriptsize{(0.03)} \\
\midrule
\multirow{3}{*}{Qwen2.5-VL 72B} & Simple & 0.95 \scriptsize{(0.01)} & 0.75 \scriptsize{(0.05)} & \textbf{0.79 \scriptsize{(0.02)}} & 0.83 \scriptsize{(0.02)} & 0.60 \scriptsize{(0.04)} & 0.35 \scriptsize{(0.02)} \\
 & Medium & 0.95 \scriptsize{(0.01)} & 0.88 \scriptsize{(0.01)} & \textbf{0.81 \scriptsize{(0.01)}} & 0.93 \scriptsize{(0.01)} & 0.51 \scriptsize{(0.03)} & 0.36 \scriptsize{(0.01)} \\
 & Hard & 0.95 \scriptsize{(0.01)} & 0.83 \scriptsize{(0.01)} & 0.72 \scriptsize{(0.00)} & 0.91 \scriptsize{(0.00)} & 0.64 \scriptsize{(0.02)} & 0.39 \scriptsize{(0.01)} \\
\midrule
\multirow{3}{*}{Qwen2.5-VL 7B} & Simple & 0.96 \scriptsize{(0.01)} & \textbf{0.89 \scriptsize{(0.04)}} & 0.73 \scriptsize{(0.02)} & 0.86 \scriptsize{(0.02)} & \textbf{0.72 \scriptsize{(0.03)}} & 0.47 \scriptsize{(0.02)} \\
 & Medium & 0.91 \scriptsize{(0.01)} & 0.92 \scriptsize{(0.01)} & 0.77 \scriptsize{(0.01)} & \textbf{0.96 \scriptsize{(0.00)}} & 0.44 \scriptsize{(0.03)} & 0.43 \scriptsize{(0.01)} \\
 & Hard & 0.96 \scriptsize{(0.01)} & 0.90 \scriptsize{(0.01)} & \textbf{0.80 \scriptsize{(0.00)}} & 0.93 \scriptsize{(0.00)} & 0.60 \scriptsize{(0.02)} & 0.37 \scriptsize{(0.01)} \\
\midrule
\multirow{3}{*}{Qwen3-VL 30B A3B} & Simple & 0.92 \scriptsize{(0.02)} & 0.80 \scriptsize{(0.02)} & 0.71 \scriptsize{(0.02)} & 0.89 \scriptsize{(0.01)} & 0.35 \scriptsize{(0.04)} & 0.64 \scriptsize{(0.03)} \\
 & Medium & 0.92 \scriptsize{(0.01)} & 0.83 \scriptsize{(0.02)} & 0.53 \scriptsize{(0.01)} & 0.86 \scriptsize{(0.01)} & 0.43 \scriptsize{(0.03)} & 0.72 \scriptsize{(0.02)} \\
 & Hard & 0.90 \scriptsize{(0.02)} & 0.82 \scriptsize{(0.02)} & 0.59 \scriptsize{(0.02)} & 0.82 \scriptsize{(0.01)} & 0.76 \scriptsize{(0.04)} & 0.51 \scriptsize{(0.03)} \\
\midrule
\multirow{3}{*}{Qwen3-VL 32B} & Simple & 0.91 \scriptsize{(0.02)} & 0.73 \scriptsize{(0.05)} & 0.56 \scriptsize{(0.03)} & 0.89 \scriptsize{(0.02)} & 0.52 \scriptsize{(0.04)} & 0.67 \scriptsize{(0.03)} \\
 & Medium & 0.90 \scriptsize{(0.01)} & 0.83 \scriptsize{(0.02)} & 0.55 \scriptsize{(0.01)} & 0.88 \scriptsize{(0.01)} & 0.48 \scriptsize{(0.04)} & 0.55 \scriptsize{(0.02)} \\
 & Hard & 0.90 \scriptsize{(0.02)} & 0.82 \scriptsize{(0.02)} & 0.43 \scriptsize{(0.01)} & 0.83 \scriptsize{(0.01)} & 0.71 \scriptsize{(0.03)} & 0.65 \scriptsize{(0.02)} \\
\midrule
\multirow{3}{*}{Qwen3-VL 8B} & Simple & 0.87 \scriptsize{(0.02)} & 0.85 \scriptsize{(0.02)} & 0.46 \scriptsize{(0.02)} & 0.83 \scriptsize{(0.01)} & 0.27 \scriptsize{(0.03)} & 0.62 \scriptsize{(0.02)} \\
 & Medium & 0.90 \scriptsize{(0.01)} & 0.91 \scriptsize{(0.01)} & 0.57 \scriptsize{(0.01)} & 0.88 \scriptsize{(0.01)} & 0.45 \scriptsize{(0.04)} & 0.54 \scriptsize{(0.02)} \\
 & Hard & 0.91 \scriptsize{(0.02)} & 0.88 \scriptsize{(0.02)} & 0.52 \scriptsize{(0.01)} & 0.87 \scriptsize{(0.01)} & 0.64 \scriptsize{(0.04)} & 0.72 \scriptsize{(0.02)} \\
\bottomrule
\end{tabular}}
\end{table*}

\begin{table*}[h]
\centering
\caption{\textbf{Individual Predicate Accuracy for Ground + CoT on ViPlan-HH.} The table shows the accuracy for each predicate in each split. Bolded values show the best accuracy for each predicate and split. Standard error of the mean is reported in parenthesis.}
\label{tab:ViPLan-HH_pred_cot}
\resizebox{!}{0.46\textheight}{
\begin{tabular}{l l c c c c c c}
\toprule
\textbf{Model} & \textbf{Split} & \texttt{holding} & \texttt{inside} & \texttt{nextto} & \texttt{ontop} & \texttt{open} & \texttt{reachable} \\
\midrule
\multirow{3}{*}{AyaVision 32B} & Simple & 0.88 \scriptsize{(0.03)} & 0.85 \scriptsize{(0.03)} & \textbf{0.66 \scriptsize{(0.02)}} & 0.82 \scriptsize{(0.02)} & 0.51 \scriptsize{(0.04)} & 0.62 \scriptsize{(0.03)} \\
 & Medium & 0.86 \scriptsize{(0.02)} & 0.93 \scriptsize{(0.01)} & 0.58 \scriptsize{(0.01)} & 0.88 \scriptsize{(0.01)} & \textbf{0.82 \scriptsize{(0.03)}} & 0.74 \scriptsize{(0.02)} \\
 & Hard & 0.87 \scriptsize{(0.02)} & 0.80 \scriptsize{(0.02)} & 0.59 \scriptsize{(0.01)} & 0.92 \scriptsize{(0.01)} & 0.80 \scriptsize{(0.03)} & 0.52 \scriptsize{(0.02)} \\
\midrule
\multirow{3}{*}{AyaVision 8B} & Simple & 0.62 \scriptsize{(0.04)} & 0.70 \scriptsize{(0.03)} & 0.47 \scriptsize{(0.02)} & \textbf{0.92 \scriptsize{(0.01)}} & 0.47 \scriptsize{(0.04)} & 0.67 \scriptsize{(0.03)} \\
 & Medium & 0.52 \scriptsize{(0.03)} & 0.72 \scriptsize{(0.03)} & 0.57 \scriptsize{(0.01)} & 0.90 \scriptsize{(0.01)} & 0.45 \scriptsize{(0.05)} & 0.59 \scriptsize{(0.02)} \\
 & Hard & 0.57 \scriptsize{(0.02)} & 0.85 \scriptsize{(0.01)} & 0.49 \scriptsize{(0.01)} & 0.89 \scriptsize{(0.00)} & 0.70 \scriptsize{(0.03)} & 0.68 \scriptsize{(0.02)} \\
\midrule
\multirow{3}{*}{Cosmos-Reason2 8B} & Simple & 0.89 \scriptsize{(0.03)} & 0.73 \scriptsize{(0.04)} & 0.60 \scriptsize{(0.03)} & 0.87 \scriptsize{(0.02)} & 0.57 \scriptsize{(0.04)} & 0.51 \scriptsize{(0.03)} \\
 & Medium & \textbf{0.98 \scriptsize{(0.01)}} & \textbf{0.98 \scriptsize{(0.01)}} & 0.55 \scriptsize{(0.02)} & 0.82 \scriptsize{(0.01)} & 0.65 \scriptsize{(0.03)} & 0.52 \scriptsize{(0.02)} \\
 & Hard & 0.79 \scriptsize{(0.02)} & 0.67 \scriptsize{(0.02)} & 0.73 \scriptsize{(0.01)} & 0.91 \scriptsize{(0.01)} & 0.68 \scriptsize{(0.03)} & 0.46 \scriptsize{(0.02)} \\
\midrule
\multirow{3}{*}{DeepSeek-VL2} & Simple & 0.72 \scriptsize{(0.05)} & 0.79 \scriptsize{(0.06)} & 0.61 \scriptsize{(0.03)} & 0.76 \scriptsize{(0.03)} & 0.53 \scriptsize{(0.06)} & 0.74 \scriptsize{(0.03)} \\
 & Medium & 0.67 \scriptsize{(0.02)} & 0.97 \scriptsize{(0.01)} & 0.51 \scriptsize{(0.01)} & 0.83 \scriptsize{(0.01)} & 0.61 \scriptsize{(0.04)} & 0.80 \scriptsize{(0.01)} \\
 & Hard & 0.42 \scriptsize{(0.03)} & 0.85 \scriptsize{(0.02)} & 0.61 \scriptsize{(0.01)} & 0.86 \scriptsize{(0.01)} & 0.88 \scriptsize{(0.02)} & 0.34 \scriptsize{(0.02)} \\
\midrule
\multirow{3}{*}{GPT-4.1} & Simple & 0.87 \scriptsize{(0.03)} & 0.68 \scriptsize{(0.04)} & 0.60 \scriptsize{(0.02)} & 0.81 \scriptsize{(0.02)} & 0.52 \scriptsize{(0.05)} & 0.66 \scriptsize{(0.03)} \\
 & Medium & 0.85 \scriptsize{(0.02)} & 0.91 \scriptsize{(0.01)} & 0.51 \scriptsize{(0.01)} & 0.87 \scriptsize{(0.01)} & 0.70 \scriptsize{(0.03)} & 0.69 \scriptsize{(0.02)} \\
 & Hard & 0.92 \scriptsize{(0.01)} & 0.78 \scriptsize{(0.02)} & 0.55 \scriptsize{(0.01)} & 0.85 \scriptsize{(0.00)} & 0.63 \scriptsize{(0.04)} & 0.64 \scriptsize{(0.02)} \\
\midrule
\multirow{3}{*}{GPT-4.1 Nano} & Simple & 0.79 \scriptsize{(0.04)} & 0.88 \scriptsize{(0.03)} & 0.60 \scriptsize{(0.03)} & 0.78 \scriptsize{(0.02)} & 0.34 \scriptsize{(0.05)} & 0.60 \scriptsize{(0.03)} \\
 & Medium & 0.82 \scriptsize{(0.02)} & 0.92 \scriptsize{(0.01)} & 0.57 \scriptsize{(0.01)} & 0.85 \scriptsize{(0.01)} & 0.55 \scriptsize{(0.04)} & 0.65 \scriptsize{(0.02)} \\
 & Hard & 0.71 \scriptsize{(0.03)} & 0.78 \scriptsize{(0.02)} & 0.58 \scriptsize{(0.01)} & 0.73 \scriptsize{(0.01)} & 0.53 \scriptsize{(0.03)} & 0.62 \scriptsize{(0.02)} \\
\midrule
\multirow{3}{*}{Gemma-3 12B} & Simple & 0.64 \scriptsize{(0.05)} & 0.83 \scriptsize{(0.04)} & 0.37 \scriptsize{(0.03)} & 0.83 \scriptsize{(0.02)} & 0.58 \scriptsize{(0.05)} & 0.71 \scriptsize{(0.03)} \\
 & Medium & 0.55 \scriptsize{(0.03)} & 0.97 \scriptsize{(0.01)} & 0.38 \scriptsize{(0.01)} & 0.69 \scriptsize{(0.01)} & 0.70 \scriptsize{(0.04)} & 0.59 \scriptsize{(0.02)} \\
 & Hard & 0.48 \scriptsize{(0.03)} & 0.89 \scriptsize{(0.01)} & 0.38 \scriptsize{(0.01)} & 0.84 \scriptsize{(0.01)} & 0.38 \scriptsize{(0.03)} & 0.41 \scriptsize{(0.02)} \\
\midrule
\multirow{3}{*}{Gemma-3 27B} & Simple & 0.83 \scriptsize{(0.03)} & 0.73 \scriptsize{(0.03)} & 0.61 \scriptsize{(0.02)} & 0.91 \scriptsize{(0.01)} & 0.43 \scriptsize{(0.04)} & 0.50 \scriptsize{(0.02)} \\
 & Medium & 0.64 \scriptsize{(0.02)} & 0.97 \scriptsize{(0.01)} & 0.39 \scriptsize{(0.01)} & 0.89 \scriptsize{(0.01)} & 0.69 \scriptsize{(0.03)} & \textbf{0.81 \scriptsize{(0.01)}} \\
 & Hard & 0.79 \scriptsize{(0.02)} & 0.96 \scriptsize{(0.01)} & 0.33 \scriptsize{(0.01)} & 0.89 \scriptsize{(0.00)} & 0.80 \scriptsize{(0.03)} & 0.82 \scriptsize{(0.01)} \\
\midrule
\multirow{3}{*}{InternVL3 78B} & Simple & 0.87 \scriptsize{(0.03)} & 0.73 \scriptsize{(0.05)} & 0.50 \scriptsize{(0.03)} & 0.75 \scriptsize{(0.02)} & 0.41 \scriptsize{(0.05)} & 0.72 \scriptsize{(0.03)} \\
 & Medium & 0.76 \scriptsize{(0.02)} & 0.95 \scriptsize{(0.01)} & 0.45 \scriptsize{(0.01)} & 0.86 \scriptsize{(0.01)} & 0.75 \scriptsize{(0.04)} & 0.73 \scriptsize{(0.02)} \\
 & Hard & 0.95 \scriptsize{(0.01)} & 0.94 \scriptsize{(0.01)} & 0.40 \scriptsize{(0.01)} & 0.86 \scriptsize{(0.01)} & 0.70 \scriptsize{(0.04)} & \textbf{0.92 \scriptsize{(0.01)}} \\
\midrule
\multirow{3}{*}{InternVL3 8B} & Simple & 0.73 \scriptsize{(0.04)} & 0.76 \scriptsize{(0.04)} & 0.53 \scriptsize{(0.02)} & 0.67 \scriptsize{(0.02)} & 0.36 \scriptsize{(0.05)} & 0.68 \scriptsize{(0.03)} \\
 & Medium & 0.67 \scriptsize{(0.02)} & 0.85 \scriptsize{(0.02)} & 0.45 \scriptsize{(0.01)} & 0.65 \scriptsize{(0.01)} & 0.36 \scriptsize{(0.04)} & 0.75 \scriptsize{(0.02)} \\
 & Hard & 0.82 \scriptsize{(0.02)} & 0.52 \scriptsize{(0.02)} & 0.40 \scriptsize{(0.01)} & 0.61 \scriptsize{(0.01)} & 0.67 \scriptsize{(0.04)} & 0.75 \scriptsize{(0.02)} \\
\midrule
\multirow{3}{*}{InternVL3.5 30B A3B} & Simple & 0.94 \scriptsize{(0.02)} & 0.66 \scriptsize{(0.04)} & 0.60 \scriptsize{(0.02)} & 0.88 \scriptsize{(0.01)} & 0.59 \scriptsize{(0.04)} & 0.58 \scriptsize{(0.03)} \\
 & Medium & 0.92 \scriptsize{(0.01)} & 0.97 \scriptsize{(0.01)} & 0.62 \scriptsize{(0.01)} & 0.85 \scriptsize{(0.01)} & 0.72 \scriptsize{(0.04)} & 0.64 \scriptsize{(0.02)} \\
 & Hard & 0.90 \scriptsize{(0.02)} & 0.87 \scriptsize{(0.02)} & 0.71 \scriptsize{(0.02)} & 0.86 \scriptsize{(0.01)} & 0.61 \scriptsize{(0.05)} & 0.49 \scriptsize{(0.03)} \\
\midrule
\multirow{3}{*}{InternVL3.5 38B} & Simple & 0.86 \scriptsize{(0.03)} & 0.78 \scriptsize{(0.03)} & 0.52 \scriptsize{(0.02)} & 0.79 \scriptsize{(0.02)} & 0.47 \scriptsize{(0.04)} & 0.71 \scriptsize{(0.03)} \\
 & Medium & 0.76 \scriptsize{(0.02)} & 0.79 \scriptsize{(0.02)} & 0.47 \scriptsize{(0.01)} & 0.80 \scriptsize{(0.01)} & 0.50 \scriptsize{(0.04)} & 0.69 \scriptsize{(0.02)} \\
 & Hard & 0.85 \scriptsize{(0.01)} & 0.86 \scriptsize{(0.01)} & 0.55 \scriptsize{(0.01)} & 0.72 \scriptsize{(0.01)} & 0.74 \scriptsize{(0.02)} & 0.78 \scriptsize{(0.01)} \\
\midrule
\multirow{3}{*}{InternVL3.5 8B} & Simple & 0.75 \scriptsize{(0.04)} & 0.84 \scriptsize{(0.03)} & 0.39 \scriptsize{(0.03)} & 0.80 \scriptsize{(0.02)} & 0.55 \scriptsize{(0.05)} & 0.73 \scriptsize{(0.03)} \\
 & Medium & 0.70 \scriptsize{(0.03)} & 0.82 \scriptsize{(0.02)} & 0.50 \scriptsize{(0.01)} & 0.89 \scriptsize{(0.01)} & 0.56 \scriptsize{(0.04)} & 0.71 \scriptsize{(0.02)} \\
 & Hard & 0.72 \scriptsize{(0.02)} & 0.89 \scriptsize{(0.01)} & 0.61 \scriptsize{(0.01)} & 0.87 \scriptsize{(0.01)} & 0.62 \scriptsize{(0.03)} & 0.35 \scriptsize{(0.02)} \\
\midrule
\multirow{3}{*}{LLaVA-Onevision 72B} & Simple & 0.86 \scriptsize{(0.04)} & 0.50 \scriptsize{(0.13)} & \textbf{0.66 \scriptsize{(0.04)}} & 0.87 \scriptsize{(0.03)} & 0.27 \scriptsize{(0.06)} & 0.59 \scriptsize{(0.04)} \\
 & Medium & 0.95 \scriptsize{(0.01)} & 0.94 \scriptsize{(0.01)} & 0.63 \scriptsize{(0.01)} & \textbf{0.94 \scriptsize{(0.01)}} & 0.61 \scriptsize{(0.04)} & 0.67 \scriptsize{(0.02)} \\
 & Hard & 0.89 \scriptsize{(0.07)} & 0.61 \scriptsize{(0.09)} & \textbf{0.76 \scriptsize{(0.04)}} & 0.92 \scriptsize{(0.03)} & 0.53 \scriptsize{(0.13)} & 0.69 \scriptsize{(0.07)} \\
\midrule
\multirow{3}{*}{LLaVA-Onevision 7B} & Simple & 0.91 \scriptsize{(0.03)} & 0.74 \scriptsize{(0.05)} & 0.32 \scriptsize{(0.03)} & 0.77 \scriptsize{(0.02)} & 0.55 \scriptsize{(0.05)} & 0.45 \scriptsize{(0.03)} \\
 & Medium & 0.89 \scriptsize{(0.02)} & 0.94 \scriptsize{(0.02)} & 0.38 \scriptsize{(0.01)} & 0.87 \scriptsize{(0.01)} & 0.69 \scriptsize{(0.04)} & 0.53 \scriptsize{(0.02)} \\
 & Hard & 0.95 \scriptsize{(0.01)} & 0.92 \scriptsize{(0.02)} & 0.45 \scriptsize{(0.01)} & \textbf{0.93 \scriptsize{(0.01)}} & 0.79 \scriptsize{(0.04)} & 0.49 \scriptsize{(0.02)} \\
\midrule
\multirow{3}{*}{Mistral-Small-3.1 24B} & Simple & 0.77 \scriptsize{(0.04)} & 0.67 \scriptsize{(0.05)} & 0.44 \scriptsize{(0.03)} & 0.69 \scriptsize{(0.03)} & 0.43 \scriptsize{(0.05)} & 0.65 \scriptsize{(0.03)} \\
 & Medium & 0.59 \scriptsize{(0.03)} & 0.89 \scriptsize{(0.02)} & 0.46 \scriptsize{(0.01)} & 0.79 \scriptsize{(0.01)} & 0.40 \scriptsize{(0.05)} & 0.73 \scriptsize{(0.02)} \\
 & Hard & 0.70 \scriptsize{(0.04)} & 0.84 \scriptsize{(0.02)} & 0.59 \scriptsize{(0.02)} & 0.79 \scriptsize{(0.02)} & 0.81 \scriptsize{(0.03)} & 0.40 \scriptsize{(0.03)} \\
\midrule
\multirow{3}{*}{Molmo 7B} & Simple & \textbf{0.99 \scriptsize{(0.01)}} & \textbf{0.95 \scriptsize{(0.04)}} & 0.49 \scriptsize{(0.02)} & 0.80 \scriptsize{(0.02)} & 0.45 \scriptsize{(0.06)} & 0.63 \scriptsize{(0.03)} \\
 & Medium & 0.72 \scriptsize{(0.02)} & 0.92 \scriptsize{(0.02)} & 0.56 \scriptsize{(0.01)} & 0.87 \scriptsize{(0.01)} & 0.77 \scriptsize{(0.04)} & 0.59 \scriptsize{(0.02)} \\
 & Hard & 0.96 \scriptsize{(0.01)} & 0.90 \scriptsize{(0.01)} & 0.75 \scriptsize{(0.01)} & 0.87 \scriptsize{(0.01)} & 0.36 \scriptsize{(0.02)} & 0.56 \scriptsize{(0.01)} \\
\midrule
\multirow{3}{*}{Phi-4 Multimodal} & Simple & 0.00 \scriptsize{(0.00)} & 0.00 \scriptsize{(0.00)} & 0.00 \scriptsize{(0.00)} & 0.00 \scriptsize{(0.00)} & \textbf{0.67 \scriptsize{(0.19)}} & 0.00 \scriptsize{(0.00)} \\
 & Medium & 0.00 \scriptsize{(0.00)} & 0.17 \scriptsize{(0.11)} & 0.01 \scriptsize{(0.00)} & 0.02 \scriptsize{(0.01)} & 0.67 \scriptsize{(0.19)} & 0.10 \scriptsize{(0.04)} \\
 & Hard & 0.22 \scriptsize{(0.08)} & 0.26 \scriptsize{(0.08)} & 0.02 \scriptsize{(0.01)} & 0.27 \scriptsize{(0.05)} & 0.64 \scriptsize{(0.14)} & 0.05 \scriptsize{(0.03)} \\
\midrule
\multirow{3}{*}{Qwen2.5-VL 72B} & Simple & 0.90 \scriptsize{(0.02)} & 0.82 \scriptsize{(0.03)} & 0.65 \scriptsize{(0.02)} & 0.80 \scriptsize{(0.02)} & 0.53 \scriptsize{(0.04)} & 0.54 \scriptsize{(0.02)} \\
 & Medium & 0.87 \scriptsize{(0.02)} & 0.95 \scriptsize{(0.01)} & \textbf{0.66 \scriptsize{(0.01)}} & 0.86 \scriptsize{(0.01)} & 0.59 \scriptsize{(0.04)} & 0.58 \scriptsize{(0.02)} \\
 & Hard & 0.92 \scriptsize{(0.01)} & \textbf{0.97 \scriptsize{(0.01)}} & 0.66 \scriptsize{(0.01)} & 0.82 \scriptsize{(0.01)} & 0.76 \scriptsize{(0.02)} & 0.68 \scriptsize{(0.01)} \\
\midrule
\multirow{3}{*}{Qwen2.5-VL 7B} & Simple & 0.84 \scriptsize{(0.03)} & 0.84 \scriptsize{(0.03)} & 0.55 \scriptsize{(0.03)} & 0.83 \scriptsize{(0.02)} & 0.49 \scriptsize{(0.04)} & 0.61 \scriptsize{(0.03)} \\
 & Medium & 0.72 \scriptsize{(0.02)} & \textbf{0.98 \scriptsize{(0.00)}} & 0.52 \scriptsize{(0.01)} & 0.90 \scriptsize{(0.00)} & 0.67 \scriptsize{(0.03)} & 0.65 \scriptsize{(0.01)} \\
 & Hard & 0.61 \scriptsize{(0.01)} & 0.96 \scriptsize{(0.00)} & 0.46 \scriptsize{(0.00)} & 0.75 \scriptsize{(0.00)} & 0.45 \scriptsize{(0.02)} & 0.58 \scriptsize{(0.01)} \\
\midrule
\multirow{3}{*}{Qwen3-VL 30B A3B} & Simple & 0.92 \scriptsize{(0.02)} & 0.68 \scriptsize{(0.04)} & 0.53 \scriptsize{(0.02)} & 0.88 \scriptsize{(0.02)} & 0.37 \scriptsize{(0.04)} & 0.60 \scriptsize{(0.03)} \\
 & Medium & 0.90 \scriptsize{(0.01)} & 0.78 \scriptsize{(0.02)} & 0.58 \scriptsize{(0.01)} & 0.87 \scriptsize{(0.01)} & 0.56 \scriptsize{(0.05)} & 0.69 \scriptsize{(0.02)} \\
 & Hard & 0.94 \scriptsize{(0.01)} & 0.89 \scriptsize{(0.01)} & 0.72 \scriptsize{(0.01)} & 0.75 \scriptsize{(0.01)} & 0.75 \scriptsize{(0.02)} & 0.44 \scriptsize{(0.01)} \\
\midrule
\multirow{3}{*}{Qwen3-VL 32B} & Simple & 0.77 \scriptsize{(0.04)} & 0.77 \scriptsize{(0.04)} & 0.56 \scriptsize{(0.03)} & 0.74 \scriptsize{(0.02)} & 0.48 \scriptsize{(0.05)} & 0.69 \scriptsize{(0.03)} \\
 & Medium & 0.91 \scriptsize{(0.01)} & 0.90 \scriptsize{(0.01)} & 0.52 \scriptsize{(0.01)} & 0.84 \scriptsize{(0.01)} & 0.60 \scriptsize{(0.04)} & 0.76 \scriptsize{(0.02)} \\
 & Hard & 0.92 \scriptsize{(0.01)} & 0.92 \scriptsize{(0.01)} & 0.55 \scriptsize{(0.01)} & 0.77 \scriptsize{(0.01)} & 0.73 \scriptsize{(0.03)} & 0.64 \scriptsize{(0.02)} \\
\midrule
\multirow{3}{*}{Qwen3-VL 8B} & Simple & 0.85 \scriptsize{(0.03)} & 0.92 \scriptsize{(0.02)} & 0.64 \scriptsize{(0.02)} & 0.82 \scriptsize{(0.02)} & 0.52 \scriptsize{(0.03)} & \textbf{0.84 \scriptsize{(0.02)}} \\
 & Medium & 0.94 \scriptsize{(0.01)} & 0.89 \scriptsize{(0.01)} & 0.61 \scriptsize{(0.01)} & 0.73 \scriptsize{(0.01)} & 0.65 \scriptsize{(0.03)} & 0.77 \scriptsize{(0.01)} \\
 & Hard & \textbf{0.97 \scriptsize{(0.01)}} & 0.96 \scriptsize{(0.01)} & 0.57 \scriptsize{(0.01)} & 0.85 \scriptsize{(0.01)} & \textbf{0.92 \scriptsize{(0.01)}} & 0.51 \scriptsize{(0.01)} \\
\bottomrule
\end{tabular}}
\end{table*}